\documentclass[10pt,twocolumn,letterpaper]{article}
\usepackage{bm}

\usepackage[pagenumbers]{cvpr} 

\usepackage{graphicx}	
\usepackage{amsmath}	
\usepackage{amssymb}	
\usepackage{booktabs}
\usepackage{cuted}
\usepackage{times}
\usepackage{microtype}
\usepackage{epsfig}
\usepackage{caption}
\usepackage{float}
\usepackage{placeins}
\usepackage{color, colortbl}
\usepackage{stfloats}
\usepackage{enumitem}
\usepackage{tabularx}
\usepackage{xstring}
\usepackage{multirow}
\usepackage{xspace}
\usepackage{url}
\usepackage{subcaption}
\usepackage[hang,flushmargin]{footmisc}
\usepackage{kotex}
\usepackage{arydshln}
\usepackage{adjustbox}
\usepackage{wrapfig}
\usepackage{algorithm,algorithmicx,algpseudocode}
\usepackage{listings}
\usepackage{bm}

\newlength\paramargin
\newlength\abovetabcapmargin
\newlength\belowtabcapmargin
\newlength\abovefigcapmargin
\newlength\belowfigcapmargin
\newlength\aboveeqmargin
\newlength\beloweqmargin

\usepackage{amsmath}

\DeclareMathOperator*{\argmin}{arg\,min}

\definecolor{cvprblue}{rgb}{0.21,0.49,0.74}
\usepackage[pagebackref,breaklinks,colorlinks,allcolors=cvprblue]{hyperref}

\def\paperID{10357} 
\def\confName{CVPR}
\def\confYear{2025}

\title{SplatFlow: Multi-View Rectified Flow Model for 3D Gaussian Splatting Synthesis}

\author{
    Hyojun Go\textsuperscript{\rm 1}\thanks{Equal contribution} \qquad
    Byeongjun Park\textsuperscript{\rm 1,2}\footnotemark[1] \qquad
    Jiho Jang \qquad
    Jin-Young Kim \vspace{0.1cm} \\
    Soonwoo Kwon \qquad
    Changick Kim\textsuperscript{\rm 2}\thanks{Corresponding author} \vspace{0.3cm} \\
    \textsuperscript{\rm 1} EverEx \qquad
    \textsuperscript{\rm 2} KAIST 
    }

\begin{document}
\maketitle
\begin{strip}
    \vspace{-18mm}
    \centering
    \resizebox{\textwidth}{!}{
    \includegraphics[width=\textwidth]{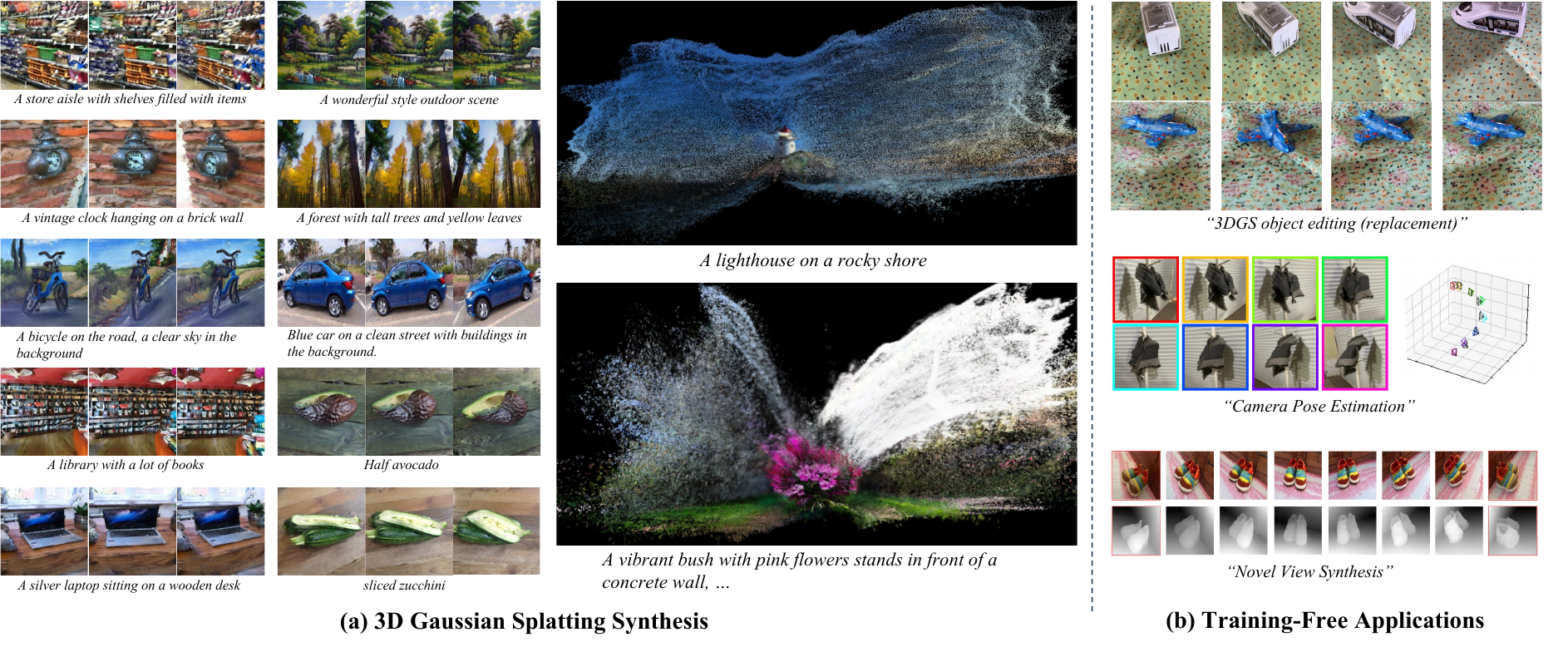}  
    }
    \vspace{\abovefigcapmargin}
    \vspace{-0.5cm}
    \captionof{figure}{\textbf{SplatFlow for 3D Gaussian Splatting synthesis and its training-free applications.} (a) Examples of direct 3D Gaussian Splatting (3DGS) generation only from text prompts, (b) Training-free applications, including 3DGS object editing, camera pose estimation, and novel view synthesis. SplatFlow seamlessly integrates these capabilities, showcasing its versatility in generating and editing complex 3D content.}
    \label{fig:teasure}
\end{strip}

\begin{abstract}
Text-based generation and editing of 3D scenes hold significant potential for streamlining content creation through intuitive user interactions.
While recent advances leverage 3D Gaussian Splatting (3DGS) for high-fidelity and real-time rendering, existing methods are often specialized and task-focused, lacking a unified framework for both generation and editing.
In this paper, we introduce SplatFlow, a comprehensive framework that addresses this gap by enabling direct 3DGS generation and editing.
SplatFlow comprises two main components: a multi-view rectified flow (RF) model and a Gaussian Splatting Decoder (GSDecoder).
The multi-view RF model operates in latent space, generating multi-view images, depths, and camera poses simultaneously, conditioned on text prompts—thus addressing challenges like diverse scene scales and complex camera trajectories in real-world settings.
Then, the GSDecoder efficiently translates these latent outputs into 3DGS representations through a feed-forward 3DGS method. 
Leveraging training-free inversion and inpainting techniques, SplatFlow enables seamless 3DGS editing and supports a broad range of 3D tasks—including object editing, novel view synthesis, and camera pose estimation—within a unified framework without requiring additional complex pipelines.
We validate SplatFlow's capabilities on the MVImgNet and DL3DV-7K datasets, demonstrating its versatility and effectiveness in various 3D generation, editing, and inpainting-based tasks. Our project page is available at \href{https://gohyojun15.github.io/SplatFlow/}{ https://gohyojun15.github.io/SplatFlow/}.
\end{abstract}    

\section{Introduction}
\label{sec:intro}
The demand for realistic 3D scene generation and editing from text has surged in applications like VR/AR, gaming, and robotics. Early works primarily utilized NeRF~\cite{mildenhall2021nerf, muller2022instant} for scene generation~\cite{zhang2024text2nerf} and editing~\cite{yu2023edit, park2023ed, khalid2023latenteditor, karim2025free, kamata2023instruct, bao2023sine}, but NeRF's volumetric rendering is computationally intensive, resulting in slower rendering speeds. Recently, 3D Gaussian Splatting (3DGS)~\cite{kerbl20233d} has emerged as a promising alternative, offering real-time and high-fidelity rendering implementations. Building on this, recent advances in 3D generation~\cite{zhuang2024tip, palandra2024gsedit, wang2025view, wu2024gaussctrl, wang2024gaussianeditor, chen2024gaussianeditor, khalid20253dego, zhang2024gaussiancube, xu2024dmvd, zhang2025gs, tang2025lgm} and editing~\cite{zhuang2024tip, palandra2024gsedit, wang2025view, wu2024gaussctrl, wang2024gaussianeditor, chen2024gaussianeditor, khalid20253dego} increasingly leverage 3DGS to achieve both speed and quality.

Despite these advancements, current methods for generating and editing 3DGS are specialized and task-focused, lacking a unified framework. For 3DGS generation, several works~\cite{chen2024text, tang2024dreamgaussian, yi2024gaussiandreamer, liu2024humangaussian, vilesov2023cg3d, li2025connecting} leverage 2D diffusion models~\cite{sohl2015deep, rombach2022high, ho2020denoising, song2020score} with Score Distillation Sampling (SDS)~\cite{poole2022dreamfusion}, which requires time-intensive per-scene optimization. To address this, recent studies have shifted towards direct 3DGS generation, combining diffusion models with reconstruction models~\cite{xu2024dmvd, zhang2025gs, tang2025lgm} or utilizing 3D generative models~\cite{zhang2024gaussiancube}. However, most of these methods are restricted to object-level 3D generation using synthetic datasets~\cite{deitke2023objaverse, deitke2024objaverse} with controlled, bounded environments.
In contrast, real-world scenes, varying in scene scales and camera trajectories, present distinctive challenges.

In 3DGS editing, several works~\cite{zhuang2024tip, palandra2024gsedit, wang2025view, wu2024gaussctrl, wang2024gaussianeditor, chen2024gaussianeditor, khalid20253dego} adopt 2D diffusion models~\cite{sohl2015deep, rombach2022high, ho2020denoising, song2020score} to guide the editing process. The main challenge is lifting 2D editing guidance to 3D while maintaining 3D-consistent modifications. One approach focuses on SDS techniques to update 3DGS~\cite{palandra2024gsedit, zhuang2024tip, xiao2024localized}, but these methods require additional stages like texture adjustments~\cite{palandra2024gsedit, xiao2024localized} and fine-grained refinements~\cite{zhuang2024tip}. Another approach aims to edit multi-view images with 3D consistency~\cite{wang2025view, wu2024gaussctrl, khalid20253dego}, introducing attention-based modules~\cite{wang2025view, wu2024gaussctrl} and autoregressive editing~\cite{khalid20253dego}, adding complexity to the editing pipeline.

In this paper, we bridge the gap between 3DGS generation and editing by designing a direct 3DGS generative model. Inspired by the success of 2D diffusion models that enable training-free editing through inversion~\cite{meng2021sdedit, couairon2022diffedit, mokady2023null}, we argue that a direct 3DGS model can similarly enable training-free editing. This approach integrates editing without requiring complex pipelines to lift 2D priors, making generation and editing adaptable within a unified framework.

We introduce SplatFlow, which consists of two main components: (1) a multi-view rectified flow (RF) model and (2) a Gaussian Splatting Decoder (GSDecoder). Similar to Latent Diffusion Models~\cite{rombach2022high}, our multi-view RF model operates in the latent space and is trained to simultaneously generate multi-view images, depths, and camera poses conditioned on text prompts. This joint modeling enables SplatFlow to effectively handle the challenges of real-world scene generation, such as various scene scales and camera trajectories, unlike synthetic object datasets~\cite{deitke2023objaverse, deitke2024objaverse}. The GSDecoder translates the latent outputs into 3DGS, based on efficient feed-forward methods~\cite{charatan2024pixelsplat, chen2025mvsplat}. To ensure compatibility with other generative models, particularly Stable Diffusion 3~\cite{esser2024scaling}, we incorporate a fixed pre-trained encoder, allowing flexible cross-model usage during sampling.

Finally, SplatFlow utilizes training-free inversion~\cite{meng2021sdedit, couairon2022diffedit, mokady2023null} and inpainting techniques~\cite{lugmayr2022repaint, rout2023theoretical, chung2022diffusion, rout2024solving, song2023pseudoinverse} for 3DGS editing and various 3D tasks. For example, it enables object editing~\cite{chen2024dge, palandra2024gsedit}, novel view synthesis~\cite{park2024bridging, rombach2021geometry, yu2024viewcrafter}, and camera pose estimation~\cite{zhang2024cameras} by jointly modeling images, depths, and poses to infer missing elements, highlighting its versatility. We validated its capabilities using the MVImgNet~\cite{yu2023mvimgnet} and DL3DV-7K~\cite{ling2024dl3dv} datasets, assessing its performance in 3D generation, editing, and inpainting-based tasks.

\section{Related Works}

\subsection{Diffusion Models and Rectified Flows}

\paragraph{Diffusion Models}
Diffusion models~\cite{sohl2015deep, ho2020denoising, song2020score} generate data through iterative denoising and have become standard in generative modeling across images~\cite{dhariwal2021diffusion, park2024switch, park2023denoising, lee2024multi, go2024addressing}, videos~\cite{ho2022imagen}, and text~\cite{li2022diffusion}. 
Trained on large-scale text-to-image datasets, they capture complex semantic relationships, producing high-quality images~\cite{rombach2022high, nichol2021glide, podell2023sdxl, chen2023pixart, sauer2025adversarial}.
An advantage is their adaptability to various tasks in a plug-and-play manner~\cite{graikos2022diffusion, go2023towards, song2023loss}, especially inpainting~\cite{lugmayr2022repaint, rout2023theoretical} and editing~\cite{hertz2022prompt, kim2022diffusionclip}. 
In painting can be framed as a linear inverse problem~\cite{daras2024survey}, for which effective posterior sampling methods have been proposed~\cite{lugmayr2022repaint, rout2023theoretical, chung2022diffusion, rout2024solving, song2023pseudoinverse}.
For editing, inversion methods~\cite{meng2021sdedit, couairon2022diffedit, mokady2023null} produce structured noise by inverting images, allowing new prompts to be effectively processed.
Building on these training-free techniques, we extend them to our 3DGS model, adapting RePaint~\cite{lugmayr2022repaint} and SDEdit~\cite{meng2021sdedit} within rectified flow models.

\vspace{\paramargin}
\paragraph{Rectified Flow Models}
Rectified Flow (RF) models~\cite{liu2022flow, albergo2022building, lipman2022flow} represent an alternative approach to traditional diffusion models by utilizing a straight-line path from data to noise instead of the typical curved forward process, aiming to simplify sampling and reduce computational costs. This linear path theoretically enables faster sampling by minimizing error accumulation across steps, and recent advances, such as Stable Diffusion 3 (SD3)~\cite{esser2024scaling}, have successfully scaled RF in text-to-image generation, demonstrating performance that surpasses traditional diffusion models.
Our SplatFlow framework builds on the capabilities of SD3, integrating RF to achieve efficient 3DGS generation.

\subsection{3D Generative Models}

\paragraph{Neural Scene Representation}
Various neural scene representations have been proposed for 3D generation tasks. Early works use explicit representations, enabling synthesis of 3D point clouds~\cite{nichol2022point} and shapes~\cite{zhou20213d, vahdat2022lion}, but struggle with realistic scene generation due to their rendering mechanisms. 
To address this, NeRF~\cite{mildenhall2021nerf} is employed for a realistic 3D generation with per-scene optimization~\cite{poole2022dreamfusion}, but its inefficient rendering is unsuitable for a fast generation.
Recently, 3DGS~\cite{kerbl20233d} enables real-time rendering, accelerating optimization-based 3D generation~\cite{tang2024dreamgaussian, ling2024align, chen2024text}. 
Building on this progress, we leverage 3DGS's efficient rendering for fast training of a generative model on large datasets.

\vspace{\paramargin}
\paragraph{Lifting Multi-View Generation}
Advances in image generation models~\cite{podell2023sdxl, chen2023pixart} and large-scale 3D object datasets~\cite{deitke2023objaverse, deitke2024objaverse} make multi-view generation compelling compared to optimization-based generation~\cite{tang2024dreamgaussian}.
The key idea is to learn 3D priors as conditions~\cite{liu2023syncdreamer, woo2024harmonyview, long2024wonder3d} or attentions~\cite{shi2023mvdream, voleti2025sv3d} from 3D datasets while leveraging text-to-image diffusion models~\cite{rombach2022high} trained on billions of text-image pairs~\cite{schuhmann2022laion}. 
This enables the synthesis of multi-view consistent images from an image or text prompt, which is then fed into reconstruction models~\cite{chen2025lara, tang2025lgm} to directly generate neural scene representations like 3DGS.
One critical success factor in these methods is that the training dataset provides rendered images from consistent, canonical viewpoints, allowing instant access to desired views regardless of the 3D object.
In this work, we focus on more challenging unbounded 3D real-world scene generation, where each scene varies in scale and camera trajectory. To handle diverse trajectories, our model learns a joint distribution of camera poses and 3D scenes.

\vspace{\paramargin}
\paragraph{Text-to-3D Scene Generation}
Generating 3D scenes from text prompts is challenging due to the unbounded scale of scenes, complicating scale determination.
One straightforward approach is to create a scene covered by a single image~\cite{shriram2024realmdreamer,zhou2025dreamscene360} generated from text-to-image diffusion models~\cite{podell2023sdxl,chen2023pixart}.
Another line of work~\cite{chung2023luciddreamer, hollein2023text2room} generates wider scenes using user-provided camera trajectories, but these methods are time-consuming and produce subpar quality due to reliance on 2D generative models.
Recently, Director3D~\cite{li2024director3d}, the work most similar to ours, generates camera poses from text and then creates multi-view images decoded into 3DGS. In contrast, SplatFlow jointly learns the distribution of camera poses and multi-view images from text, enhancing generation quality and enabling flexible 3D scene editing in a unified framework.

\subsection{3D Editing}
In recent years, NeRF-based methods~\cite{yu2023edit, yi2024diffusion, richardson2023texture, park2023ed, park2024point, khalid2023latenteditor, karim2025free, kamata2023instruct, bao2023sine} have gained popularity for 3D editing, surpassing point clouds and meshes~\cite{wang2023mesh, zhang2024point, zhang2024dragtex}.
However, due to the inefficient rendering, these methods are time-consuming and lack precise control. Conversely, incorporating 3DGS into 3D editing~\cite{chen2024gaussianeditor, wang2024gaussianeditor} improves effectiveness, speed, and control.

For guiding 3D editing, several methods have leveraged 2D diffusion models~\cite{sohl2015deep, rombach2022high, ho2020denoising, song2020score} for structured 3D editing, but it often requires additional refinement stages.
For instance, GSEdit~\cite{palandra2024gsedit} and Xiao~\etal~\cite{xiao2024localized} apply texture enhancement, while TIP-Editor~\cite{zhuang2024tip} adds fine-grained adjustments to reduce SDS artifacts.
Another line of research in 3D editing emphasizes achieving multi-view consistency by sequentially updating each view. 
For example, GaussCtrl~\cite{wu2024gaussctrl} and 3DEdit~\cite{wang2025view} utilize attention-based modules to propagate edits made from one perspective across multiple views, ensuring that changes are consistently reflected throughout the 3D scene. Similarly, 3D-Ego~\cite{khalid20253dego} adopts an autoregressive editing framework, updating each view to enhance multi-view consistency.

Unlike prior methods that require complex procedures to lift 2D priors for 3D consistency, SplatFlow directly models multi-view consistency within the latent space, efficiently decoding edits through our Gaussian Splatting Decoder.

\section{Preliminary: Rectified Flows}
Rectified Flows (RFs)~\cite{liu2022flow, albergo2022building, lipman2022flow} consider generative models that construct a noise distribution $q_0$ and a time-varying vector field $v_t(\bm{x}_t)$ to sample data $x_0$ from a data distribution $p_0$ via an ordinary differential equation (ODE) as:
\begin{equation}
    \label{eq:rectified_flow_ode}
    \mathrm{d}X_t = v(X_t) \, \mathrm{d}t, \quad X_0 \sim q_0, \quad t \in [0, 1],
\vspace{\beloweqmargin}
\end{equation}
where $q_0$ typically follows a Gaussian distribution $\mathcal{N}(0, I)$.
RF employs a neural network with parameters $\theta$ for modeling the vector field as $v_t(X_t) = -u_\theta(X_t, 1-t)$.

To train a neural network as the vector field, RFs utilize paired samples from $p_0$ and $q_0$ (here simplified as $p_1$), along a straight line path $Y_t=tY_1 + (1-t) Y_0$. This setup produces the marginal distribution of $Y_t$ as $p_t(\bm{y}_t) = \int p_t(\bm{y}_t |\bm{y}_1) p_1(\bm{y}_1) \mathrm{d}\bm{y}_1.$ Starting from an initial $Y_0=\bm{y}_0$ and moving to a final state $Y_1=\bm{y}_1$, a straight line path induces an ODE expressed as $\mathrm{d}Y_t = u_t(Y_t | Y_1)\mathrm{d}t$ where the vector field $u_t(Y_t|Y_1)$ represents the conditional change from $\bm{y}_0$ to $\bm{y}_1$, given by $u_t(Y_t|Y_1) = \bm{y}_1-\bm{y}_0$.
The marginal vector field $u_t(\bm{y}_t)$ is computed by averaging over the conditional vector field as~\cite{lipman2022flow}:
\begin{equation} 
u_t(\bm{y}_t) = \int u_t(\bm{y}_t | \bm{y}_1) \frac{p_t(\bm{y}_t | \bm{y}_1)}{p_t(\bm{y}_t)} p_1(\bm{y}_1)\mathrm{d}\bm{y}_1. 
\vspace{\beloweqmargin}
\end{equation}
To approximate $u_t(\bm{y}_t)$, the neural network is trained with a flow-matching object, defined by: 
\begin{equation} 
\mathcal{L}_{\text{FM}} := \mathbb{E}_{t, Y_t} \left[ \left|\left| u_t(Y_t) - u_\theta(Y_t, t) \right|\right|_2^2 \right]. 
\vspace{\beloweqmargin}
\end{equation}
Since directly computing this objective can be computationally challenging, an alternative approach, conditional flow matching, is used to simplify training: 
\begin{equation} 
~\label{eq:cfm_loss}
\mathcal{L}_{\text{CFM}} := \mathbb{E}_{t, Y_t, Y_1\sim p_1}\left[ \left\| u_t(Y_t | Y_1) - u_\theta(Y_t, t) \right\|_2^2 \right].
\vspace{\beloweqmargin}
\end{equation}
The gradients of $L_{\text{CFM}}$ and $L_{\text{FM}}$ are theoretically identical~\cite{lipman2022flow}, making them equivalent in training.
However, $L_{\text{CFM}}$ offers better computational efficiency and is therefore preferred.
Ultimately, the learned vector field for the ODE in Eq.~\ref{eq:rectified_flow_ode} is given by $v_t(X_t)=-u_\theta(X_t, 1-t)$, allowing RFs to generate samples that follow the data distribution using the learned vector field.



\section{Methodology}
\label{sec:method}
\begin{figure*}[t]
    \centering
    \includegraphics[width=0.95\linewidth]{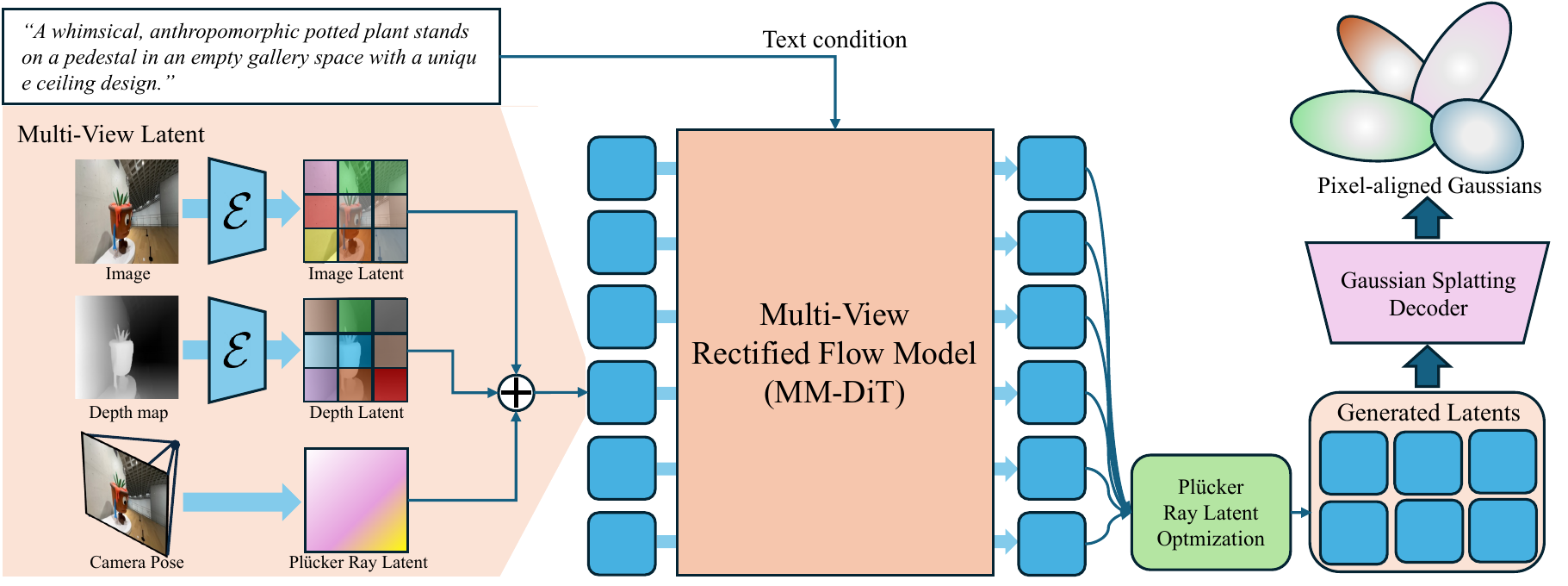}

    \vspace{\abovefigcapmargin}
    \vspace{-0.1cm}
    \caption{\textbf{Overview of SplatFlow.} SplatFlow consists of two main components: a multi-view Rectified Flow (Section~\ref{sec:4.3_multiview_rf}) model and a Gaussian Splat Decoder (Section~\ref{sec:4.2_gsdecoder}). Conditioned on text, RF model generates multi-view latents—including image, depth, and Plücker ray coordinates. After an optimization process to estimate camera poses, the GSDecoder decodes these latents into pixel-aligned 3DGS.}
    \vspace{\belowfigcapmargin}
    \label{fig:architecture_figure}
\end{figure*}

\subsection{Overview of SplatFlow}

We introduce \textit{SplatFlow}, a 3DGS generation model designed to perform various 3D editing and 3D-related tasks in a training-free manner, despite being trained solely for generation. As shown in Fig.~\ref{fig:architecture_figure}, our model consists of two main components: 1) a multi-view Rectified Flow (RF) model and 2) a Gaussian Splatting Decoder (GSDecoder).
First, the multi-view RF model generates multi-view camera poses, depths, and images conditioned on text prompts. Then, the GSDecoder translates these latent representations into pixel-aligned 3DGS, constructing a real-world scene structure.

Also, both the multi-view RF model and GSDecoder operate within a latent space, adopting a Latent Diffusion Model (LDM)~\cite{rombach2022high}. By freezing and sharing the encoder of Stable Diffusion 3 (SD3)~\cite{esser2024scaling}, our design enables the RF model to share the latent space with SD3, allowing us to leverage its knowledge for enhanced generation and editing capabilities.

Finally, our model leverages a training-free approach~\cite{lugmayr2022repaint, meng2021sdedit}, enabling a broad range of editing applications, such as object replacement, as well as 3D-related tasks including camera pose estimation and novel view synthesis.

In the following sections, we present each component in detail. Section~\ref{sec:4.2_gsdecoder} introduces the GSDecoder, explaining details in constructing pixel-aligned 3DGS. Section~\ref{sec:4.3_multiview_rf} describes the multi-view RF model, detailing how it jointly generates camera poses, depths, and images. Finally, Section~\ref{sec:4.4_sampling_inpainting} outlines the sampling process and demonstrates how our model applies training-free techniques.

\subsection{Gaussian Splatting Decoder (GSDecoder)}
\label{sec:4.2_gsdecoder}

Recently, feed-forward 3DGS methods~\cite{charatan2024pixelsplat, chen2025mvsplat, szymanowicz2024splatter, tang2025lgm} have enabled fast 3DGS reconstruction from sparse views by training on large datasets, achieving much faster reconstruction than per-scene optimization methods~\cite{kerbl20233d}.
Leveraging this advantage, our GSDecoder $G_\phi$ (parameterized by $\phi$) is designed to decode pixel-aligned 3DGS from latent representations of $K$ sparse views, given their corresponding camera poses $\mathcal{P}=\{\bm{P}_i\}_{i=1}^K$ with $\bm{P}_i = \bm{K}_i \left[ \bm{R}_i|\bm{T}_i \right]$, where $\bm{K}_i$, $\bm{R}_i$, and $\bm{T}_i$ denote the intrinsic matrix, rotation matrix, and translation vector of the $i$-th view, respectively.

A straightforward approach is to design $G_\phi$ to directly output the 3D Gaussian parameters from the image latents obtained through the encoder $\mathcal{E}$ as $G_\phi(\{ (\mathcal{E}(\bm{I}_i), \bm{P}_i )\}_{i=1}^K)=\{(\bm{\mu}_j, \alpha_j, \bm{\Sigma}_i, \bm{c}_j)\}_{j=1}^{H \times W \times K}$, where each 3D Gaussian parameter includes position $\bm{\mu}_j$, opacity $\alpha_j$, covariance $\bm{\Sigma}_j$, and color $\bm{c}_j$ in a pixel-aligned manner with each $i$-th image $\bm{I}_i \in \mathbb{R}^{H \times W}$. 
However, as the image latents are produced through a shared and frozen encoder (to leverage compatibility with 2D generative models during sampling), it might lose fine-grained spatial details, leading to abstract representations. 
This can limit the preservation of specific 3D structural information. 
To address these limitations, we propose two improvements to enhance GSDecoder's performance.

\vspace{\paramargin}
\paragraph{Depth latent integration}
As demonstrated in~\cite{ke2024repurposing}, the encoder $\mathcal{E}$ can effectively encode depth maps. 
Building on this, we incorporate the depth latents $\{\mathcal{E}(\bm{D}_i)\}_{i=1}^K$ of each depth map $\bm{D}_i$ as an additional input to the GSDecoder to further enhance 3D structural information.
We use DepthAnythingV2~\cite{yang2024depth} to extract the depth maps and normalize them to the range [-1, 1], stacking them as three channels similar to RGB for constructing $\bm{D}_i$.
We observe that this improves the convergence speed and performance of the GSDecoder.

\vspace{\paramargin}
\paragraph{Adversarial loss}
Applying adversarial losses has demonstrated improvements in the visual quality of decoded output~\cite{esser2021taming, rombach2022high}. 
However, using adversarial loss during training can cause instability, particularly in the early stages. We observe that applying the vision-aided loss~\cite{kumari2022ensembling} after the GSDecoder has reached a certain level of convergence significantly improves visual quality without destabilizing training.

\vspace{\paramargin}
\paragraph{Training} 
In summary, our GSDecoder predicts 3DGS $\{(\bm{\mu}_j, \alpha_j, \bm{\Sigma}_j, \bm{c}_j)\}_{j=1}^{H \times W \times K} $ from $\{(\bm{I}_i, \bm{D}_i, \bm{P}_i)\}_{i=1}^K$.
The model is trained with a combined loss function consisting of LPIPS~\cite{zhang2018unreasonable}, Mean Squared Error loss, and vision-aided loss~\cite{kumari2022ensembling}, computed between the target images of novel views and rendered views from 3DGS. 
Specifically, the vision-aided loss is activated only after a certain threshold iteration to improve stability and perceptual quality.
Additionally, the architecture is initialized with the parameters of the Stable Diffusion 3 decoder, incorporating cross-view attention mechanisms, and the channel dimensions are increased to meet the specific requirements of 3D Gaussian Splatting. Further details on the GSDecoder are provided in Appendix~\textcolor{cvprblue}{A}.

\subsection{Multi-View Rectified Flow Model}
\label{sec:4.3_multiview_rf}

With the GSDecoder in place, the 3D Gaussian Splatting (3DGS) generation process becomes feasible if we can produce consistent multi-view latents of images \( \mathcal{E}(\bm{I}_i) \), depth maps \( \mathcal{E}(\bm{D}_i) \), and camera poses \( \bm{P}_i \) for \( i = 1, \dots, K \). Our multi-view rectified flow (RF) model is trained to sample from the joint distribution of these latents, conditioned on text \( C \) as \( p(\{\bm{I}_i\}_{i=1}^K, \{\bm{D}_i\}_{i=1}^K, \{\bm{P}_i\}_{i=1}^K | C) \), allowing for simultaneous generation of multi-view images, depths, and poses. The rationale for modeling a joint distribution instead of each component separately is twofold. First, this approach allows the model to handle various tasks by reformulating them as inpainting problems, as discussed in Section~\ref{sec:4.4_sampling_inpainting}. Second, joint modeling is crucial for real-world scenes, which require adaptive camera poses tailored to each scene~\cite{li2024director3d}.

To achieve this, our multi-view RF model treats the concatenation of multi-view image latents, depth latents, and Plücker ray coordinates~\cite{plucker1828analytisch, zhang2024cameras} along the channel dimension as input data. Formally, each ray \( \bm{r}_i \) is represented as \( \langle \bm{d}_i, \bm{m}_i \rangle \), where \( \bm{d}_i = \bm{R}_i^\top \bm{K}_i^{-1} \bm{w}_i \) denotes the direction vector and \( \bm{m}_i = (-\bm{R}_i^\top \bm{T}_i) \times \bm{d}_i \) represents the moment vector, with \( \bm{w}_i \) referring to the 2D pixel coordinates. By matching the spatial resolution of \( \bm{w}_i \) with the image and depth latents, each represented as \( \mathbb{R}^{n \times h \times w} \), \( \bm{r}_i \) can be expressed as \( \mathbb{R}^{6 \times h \times w} \), enabling concatenation along the channel axis.

Through this setup, we define \( Y_0 \) as the multi-view latents \( \bm{X} \), where each element \( \bm{X}_i = \langle \mathcal{E}(\bm{I}_i), \mathcal{E}(\bm{D}_i), \bm{r}_i \rangle\) results in \( Y_0 = (\bm{X}_1, \bm{X}_2, \dots, \bm{X}_K) \in \mathbb{R}^{K \times (2n + 6) \times h \times w} \). We then train the multi-view RF model \( u_\theta \) with the conditional flow matching loss defined in Eq.~\ref{eq:cfm_loss}. Specifically, we fine-tune Stable Diffusion 3~\cite{esser2024scaling} by adjusting the channel dimensions in the input and output layers and incorporating cross-view attention. Further details are provided in Appendix~\textcolor{blue}{A}.

\vspace{\paramargin}
\paragraph{Sampling process} To recover the camera pose from ray \( \bm{r}_i \), we follow the procedure in~\cite{zhang2024cameras} and further refine the pose parameters to ensure that multiple views share the same intrinsic. However, we find that naively solving the ODE can lead to degraded camera pose accuracy. We hypothesize this degradation results from deviations from the camera pose manifold, which introduces errors in camera trajectory estimation.
To address this, at each sampling step \( t = t_k \), we predict the sampling destination at \( t = 0 \), recover the camera poses from this prediction, and reconstruct the associated camera rays. These rays are then forwarded to \( t = t_k \) to maintain the solution within the valid ray manifold at \( t = t_k \).

\vspace{\paramargin}
\paragraph{Stable Diffusion 3 guidance} To improve multi-view image generation quality, we integrate vector fields from Stable Diffusion 3, known for robust generalization, with our multi-view RF model to solve the ODE for randomly selected view's image latents at specific sampling steps.

\subsection{Inference Given Partial Data}
\label{sec:4.4_sampling_inpainting}

In this section, we demonstrate how SplatFlow can be effectively utilized for 3D editing and applications by applying training-free inversion techniques~\cite{meng2021sdedit, couairon2022diffedit, mokady2023null} and inpainting methods~\cite{lugmayr2022repaint, rout2023theoretical, chung2022diffusion, rout2024solving, song2023pseudoinverse}. 
We modify SDEdit~\cite{meng2021sdedit} inversion for rectified flow-based inversion and adopt the RePaint~\cite{lugmayr2022repaint} to support rectified flow inpainting. Additional technical details are provided in Appendix~\textcolor{cvprblue}{A}.


\vspace{\paramargin}
\paragraph{3DGS editing}
Our GSDecoder is designed to directly decode 3DGS, enabling efficient editing by modifying only the multi-view latents. Additionally, with the RF model's joint modeling of multi-view image latents, edits can be achieved without extra modules (e.g., cross-view attention mechanisms~\cite{wu2024gaussctrl, wang2025view}), streamlining the process through simple inversion and sampling. Specifically, given a target text and input multi-view latents, we apply a modified SDEdit for inversion at a chosen timestep $t_k$. Conditioned on the target text, we then generate the edited multi-view latents by solving the ODE in Eq.~\ref{eq:rectified_flow_ode} following our sampling procedure.

\vspace{\paramargin}
\paragraph{Inpainting application}
Our model is also applicable to various 3D tasks, as it jointly models multi-view images, depths, and camera poses. This joint modeling allows known parts of the data (e.g., some views or camera parameters) to act as constraints, enabling the prediction of unknown parts through inpainting. In this work, we focus on two specific tasks: 1) \textit{camera pose estimation} from multi-view images and depths, and 2) \textit{novel view synthesis} using a subset of the \( K \) multi-view latents and camera poses for novel viewpoints.

\section{Experimental Results}
\label{sec:exp}

In this section, we demonstrate the capabilities of our proposed SplatFlow across 3DGS generation, 3DGS editing, and various 3D tasks. We begin by detailing the implementation of SplatFlow in Section~\ref{sec:exp:implenetation}. Following this, we present comparative results on 3DGS generation in real-world datasets in Section~\ref{sec:exp:3dgs_gen}. In Section~\ref{sec:exp:sdgs_editing}, we evaluate the 3DGS editing performance of SplatFlow against other methods. Finally, Section~\ref{sec:exp:inpainting_result} showcases the versatility of our multi-view RF model in handling diverse 3D tasks through training-free inpainting techniques. Extensive information on all our experiments is provided in
Appendix \textcolor{cvprblue}{B}
with additional results that could not be included here in the Appendix.

\begin{figure*}[t!]
    \centering
    \setlength\tabcolsep{2pt}
    
    \begin{tabular}{c:cc:cc:cc:cc}
        \multirow{2}{*}{%
            \parbox{0.1\linewidth}{%
                \vspace{-2pt}%
                \centering
                \footnotesize
                \textit{Stack of fries on a checkered surface}%
                    }%
        } 
        &
        \adjincludegraphics[clip,width=0.1\linewidth,trim={0 0 0 0}]{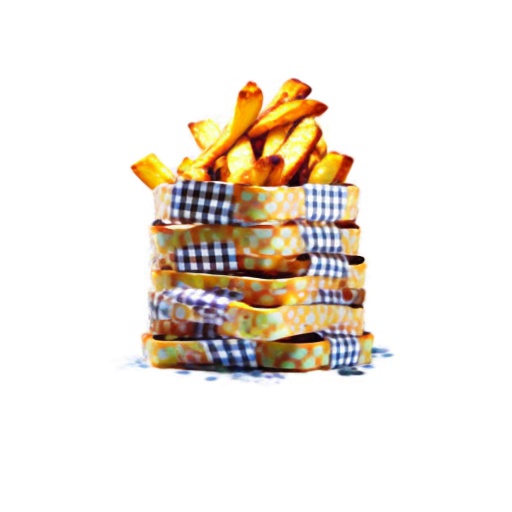} &
        \adjincludegraphics[clip,width=0.1\linewidth,trim={0 0 0 0}]{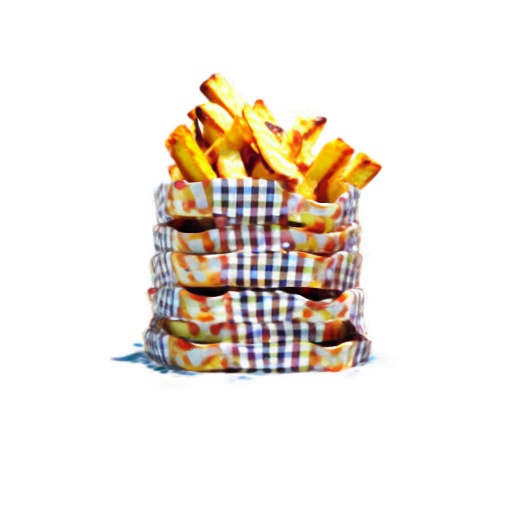} &
          \adjincludegraphics[clip,width=0.1\linewidth,trim={0 0 0 0}]{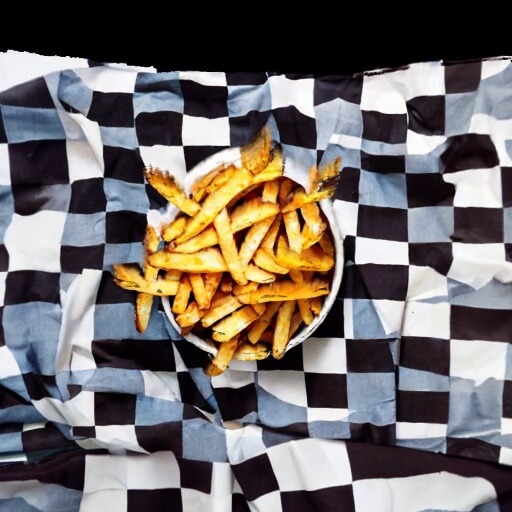} &
        \adjincludegraphics[clip,width=0.1\linewidth,trim={0 0 0 0}]{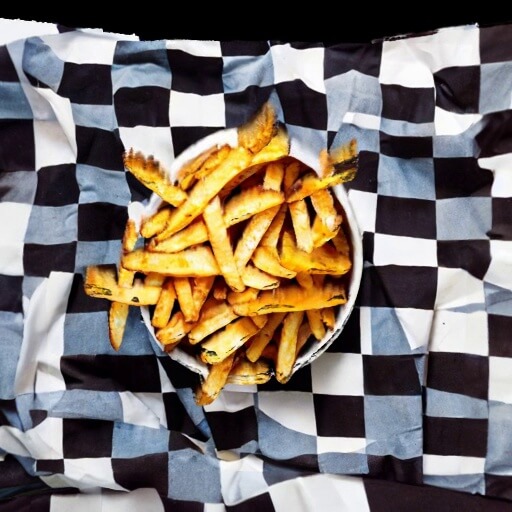} &
        \adjincludegraphics[clip,width=0.1\linewidth,trim={0 0 0 0}]{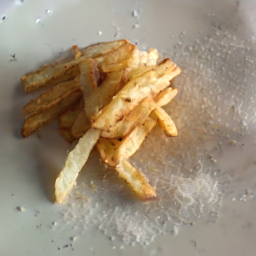} &
        \adjincludegraphics[clip,width=0.1\linewidth,trim={0 0 0 0}]{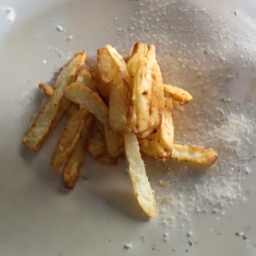} &
        \adjincludegraphics[clip,width=0.1\linewidth,trim={0 0 0 0}]{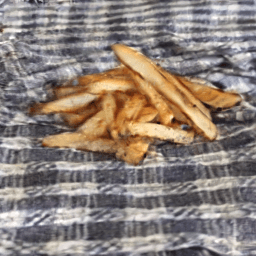} &
        \adjincludegraphics[clip,width=0.1\linewidth,trim={0 0 0 0}]{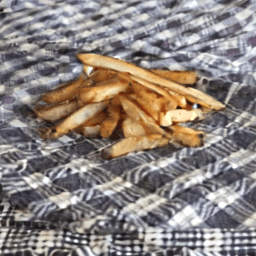} \\
         &
        \adjincludegraphics[clip,width=0.1\linewidth,trim={0 0 0 0}]{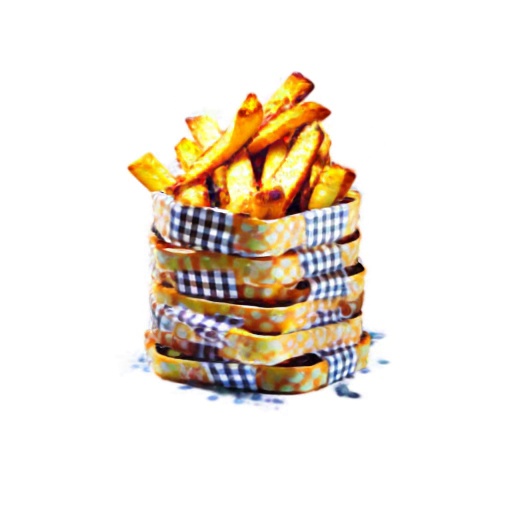} &
        \adjincludegraphics[clip,width=0.1\linewidth,trim={0 0 0 0}]{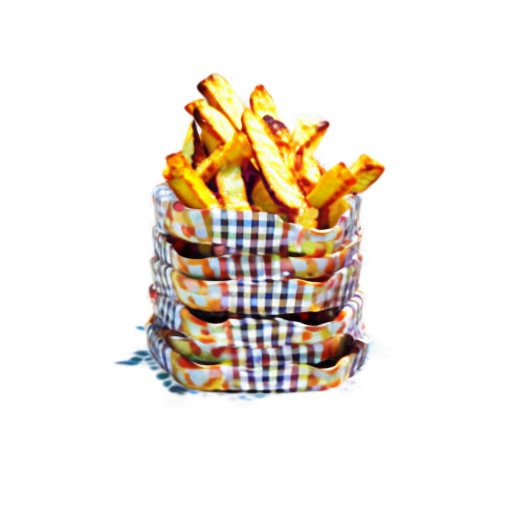} &
        \adjincludegraphics[clip,width=0.1\linewidth,trim={0 0 0 0}]{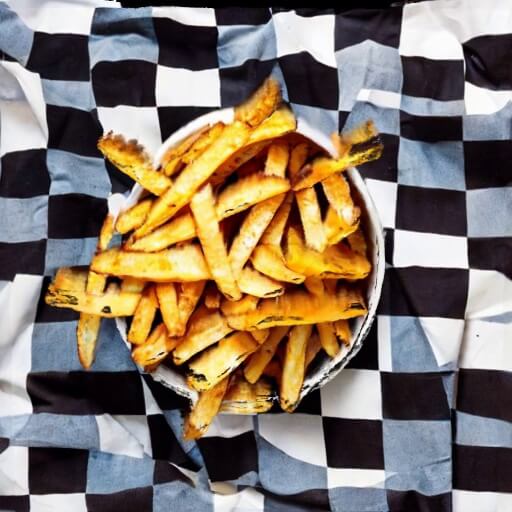} &
        \adjincludegraphics[clip,width=0.1\linewidth,trim={0 0 0 0}]{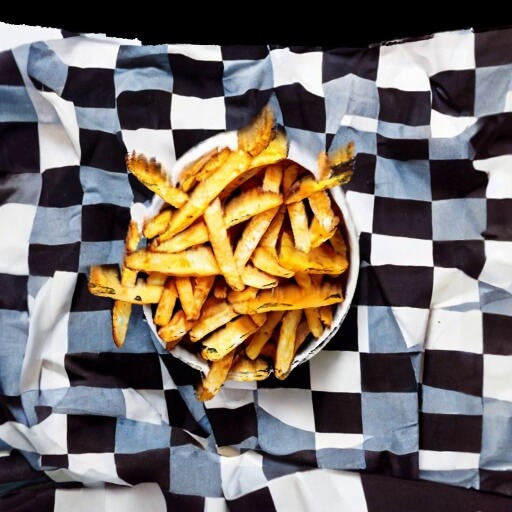} &
        \adjincludegraphics[clip,width=0.1\linewidth,trim={0 0 0 0}]{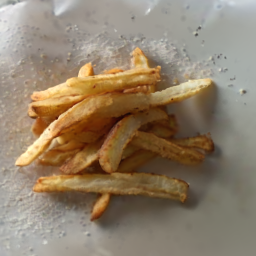} &
        \adjincludegraphics[clip,width=0.1\linewidth,trim={0 0 0 0}]{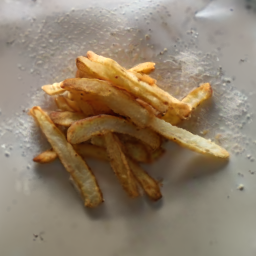} &
        \adjincludegraphics[clip,width=0.1\linewidth,trim={0 0 0 0}]{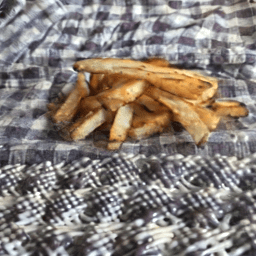} &
        \adjincludegraphics[clip,width=0.1\linewidth,trim={0 0 0 0}]{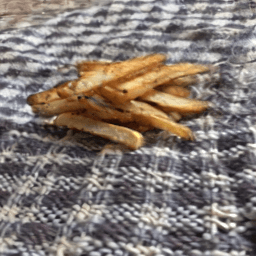} \\
        \arrayrulecolor{gray}\midrule
        \multirow{2}{*}{%
            \parbox{0.1\linewidth}{%
                \vspace{-8pt}%
                \centering
                \footnotesize
                \textit{Broom with straw bristles, leaning against a sign.}%
                    }%
        }  &
        \adjincludegraphics[clip,width=0.1\linewidth,trim={0 0 0 0}]{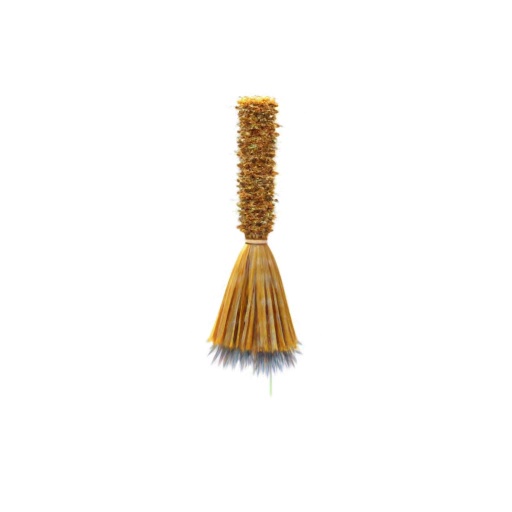} &
        \adjincludegraphics[clip,width=0.1\linewidth,trim={0 0 0 0}]{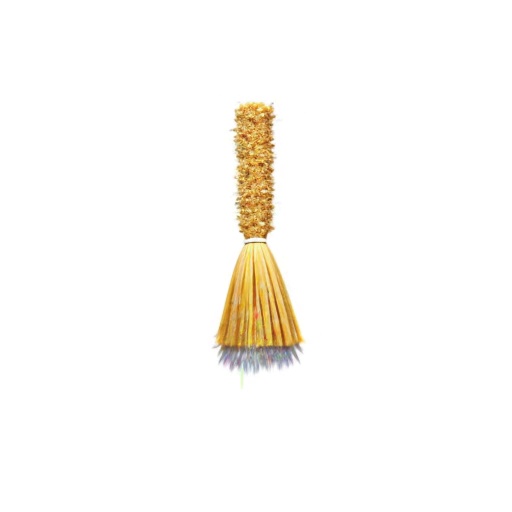} &
        \adjincludegraphics[clip,width=0.1\linewidth,trim={0 0 0 0}]{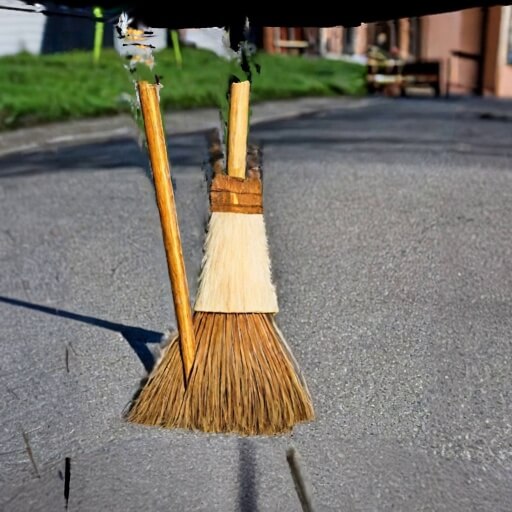} &
        \adjincludegraphics[clip,width=0.1\linewidth,trim={0 0 0 0}]{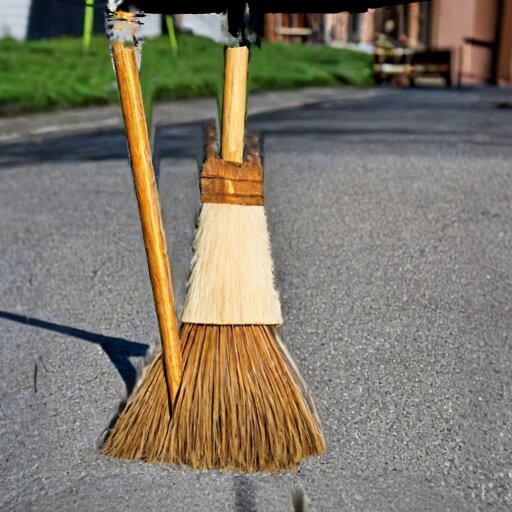} &
        \adjincludegraphics[clip,width=0.1\linewidth,trim={0 0 0 0}]{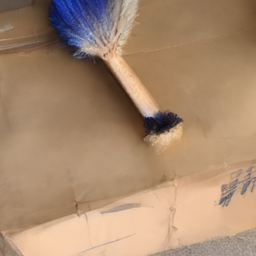} &
        \adjincludegraphics[clip,width=0.1\linewidth,trim={0 0 0 0}]{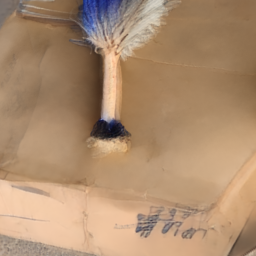} &
        \adjincludegraphics[clip,width=0.1\linewidth,trim={0 0 0 0}]{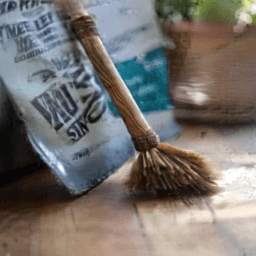} &
        \adjincludegraphics[clip,width=0.1\linewidth,trim={0 0 0 0}]{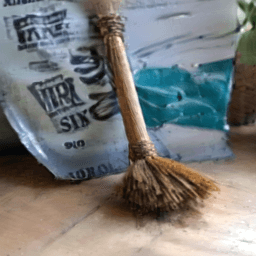} \\
         &
        \adjincludegraphics[clip,width=0.1\linewidth,trim={0 0 0 0}]{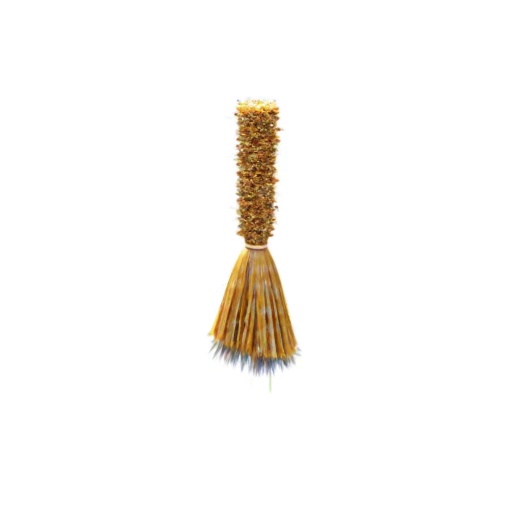} &
        \adjincludegraphics[clip,width=0.1\linewidth,trim={0 0 0 0}]{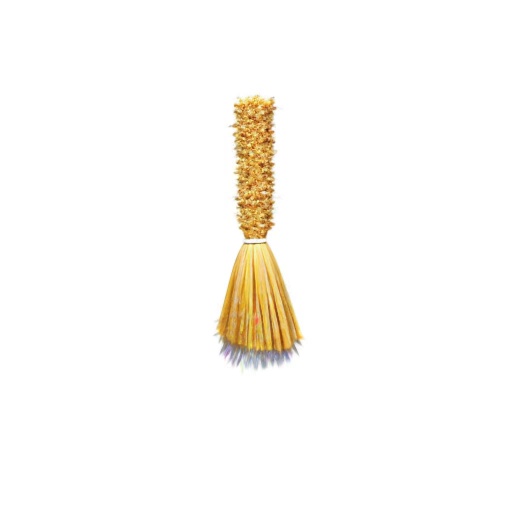} &
        \adjincludegraphics[clip,width=0.1\linewidth,trim={0 0 0 0}]{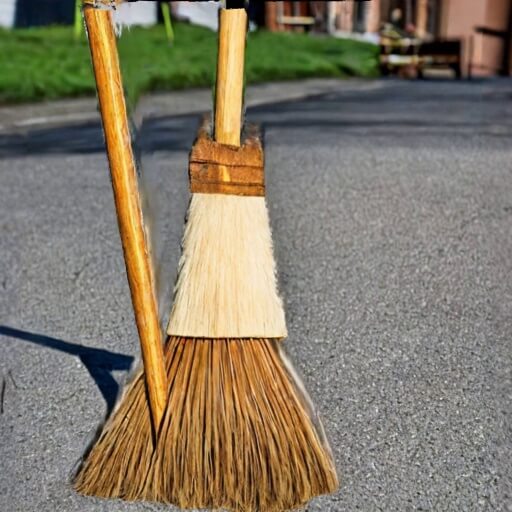} &
        \adjincludegraphics[clip,width=0.1\linewidth,trim={0 0 0 0}]{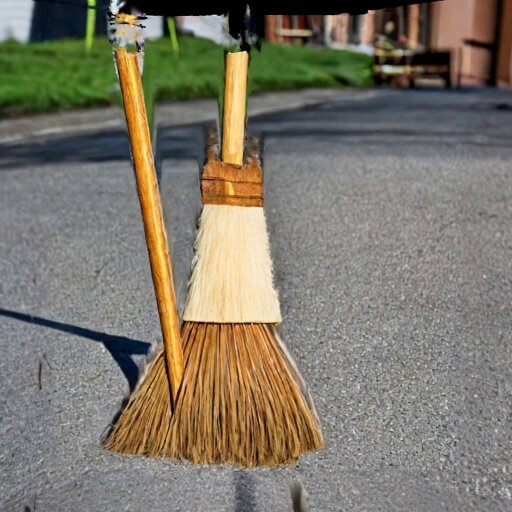} &
        \adjincludegraphics[clip,width=0.1\linewidth,trim={0 0 0 0}]{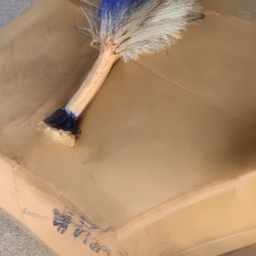} &
        \adjincludegraphics[clip,width=0.1\linewidth,trim={0 0 0 0}]{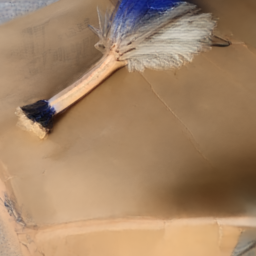} &
        \adjincludegraphics[clip,width=0.1\linewidth,trim={0 0 0 0}]{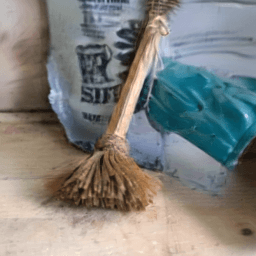} &
        \adjincludegraphics[clip,width=0.1\linewidth,trim={0 0 0 0}]{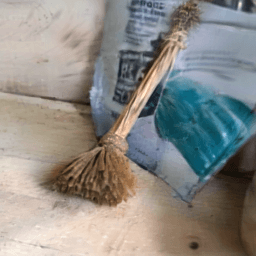} \\
        \midrule

        \multirow{2}{*}{%
            \parbox{0.1\linewidth}{%
                \vspace{-28pt}%
                \centering
                \footnotesize
                \textit{A mannequin dressed in a red gown with yellow accents stands amidst an array of vibrant flowers and lush greenery.}%
                    }%
        } &
        \adjincludegraphics[clip,width=0.1\linewidth,trim={0 0 0 0}]{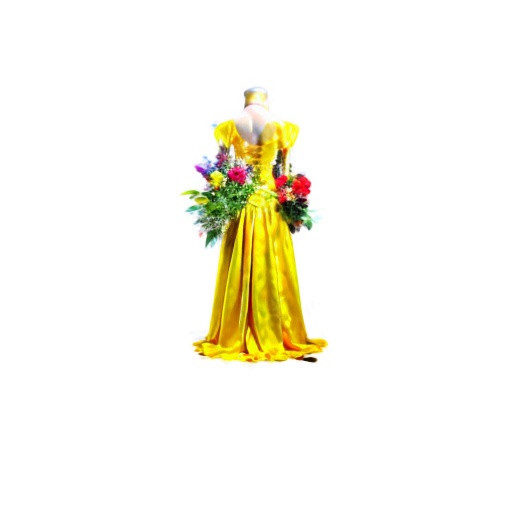} &
        \adjincludegraphics[clip,width=0.1\linewidth,trim={0 0 0 0}]{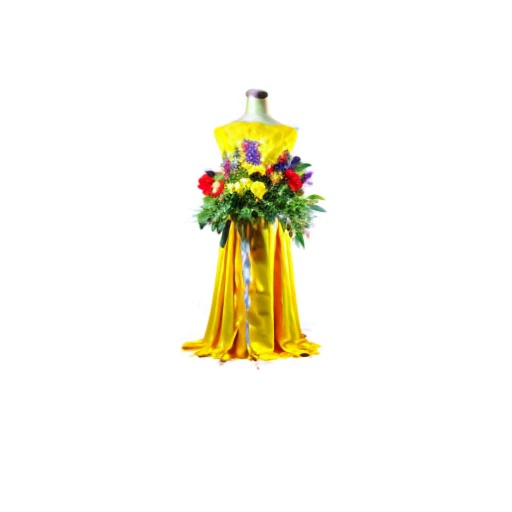} &
        \adjincludegraphics[clip,width=0.1\linewidth,trim={0 0 0 0}]{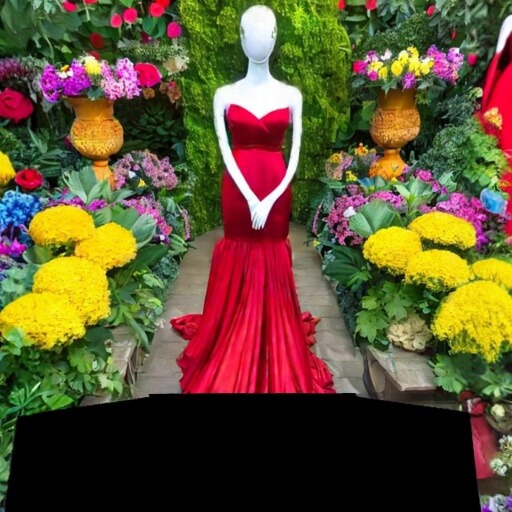} &
        \adjincludegraphics[clip,width=0.1\linewidth,trim={0 0 0 0}]{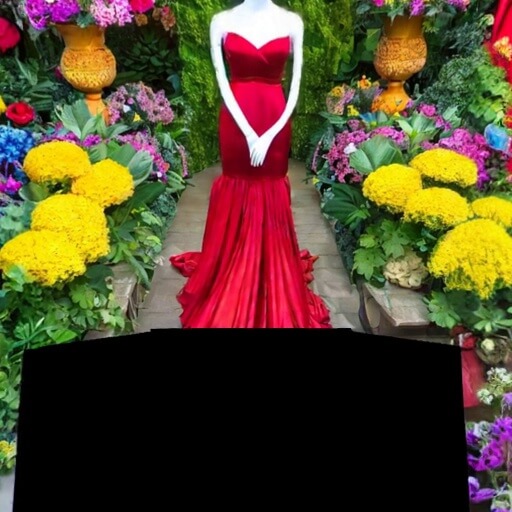} &
        \adjincludegraphics[clip,width=0.1\linewidth,trim={0 0 0 0}]{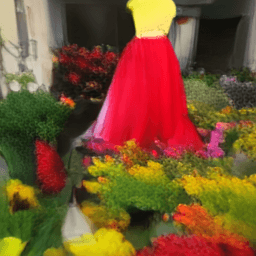} &
        \adjincludegraphics[clip,width=0.1\linewidth,trim={0 0 0 0}]{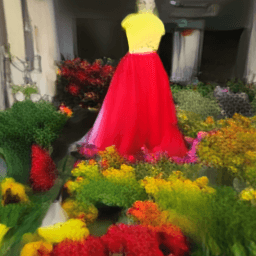} &
        \adjincludegraphics[clip,width=0.1\linewidth,trim={0 0 0 0}]{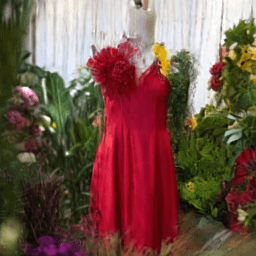} &
        \adjincludegraphics[clip,width=0.1\linewidth,trim={0 0 0 0}]{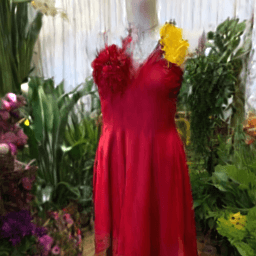} \\
         &
        \adjincludegraphics[clip,width=0.1\linewidth,trim={0 0 0 0}]{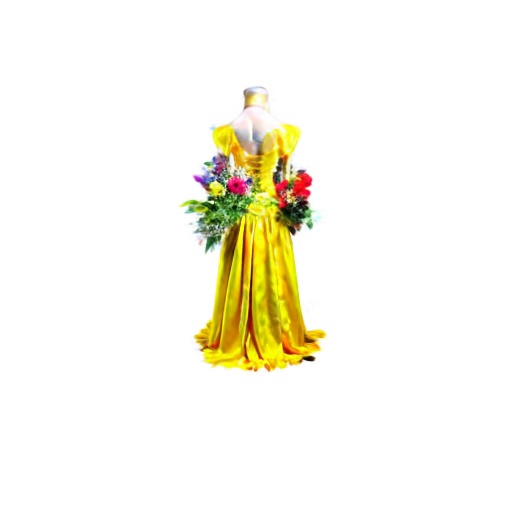} &
        \adjincludegraphics[clip,width=0.1\linewidth,trim={0 0 0 0}]{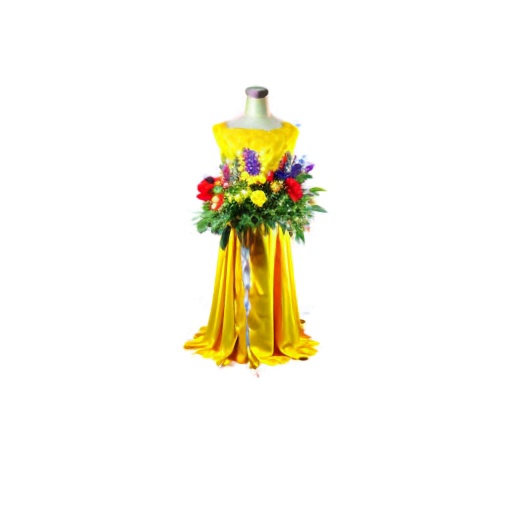} &
        \adjincludegraphics[clip,width=0.1\linewidth,trim={0 0 0 0}]{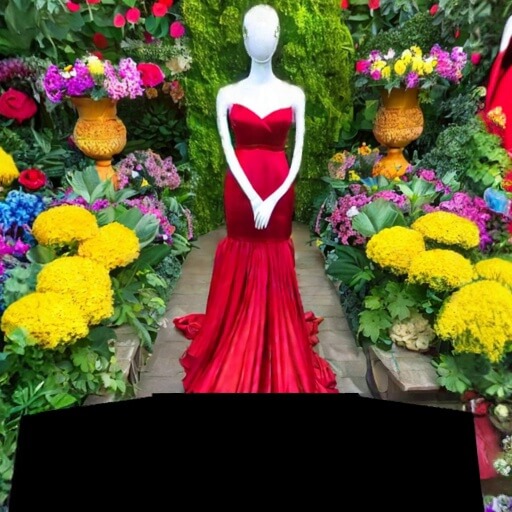} &
        \adjincludegraphics[clip,width=0.1\linewidth,trim={0 0 0 0}]{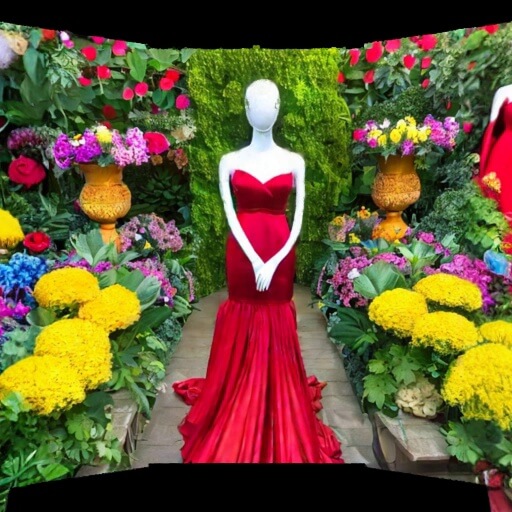} &
        \adjincludegraphics[clip,width=0.1\linewidth,trim={0 0 0 0}]{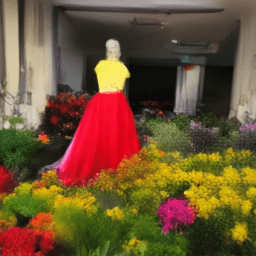} &
        \adjincludegraphics[clip,width=0.1\linewidth,trim={0 0 0 0}]{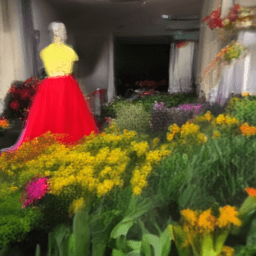} &
        \adjincludegraphics[clip,width=0.1\linewidth,trim={0 0 0 0}]{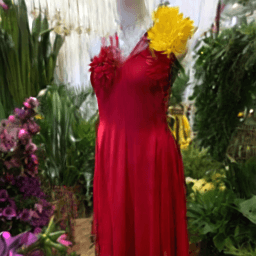} &
        \adjincludegraphics[clip,width=0.1\linewidth,trim={0 0 0 0}]{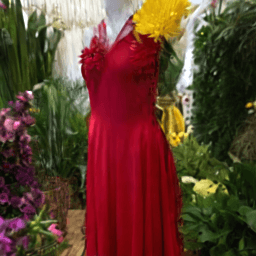} \\ 
        \midrule
        \multirow{2}{*}{%
            \parbox{0.1\linewidth}{%
                \vspace{-28pt}%
                \centering
                \footnotesize
                \textit{An overgrown and neglected area with a large bush or small tree covered in ivy, a wooden fence, and a paved path.}%
                    }%
        } &
        \adjincludegraphics[clip,width=0.1\linewidth,trim={0 0 0 0}]{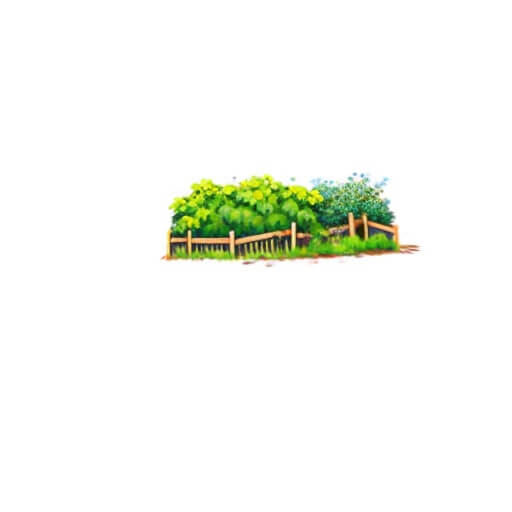} &
        \adjincludegraphics[clip,width=0.1\linewidth,trim={0 0 0 0}]{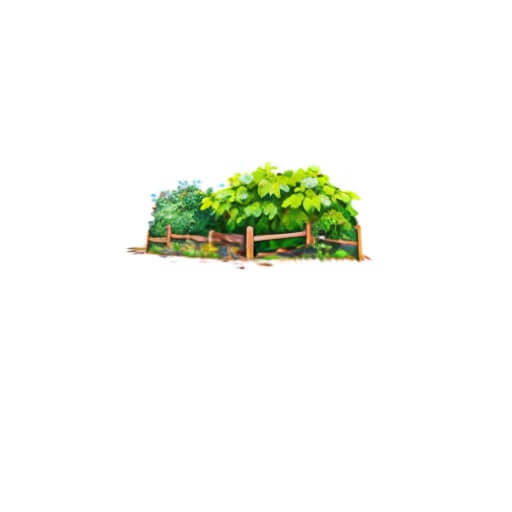} &
        \adjincludegraphics[clip,width=0.1\linewidth,trim={0 0 0 0}]{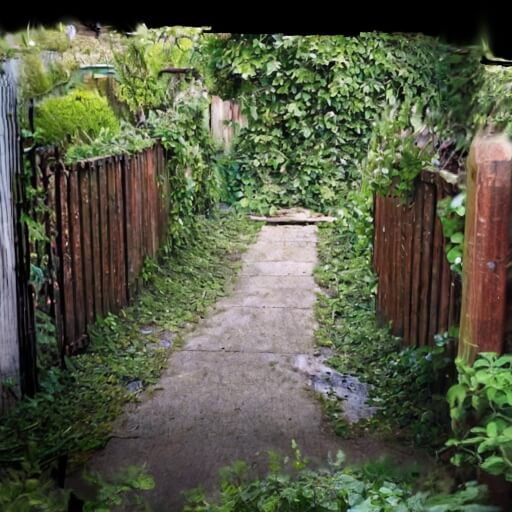} &
        \adjincludegraphics[clip,width=0.1\linewidth,trim={0 0 0 0}]{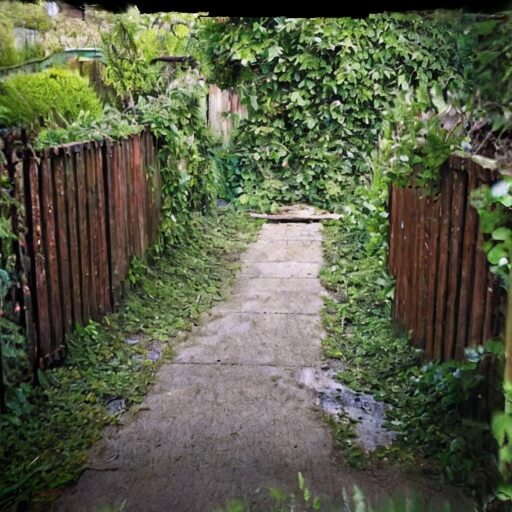} &
        \adjincludegraphics[clip,width=0.1\linewidth,trim={0 0 0 0}]{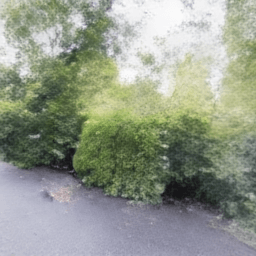} &
        \adjincludegraphics[clip,width=0.1\linewidth,trim={0 0 0 0}]{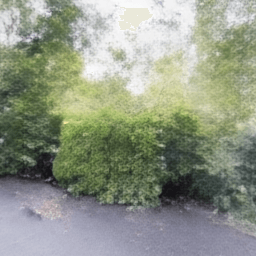} &
        \adjincludegraphics[clip,width=0.1\linewidth,trim={0 0 0 0}]{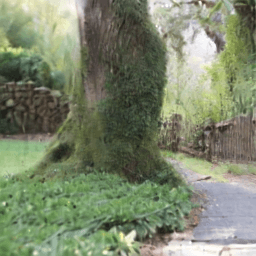} &
        \adjincludegraphics[clip,width=0.1\linewidth,trim={0 0 0 0}]{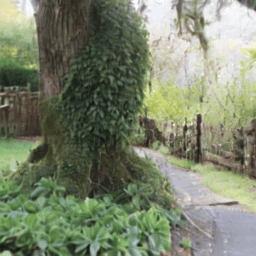} \\
         &
        \adjincludegraphics[clip,width=0.1\linewidth,trim={0 0 0 0}]{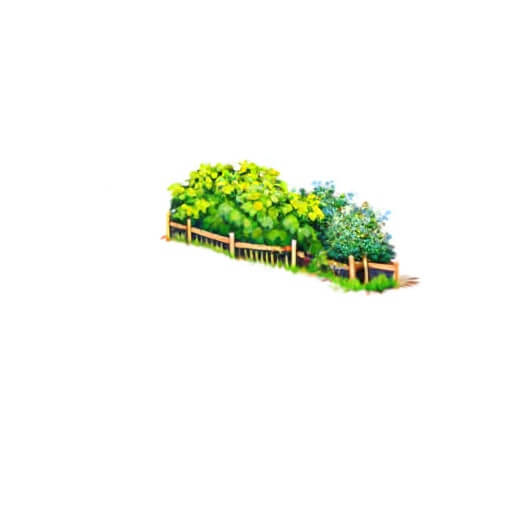} &
        \adjincludegraphics[clip,width=0.1\linewidth,trim={0 0 0 0}]{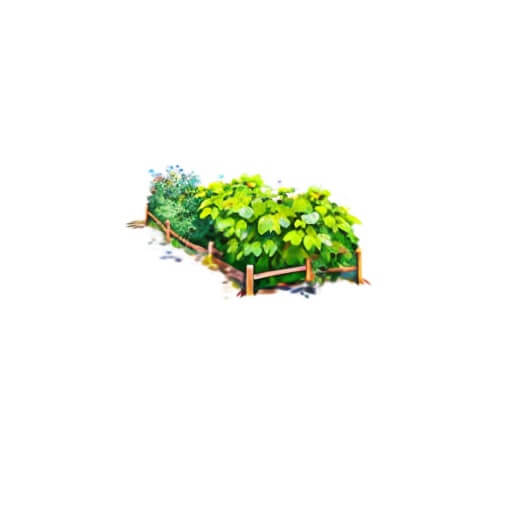} &
        \adjincludegraphics[clip,width=0.1\linewidth,trim={0 0 0 0}]{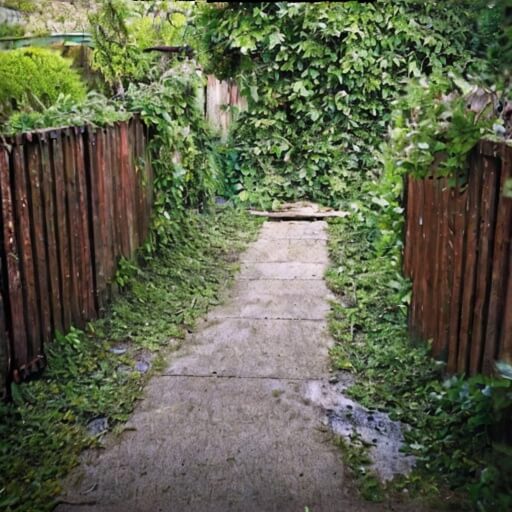} &
        \adjincludegraphics[clip,width=0.1\linewidth,trim={0 0 0 0}]{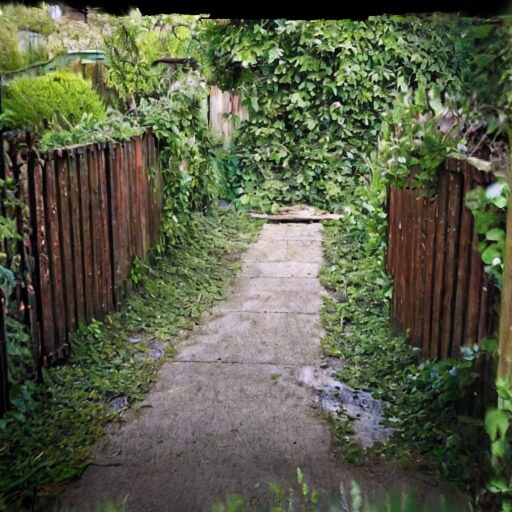} &
        \adjincludegraphics[clip,width=0.1\linewidth,trim={0 0 0 0}]{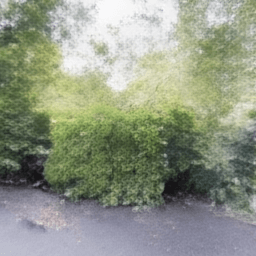} &
        \adjincludegraphics[clip,width=0.1\linewidth,trim={0 0 0 0}]{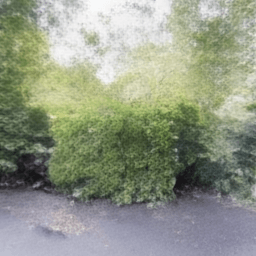} &
        \adjincludegraphics[clip,width=0.1\linewidth,trim={0 0 0 0}]{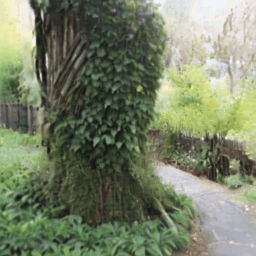} &
        \adjincludegraphics[clip,width=0.1\linewidth,trim={0 0 0 0}]{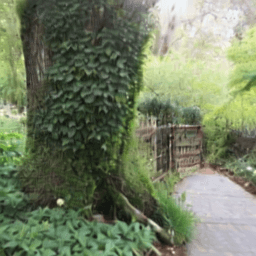} \\
        \multicolumn{1}{c}{\small Text} & \multicolumn{2}{c}{\small DreamScene~\cite{li2024dreamscene}} & \multicolumn{2}{c}{\small LucidDreamer~\cite{chung2023luciddreamer}} & \multicolumn{2}{c}{\small Director3D~\cite{li2024director3d}} & \multicolumn{2}{c}{\small Ours}
    \end{tabular}
    \vspace{\abovefigcapmargin}
    \caption{\textbf{Qualitative results in text-to-3DGS generation on MVImgNet and DL3DV validation sets.} The first two rows are rendered scenes from the MVImgNet dataset, while the last two rows are from the DL3DV dataset. Our SplatFlow produces cohesive and realistic scenes with sharp details, accurately capturing the intricacies of real-world environments and accommodating diverse camera trajectories.}
    \vspace{\belowfigcapmargin}
    \vspace{-1mm}
    \label{fig:qual_scene_generation}
\end{figure*}

\subsection{Experimental Setups}
\label{sec:exp:implenetation}
Due to space constraints, we provide a concise overview of our experimental setups here. We used the MVImgNet~\cite{yu2023mvimgnet} and DL3DV-7K~\cite{ling2024dl3dv} datasets, both containing real-world scenes with images and corresponding camera poses. For evaluation, we utilized 1.25K scenes from MVImgNet for each task and 300 sequences from DL3DV-7K, using the remaining for training. The multi-view RF model and GSDecoder were trained with $K = 8$-view setup. Text descriptions for scenes were extracted using the Llava-One Vision Qwen 7B~\cite{li2024llava}. We used SDS++~\cite{li2024director3d} for further improvements.

\subsection{Results on Text-to-3DGS Generation}
\label{sec:exp:3dgs_gen}
We evaluated our model against various baselines on the DL3DV and MVImgNet validation sets for real-world scene text-to-3DGS generation. Our evaluation protocol utilized the rendered image of the generated 3DGS. We used the FID score~\cite{heusel2017gans} to evaluate image quality and the CLIP score~\cite{hessel2021clipscore} to measure alignment with text prompts.

For quantitative comparison, we adopt Director3D~\cite{li2024director3d}, which leverages the full sequences of MVImgNet, DL3DV-10K, and LAION~\cite{schuhmann2022laion} datasets. In contrast, our model was trained on a notably smaller dataset, utilizing only a portion of MVImgNet and DL3DV-7K, without access to the full DL3DV-10K or LAION data. For qualitative results, we also compared against other scene generation methods, including LucidDreamer~\cite{chung2023luciddreamer}, DreamScene~\cite{li2024dreamscene}, and Director3D~\cite{li2024director3d}.

\begin{table}[t!]
    \centering
    \setlength\tabcolsep{2.5pt}
    \resizebox{\linewidth}{!}{
    \begin{tabular}{lcccc}
       \toprule
       \multirow{2}{*}{Method} & \multicolumn{2}{c}{MVImgNet~\cite{yu2023mvimgnet}} & \multicolumn{2}{c}{DL3DV~\cite{ling2024dl3dv}} \\
       \arrayrulecolor{gray}\cmidrule(lr){2-3}\cmidrule(lr){4-5} 
                               & FID-10K$\downarrow$ & CLIPScore$\uparrow$  & FID-2.4K$\downarrow$ & CLIPScore$\uparrow$  \\
       \midrule
       Director3D~\cite{li2024director3d} & 39.55 & 30.48 & 88.44 & 30.04     \\ 
       Director3D (w/ SDS++)~\cite{li2024director3d} & 41.80 & 31.00 & 95.88 & 31.68    \\
       \midrule
      \rowcolor{gray!25} \textbf{SplatFlow} & \textbf{34.85} & 31.43 & \textbf{79.91} & 30.06  \\ 
       \rowcolor{gray!25} \textbf{SplatFlow (w/ SDS++)} &  35.46 & \textbf{32.30} & 85.31 & \textbf{31.90} \\
       \arrayrulecolor{black}\bottomrule
    \end{tabular}
    }
    
    \vspace{\abovetabcapmargin}
    \caption{\textbf{Quantitative results in text-to-3DGS generation on the MVImgNet and DL3DV datasets.} We compared our SplatFlow with and without the SDS++~\cite{li2024director3d}, against Director3D~\cite{li2024director3d}.}
    \vspace{\belowtabcapmargin}
    \label{tab:3dgs_quant}
\end{table}

\vspace{\paramargin}
\paragraph{Quantitative results}
Table~\ref{tab:3dgs_quant} presents the quantitative comparison of SplatFlow and the baseline methods on both MVImgNet and DL3DV datasets. SplatFlow consistently outperforms Director3D~\cite{li2024director3d} in terms of FID and CLIP score, demonstrating its effectiveness in generating high-quality images and maintaining strong alignment with text prompts, even with a smaller training dataset. Specifically, on MVImgNet, SplatFlow achieves a significantly lower FID (34.85 vs. 39.55 for Director3D), indicating better image quality, and a higher CLIP score (31.43 vs. 30.48), showing improved text-to-image alignment. Similarly, on DL3DV, SplatFlow achieves an FID of 79.91, outperforming Director3D’s 88.44, with a comparable CLIP score.
Notably, with the addition of SDS++ for refinement, both SplatFlow and Director3D show increased CLIP scores, further enhancing text alignment, though at a minor trade-off in FID. This highlights the effectiveness of SDS++ for boosting alignment with text prompts across different 3DGS generation models.

\vspace{\paramargin}
\paragraph{Qualitative results}
Figure~\ref{fig:qual_scene_generation} shows the qualitative comparisons between our method and baselines on DL3DV and MVImgNet validation sets. 
Overall, SplatFlow demonstrates clear advantages over the baseline methods. First, DreamScene~\cite{li2024dreamscene}, which relies on Score Distillation Sampling (SDS), often struggles to generate realistic real-world scenes, resulting in outputs that lack natural textures and environmental details. Second, LucidDreamer~\cite{chung2023luciddreamer}, which is based on single-view inpainting, faces challenges when handling scenes with large camera trajectory variations, leading to inconsistent or incomplete views. Third, Director3D~\cite{li2024director3d} tends to produce slightly blurred outputs, lacking the sharpness needed for intricate details.
In contrast, our SplatFlow can generate coherent and realistic scenes with clear details, effectively capturing the complexities of real-world environments and various camera trajectories.
Additional qualitative results of SplatFlow can be found in Appendix~\textcolor{cvprblue}{C}.

\setlength{\columnsep}{10pt}%
\begin{wraptable}[6]{R}{0.5\linewidth}
    \centering
    \vspace{-4mm}
    \setlength\tabcolsep{1.2pt}
    \resizebox{\linewidth}{!}{%
    \begin{tabular}{lccc}
       \toprule
       Method & CLIPScore$\uparrow$ & CLIP D-sim$\uparrow$ \\
       \arrayrulecolor{gray}\midrule
       DGE~\cite{chen2024dge} & 27.43 & 0.102\\
       \midrule
       SplatFlow & 28.47 & 0.169 \\
       \rowcolor{gray!25} \; \textbf{+) SDS++} & \textbf{31.30} & \textbf{0.224}\\
       \arrayrulecolor{black}\bottomrule
    \end{tabular}
    }
    \vspace{\abovetabcapmargin}
    \caption{\textbf{3D object replacement.}}
    \label{tab:edit}
\end{wraptable}


\subsection{Result on 3DGS Editing}
\label{sec:exp:sdgs_editing}
\begin{figure}[t!]
    \centering
    \includegraphics[width=1.0\linewidth]{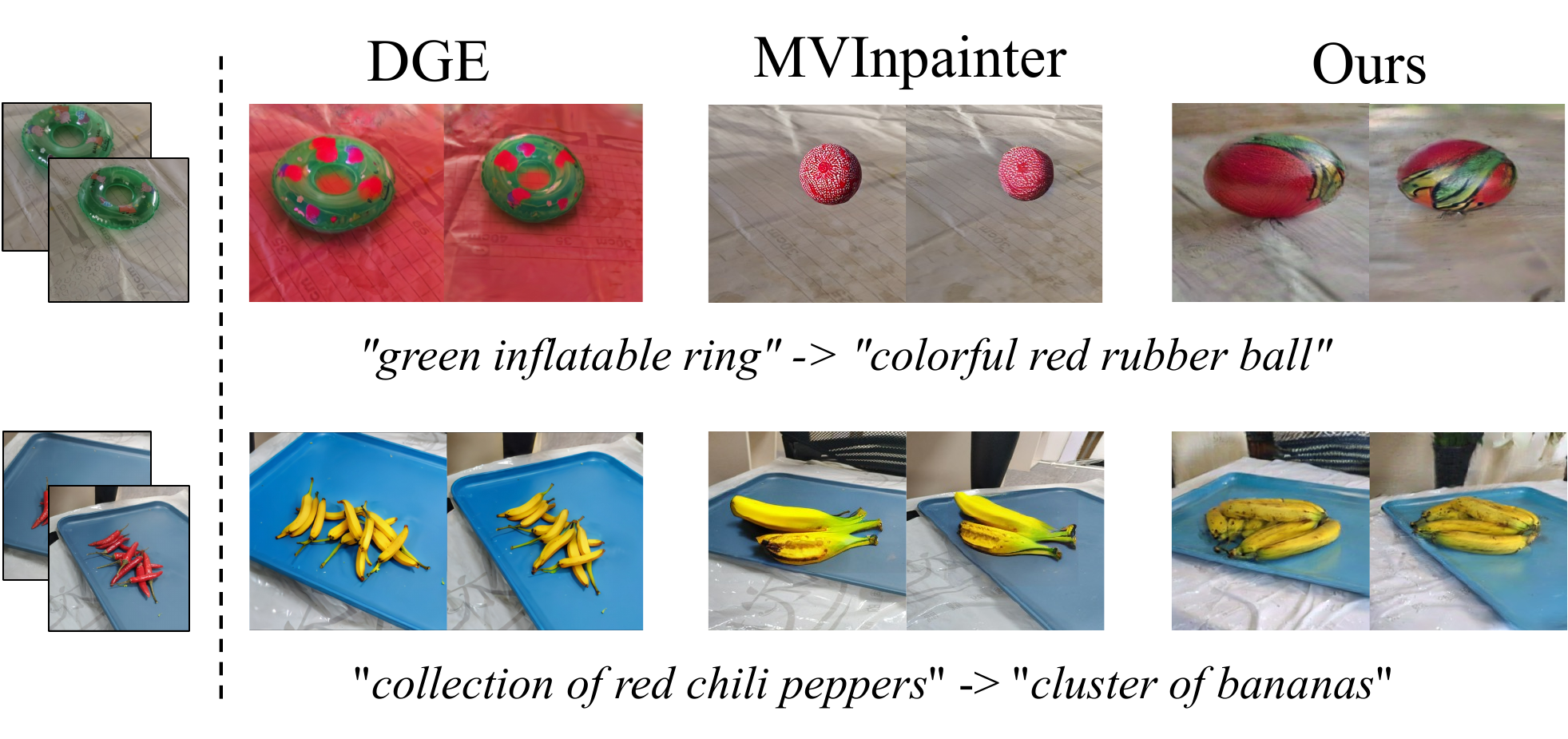}

    \vspace{-1.5mm}
    \vspace{\abovefigcapmargin}
    \caption{\textbf{Qualitative results in 3D editing with MVInpainter~\cite{cao2024mvinpainter} and DGE~\cite{chen2024dge}.} We show rendered scenes except for MVInpainter.}
    \vspace{\belowfigcapmargin}
    \vspace{1.5mm}
    \label{fig:edit}
\end{figure}

We demonstrate the effectiveness of SplatFlow in the 3DGS editing task by creating a benchmark with 100 sampled scenes from the MVImgNet validation set. Images and texts were provided to GPT-4 to identify main objects and generate text captions for edited images, which were then used for 3D object replacement.
For editing, we used the same mask in~\cite{cao2024mvinpainter} for masking out the object, then inpainted this area based on the provided editing captions.
We compared this pipeline with an optimization-based 3DGS editing method, DGE~\cite{chen2024dge}, and conducted qualitative comparisons with MVInpainter~\cite{cao2024mvinpainter}, which propagates an edited image from pretrained generative models across multiple views. Following standard protocols~\cite{kamata2023instruct, chen2024gaussianeditor}, we measured the CLIP score and directional similarity (CLIP D-sim).

\vspace{\paramargin} \paragraph{Quantitative results} As shown in Table~\ref{tab:edit}, SplatFlow achieves higher CLIPScore and CLIP D-sim than DGE~\cite{chen2024dge}, indicating more accurate and effective editing. SDS++ refinement further boosts performance, demonstrating enhanced alignment with target edits.

\vspace{\paramargin} \paragraph{Qualitative results} Figure~\ref{fig:edit} demonstrates that SplatFlow performs comparably to MVInpainter, a specialized 2D inpainting method that leverages an already edited high-quality image. On the other hand, DGE struggles to replace objects fully, often just altering styles instead of achieving complete transformations. This underscores SplatFlow's superior ability in precise 3D object replacement. Please refer to Appendix~\textcolor{cvprblue}{C}
for further qualitative results.

\subsection{Result on Inpainting Application}
\label{sec:exp:inpainting_result}
\begin{table}[t!]
    \centering
    \setlength\tabcolsep{4.5pt}
    \resizebox{\linewidth}{!}{
    \begin{tabular}{lcccccc}
       \toprule
       \multirow{2}{*}{Method} & \multicolumn{3}{c}{Rotation$\uparrow$} & \multicolumn{3}{c}{Camera Center$\uparrow$} \\
       \arrayrulecolor{gray}\cmidrule(lr){2-4} \cmidrule(lr){5-7}
                               & @5 & @10 & @15 & @0.05 & @0.1 & @0.2 \\
       \midrule
       RelPose++~\cite{lin2023relpose++} & 19.4 & 37.7 & 51.4 & 0.6 & 12.5 & 55.0    \\
       Ray Regression~\cite{zhang2024cameras} & 10.4 & 25.6 & 50.1 & 15.3 & 47.9 & 82.9    \\
       Ray Diffusion~\cite{zhang2024cameras} & 17.5 & 38.7 & 59.6 & 24.1 & 60.9 & 87.6    \\
       \midrule
       \rowcolor{gray!25} \textbf{SplatFlow (w/ depth)} & 26.8 & 52.6 & 59.3  & 62.3 & 91.6 & 99.4 \\
      \rowcolor{gray!25} \textbf{SplatFlow (w/o depth)} & \textbf{28.8} & \textbf{54.5} & \textbf{63.9} & \textbf{64.9} & \textbf{94.0} & \textbf{99.7}  \\
       
       \arrayrulecolor{black}\bottomrule
    \end{tabular}
    }
    \vspace{-1mm}
    \vspace{\abovetabcapmargin}
    \caption{\textbf{Results in camera pose estimation on MVImgNet validation set.} @$Q$ represents the accuracy threshold for rotations (degrees) and camera centers (units). }
    \vspace{-2mm}
    \label{tab:repaint_pose}
\end{table}

\begin{figure}[t!]
    \centering
    \setlength\tabcolsep{5pt}
    \begin{tabular}{cc}

        \adjincludegraphics[clip,width=0.6\linewidth,trim={0 0 0 0}]{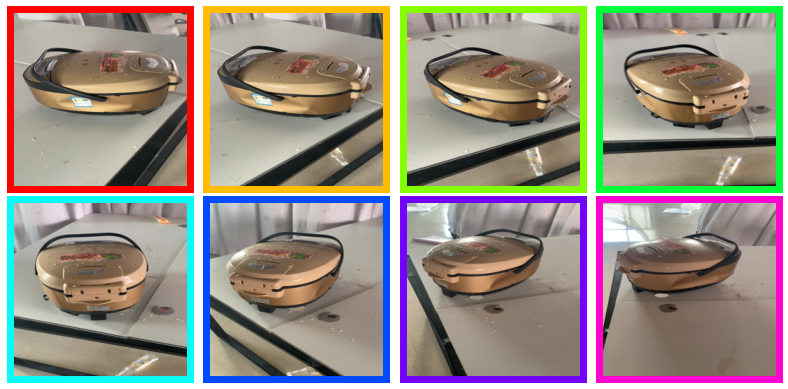} &
        \adjincludegraphics[clip,width=0.31\linewidth,trim={0 0 0 0}]{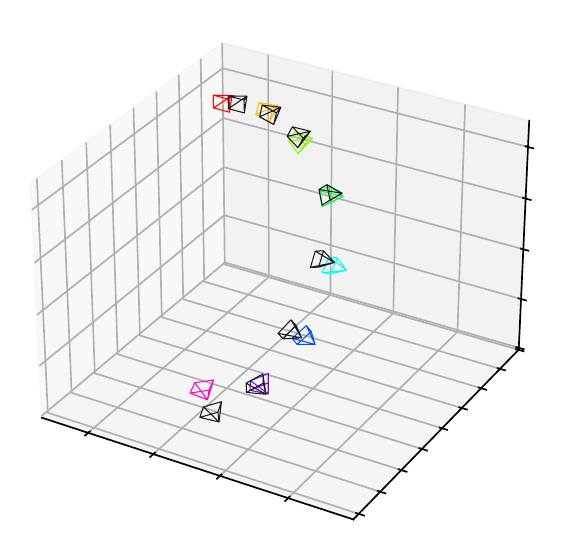} \\

        \adjincludegraphics[clip,width=0.6\linewidth,trim={0 0 0 0}]{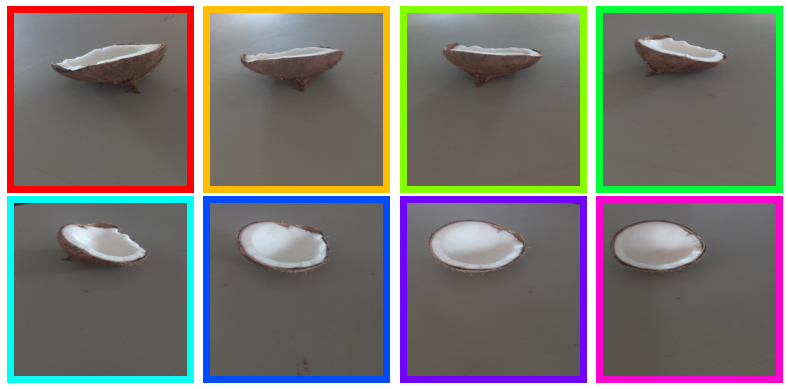} &
        \adjincludegraphics[clip,width=0.31\linewidth,trim={0 0 0 0}]{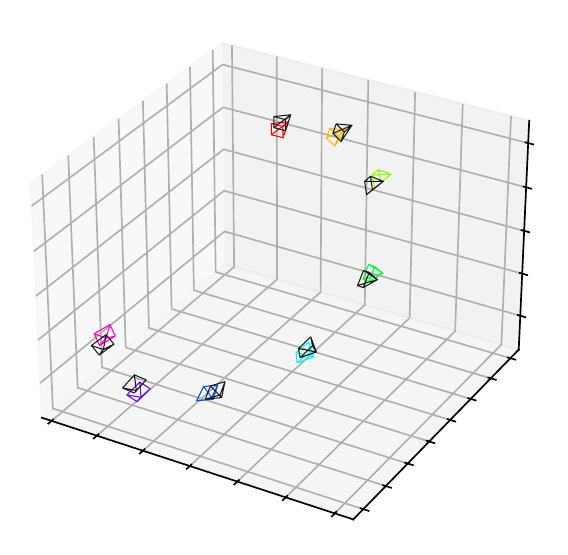} \\
        Input Images & Camera Poses \\
    \end{tabular}
    \vspace{-0.1cm}
    \vspace{\abovefigcapmargin}
    \caption{\textbf{Qualitative results for camera pose estimation.} Camera poses are estimated from multi-view images. Image border colors match each camera, with black cameras indicating GT poses.}
    \vspace{\belowfigcapmargin}

    \label{fig:repaint_pose}
\end{figure}

Here, we present the adaptability of our multi-view RF model to camera pose estimation and novel-view synthesis via the inpainting technique on the MVImgNet dataset~\cite{yu2023mvimgnet}.

\vspace{\paramargin}
\paragraph{Camera pose estimation }
To demonstrate the effectiveness of our multi-view RF model in camera pose estimation, we primarily compare our approach to RayDiffusion~\cite{zhang2024cameras} and RelPose++~\cite{lin2023relpose++}. We conducted comparisons using an 8-view setup, inpainting the ray latent with and without the depth latent from DepthAnythingV2~\cite{yang2024depth}.
We used rotation accuracy@$Q_{R}$ and translation accuracy@$Q_{T}$ as evaluation metrics, measuring the proportion of relative camera rotations within $Q_{R}$ degrees of ground truth and camera centers within $Q_{T}$ units of scene scale, respectively.

Table~\ref{tab:repaint_pose} and Fig.~\ref{fig:repaint_pose} present the quantitative and qualitative results. Overall, the multi-view RF model achieves superior performance across most scenarios, outperforming the baseline methods. Interestingly, the multi-view RF model achieves higher accuracy without depth latents than with them. We hypothesize that this outcome is due to the joint generation of depth and camera poses, which may yield more detailed and contextually enriched depth estimates.

\vspace{\paramargin}
\paragraph{Novel view-synthesis}
We evaluated the performance of our multi-view RF model in novel view synthesis under two scenarios: 1) interpolation, with the uniformly sampled input views, and 2) extrapolation, with the input views positioned in the center. This comparison validates our multi-view RF model's capability of 3D reasoning based on relative viewpoints and evaluates its effectiveness in generating novel views. We evaluated PSNR, SSIM, and LPIPS~\cite{zhang2018unreasonable} against the ground-truth images for reconstruction quality, as well as absolute relative difference and $\delta_1$ accuracy using depth maps produced by DepthAnythingV2~\cite{yang2024depth}.

\begin{table}[t!]
    \centering
    \setlength\tabcolsep{4pt}
    \resizebox{\linewidth}{!}{
    \begin{tabular}{lccccc}
       \toprule
       \multirow{2}{*}{Type} & \multicolumn{3}{c}{RGB} & \multicolumn{2}{c}{Depth} \\
       \arrayrulecolor{gray}\cmidrule(lr){2-4} \cmidrule(lr){5-6}
                       & PSNR$\uparrow$ & SSIM$\uparrow$ & LPIPS$\downarrow$ & AbsRel$\downarrow$ & $\delta_{1}$$\uparrow$ \\
        \midrule
        Interpolation (N=2) & 14.73 & 0.571 & 0.648 & 0.588  & 0.731  \\
        Interpolation (N=4) & 17.05 & 0.590 & 0.551 & 0.498  & 0.761 \\
        Interpolation (N=6) & 18.82 & 0.626 & 0.483 & 0.415  & 0.775 \\
        
        \arrayrulecolor{gray}\midrule
        
        Extrapolation (N=2) & 15.15 & 0.577 & 0.627 & 0.771 & 0.715  \\
        Extrapolation (N=4) & 16.80 & 0.595 & 0.554 & 0.679  & 0.727 \\
        Extrapolation (N=6) & 17.96 & 0.613 & 0.503 & 0.602 & 0.747  \\
       \arrayrulecolor{black}\bottomrule
    \end{tabular}
    }
    \vspace{-0.1cm}
    \vspace{\abovetabcapmargin}
    \caption{\textbf{Novel view synthesis results on MVImgNet.} We use $N$ input views to synthesize $K - N$ novel views, with uniformly sampled views for interpolation and central views for extrapolation.}
    \vspace{-2mm}
    \label{tab:repaint_scene}
\end{table}

\begin{figure}[t!]
    \centering
    \setlength\tabcolsep{1.5pt}
    \begin{tabular}{cccccccc}
        \toprule

        \adjincludegraphics[clip,width=0.11\linewidth,trim={0 0 0 0},cfbox=red 0.2pt 0.2pt]{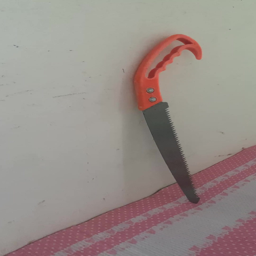} &
        \adjincludegraphics[clip,width=0.11\linewidth,trim={0 0 0 0}]{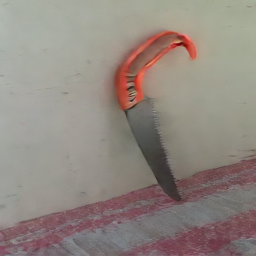} &
        \adjincludegraphics[clip,width=0.11\linewidth,trim={0 0 0 0},cfbox=red 0.2pt 0.2pt]{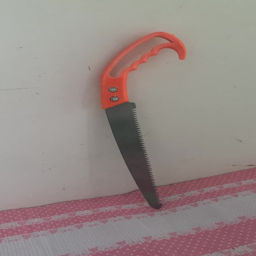} &
        \adjincludegraphics[clip,width=0.11\linewidth,trim={0 0 0 0}]{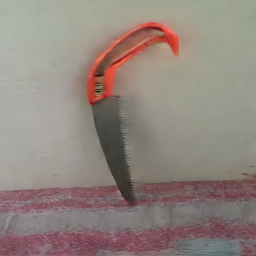} &
        \adjincludegraphics[clip,width=0.11\linewidth,trim={0 0 0 0},cfbox=red 0.2pt 0.2pt]{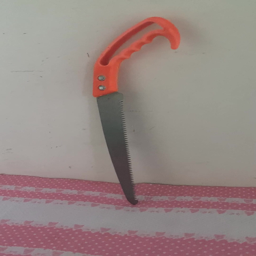} &
        \adjincludegraphics[clip,width=0.11\linewidth,trim={0 0 0 0}]{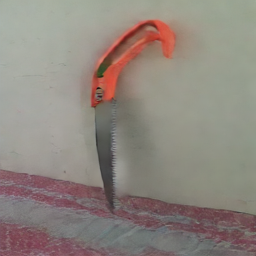} &
        \adjincludegraphics[clip,width=0.11\linewidth,trim={0 0 0 0}]{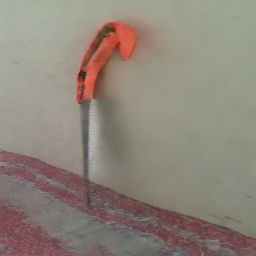} &
        \adjincludegraphics[clip,width=0.11\linewidth,trim={0 0 0 0},cfbox=red 0.2pt 0.2pt]{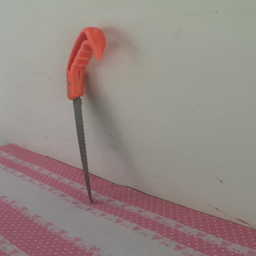} \\

        \adjincludegraphics[clip,width=0.11\linewidth,trim={0 0 0 0},cfbox=red 0.2pt 0.2pt]{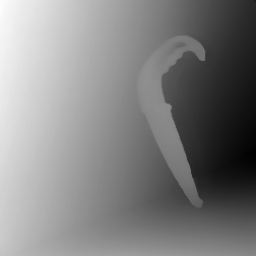} &
        \adjincludegraphics[clip,width=0.11\linewidth,trim={0 0 0 0}]{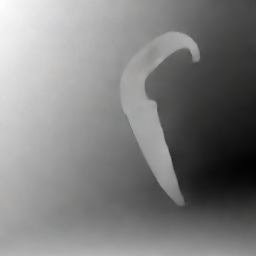} &
        \adjincludegraphics[clip,width=0.11\linewidth,trim={0 0 0 0},cfbox=red 0.2pt 0.2pt]{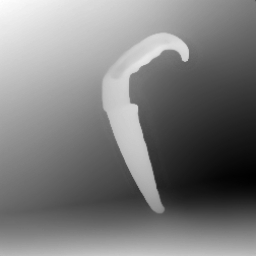} &
        \adjincludegraphics[clip,width=0.11\linewidth,trim={0 0 0 0}]{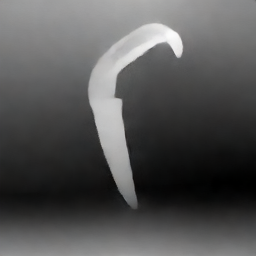} &
        \adjincludegraphics[clip,width=0.11\linewidth,trim={0 0 0 0},cfbox=red 0.2pt 0.2pt]{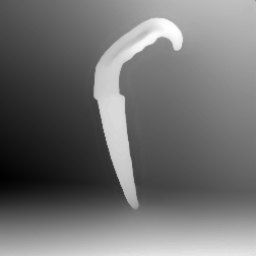} &
        \adjincludegraphics[clip,width=0.11\linewidth,trim={0 0 0 0}]{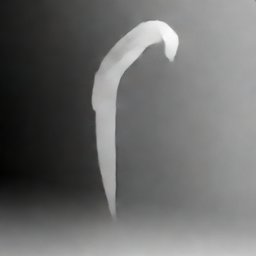} &
        \adjincludegraphics[clip,width=0.11\linewidth,trim={0 0 0 0}]{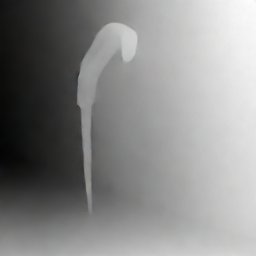} &
        \adjincludegraphics[clip,width=0.11\linewidth,trim={0 0 0 0},cfbox=red 0.2pt 0.2pt]{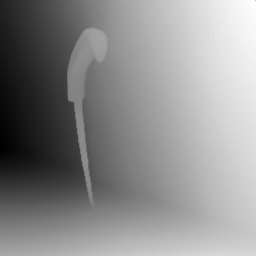} \\
        
        \adjincludegraphics[clip,width=0.11\linewidth,trim={0 0 0 0},cfbox=red 0.2pt 0.2pt]{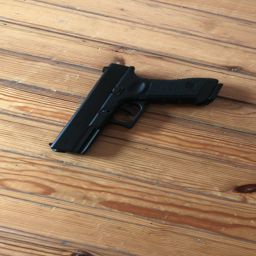} &
        \adjincludegraphics[clip,width=0.11\linewidth,trim={0 0 0 0}]{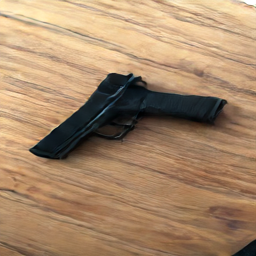} &
        \adjincludegraphics[clip,width=0.11\linewidth,trim={0 0 0 0}]{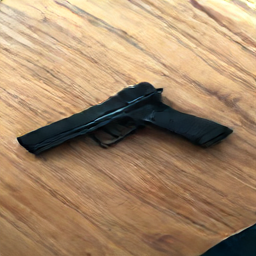} &
        \adjincludegraphics[clip,width=0.11\linewidth,trim={0 0 0 0}]{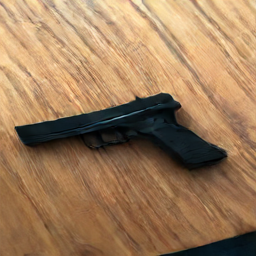} &
        \adjincludegraphics[clip,width=0.11\linewidth,trim={0 0 0 0}]{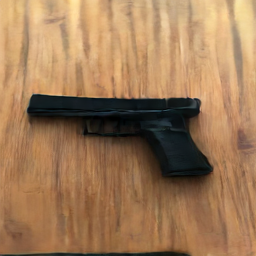} &
        \adjincludegraphics[clip,width=0.11\linewidth,trim={0 0 0 0}]{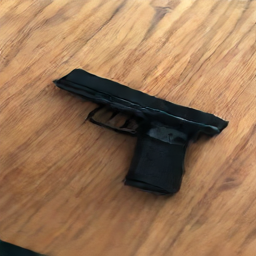} &
        \adjincludegraphics[clip,width=0.11\linewidth,trim={0 0 0 0}]{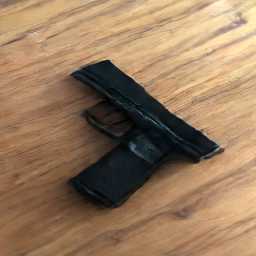} &
        \adjincludegraphics[clip,width=0.11\linewidth,trim={0 0 0 0},cfbox=red 0.2pt 0.2pt]{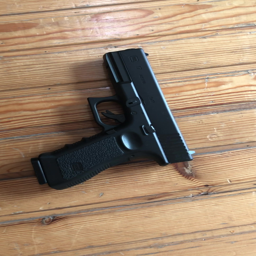} \\

        \adjincludegraphics[clip,width=0.11\linewidth,trim={0 0 0 0},cfbox=red 0.2pt 0.2pt]{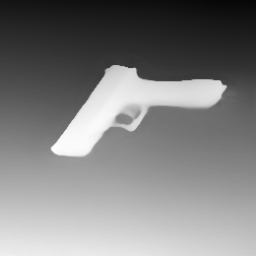} &
        \adjincludegraphics[clip,width=0.11\linewidth,trim={0 0 0 0}]{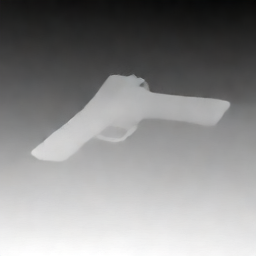} &
        \adjincludegraphics[clip,width=0.11\linewidth,trim={0 0 0 0}]{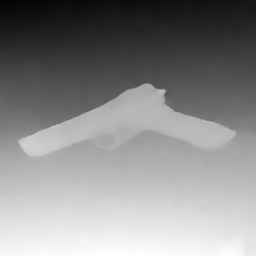} &
        \adjincludegraphics[clip,width=0.11\linewidth,trim={0 0 0 0}]{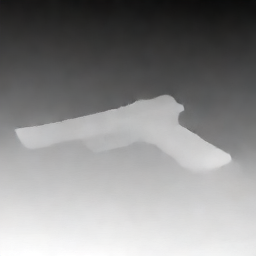} &
        \adjincludegraphics[clip,width=0.11\linewidth,trim={0 0 0 0}]{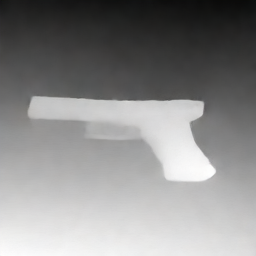} &
        \adjincludegraphics[clip,width=0.11\linewidth,trim={0 0 0 0}]{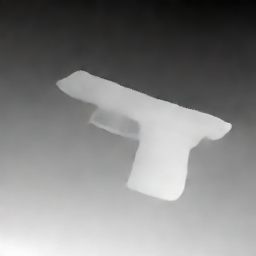} &
        \adjincludegraphics[clip,width=0.11\linewidth,trim={0 0 0 0}]{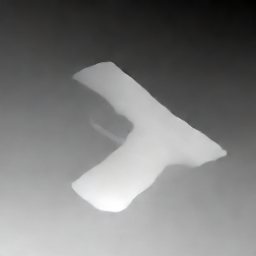} &
        \adjincludegraphics[clip,width=0.11\linewidth,trim={0 0 0 0},cfbox=red 0.2pt 0.2pt]{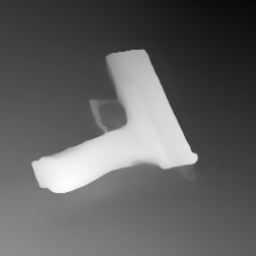} \\
        
        \arrayrulecolor{gray}\midrule

        \adjincludegraphics[clip,width=0.11\linewidth,trim={0 0 0 0}]{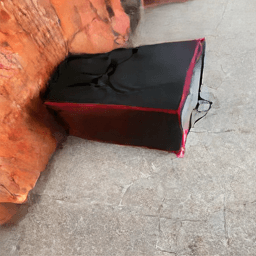} &
        \adjincludegraphics[clip,width=0.11\linewidth,trim={0 0 0 0}]{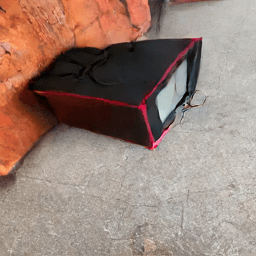} &
        \adjincludegraphics[clip,width=0.11\linewidth,trim={0 0 0 0}]{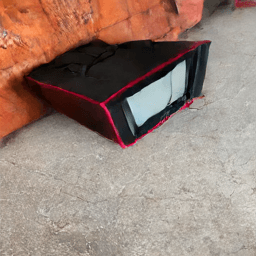} &
        \adjincludegraphics[clip,width=0.11\linewidth,trim={0 0 0 0},cfbox=red 0.2pt 0.2pt]{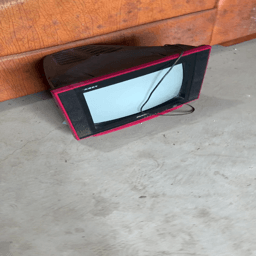} &
        \adjincludegraphics[clip,width=0.11\linewidth,trim={0 0 0 0},cfbox=red 0.2pt 0.2pt]{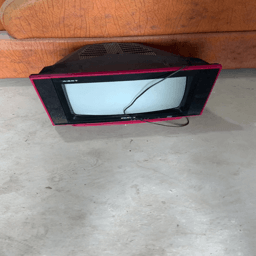} &
        \adjincludegraphics[clip,width=0.11\linewidth,trim={0 0 0 0}]{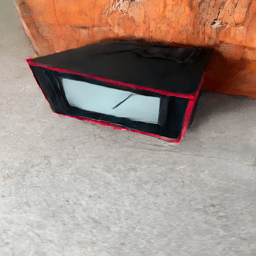} &
        \adjincludegraphics[clip,width=0.11\linewidth,trim={0 0 0 0}]{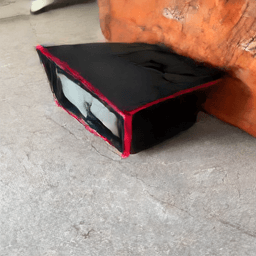} &
        \adjincludegraphics[clip,width=0.11\linewidth,trim={0 0 0 0}]{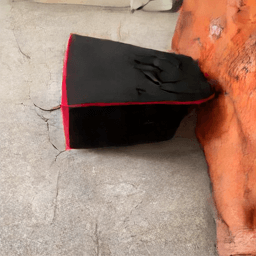} \\

        \adjincludegraphics[clip,width=0.11\linewidth,trim={0 0 0 0}]{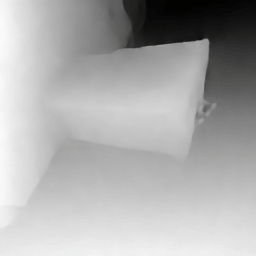} &
        \adjincludegraphics[clip,width=0.11\linewidth,trim={0 0 0 0}]{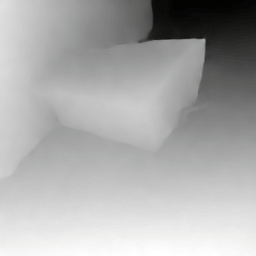} &
        \adjincludegraphics[clip,width=0.11\linewidth,trim={0 0 0 0}]{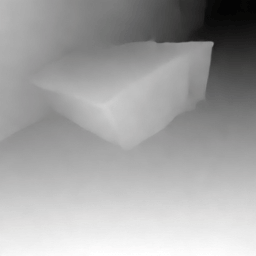} &
        \adjincludegraphics[clip,width=0.11\linewidth,trim={0 0 0 0},cfbox=red 0.2pt 0.2pt]{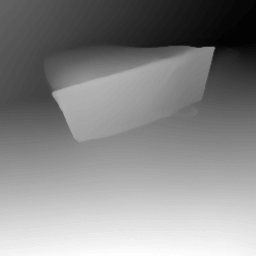} &
        \adjincludegraphics[clip,width=0.11\linewidth,trim={0 0 0 0},cfbox=red 0.2pt 0.2pt]{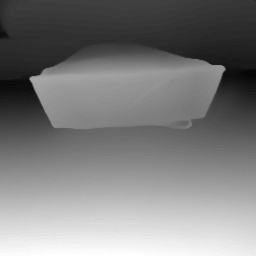} &
        \adjincludegraphics[clip,width=0.11\linewidth,trim={0 0 0 0}]{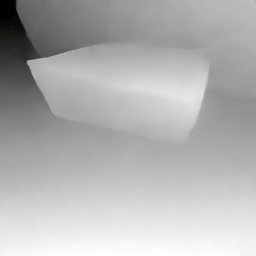} &
        \adjincludegraphics[clip,width=0.11\linewidth,trim={0 0 0 0}]{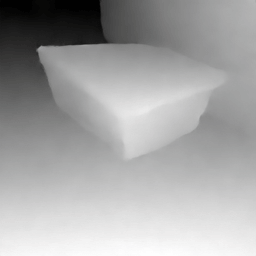} &
        \adjincludegraphics[clip,width=0.11\linewidth,trim={0 0 0 0}]{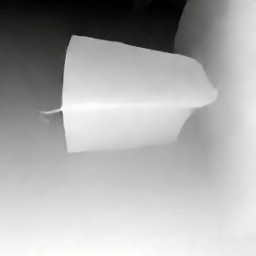} \\
        \arrayrulecolor{black}\bottomrule
    \end{tabular}
    \vspace{-0.1cm}
    \vspace{\abovefigcapmargin}
    \caption{\textbf{Qualitative results for novel view synthesis.} Novel view synthesis is performed from the red-box images and depths.}
    \vspace{\belowfigcapmargin}
    \vspace{-0.1cm}
    \label{fig:repaint_scene}
\end{figure}

As shown in Fig.~\ref{fig:repaint_scene}, our multi-view RF model successfully generates novel views for both scenarios. Also, Table~\ref{tab:repaint_scene} shows that interpolation outperforms extrapolation in both RGB reconstruction and depth estimation as the number of input views increases, indicating that interpolation leverages 3D reasoning for accurate novel view generation, whereas extrapolation focuses on generating diverse novel views.




\section{Conclusion}

In this paper, we have introduced \textit{SplatFlow}, a novel framework that unifies 3D Gaussian Splatting (3DGS) generation and editing within a single, efficient model. By designing a multi-view Rectified Flow (RF) model alongside a Gaussian Splatting Decoder (GSDecoder), we achieved direct 3DGS generation capable of addressing real-world challenges, such as varying scales and complex camera trajectories. Our approach utilizes joint modeling of multi-view images, depths, and camera poses, enabling seamless integration of training-free inversion and inpainting techniques for a diverse set of 3D tasks.
As we present the versatility of our SplatFlow, we believe SplatFlow represents an important step toward a versatile foundational model for 3D content.

{
    \small
    \bibliographystyle{ieeenat_fullname}
    \bibliography{main}
}

\clearpage
\maketitlesupplementary
\appendix

\section{Methodology Details}
\label{app:sec:method_details}

In this section, we provide additional details about the methodology of our proposed SplatFlow model, extending the description given in
Section~\textcolor{blue}{4}.
We begin by elaborating on the architecture and training details of the Gaussian Splatting Decoder (GSDecoder) in Section~\ref{app:sec:gs_decoder}, covering the specific modifications made to adapt the Stable Diffusion 3 decoder to 3D Gaussian Splatting (3DGS). Following that, we provide details on the Multi-View Rectified Flow (RF) model, including the architecture, loss functions, and several modifications made to the sampling process to jointly generate multi-view images, depths, and camera poses in Section~\ref{app:sec:mvrf}. Finally, we describe the editing process employed by SplatFlow, discussing how training-free inversion and inpainting techniques are applied to facilitate seamless 3D editing in Section~\ref{app:sec:editing}.

\subsection{Gaussian Splatting Decoder (GSDecoder)}
\label{app:sec:gs_decoder}

\paragraph{Architecture}
Our GSDecoder architecture builds upon the Stable Diffusion 3 decoder architecture~\cite{esser2024scaling}, with key modifications to adapt it for 3D Gaussian Splatting (3DGS). Specifically, we adjusted the input channel size to accommodate the concatenated latents of images, depths, and rays, and altered the output channel size to produce pixel-aligned 3DGS parameters. Furthermore, we modified the attention layers to incorporate cross-view attention, enabling the attention mechanism to operate across all view tokens simultaneously, rather than processing tokens for each view independently. 
We initialized the GSDecoder weights using the pre-trained weights from the Stable Diffusion decoder for all layers, except for the input and output layers. For these layers, we initialized the first channels with the corresponding Stable Diffusion weights, and the remaining channels were initialized by copying these values.

\vspace{\paramargin}
\paragraph{Loss function}
Our GSDecoder is trained using a weighted sum of three losses: mean squared error loss, LPIPS~\cite{zhang2018unreasonable}, and vision-aided GAN loss~\cite{kumari2022ensembling}. Specifically, the vision-aided GAN loss leverages backbones from DINO~\cite{caron2021emerging} and CLIP~\cite{radford2021learning}, and we incorporated differentiable augmentation~\cite{karras2020training} along with a multi-level version of hinge loss\footnote{\href{https://github.com/nupurkmr9/vision-aided-gan}{https://github.com/nupurkmr9/vision-aided-gan}}. 
Therefore, our loss can be represented as:
\vspace{-2mm}
\begin{equation}
    L_{\text{decoder}} = w_1 L_{\text{mse}} + w_2 L_{\text{LPIPS}} + w_3 L_{\text{vision-aided}},
\end{equation}
where $L_{\text{mse}}$, $L_{\text{LPIPS}}$, and $L_{\text{vision-aided}}$ represent the mean squared error loss, LPIPS loss, and vision-aided GAN loss, respectively, all computed between the rendered images from the 3DGS and the target view images. 
The weight factors for each loss are denoted by $w_1$, $w_2$, and $w_3$, and We set $w_1=1$ and $w_2=0.05$. 
Regarding $w_3$, we turn it after training undergoes specific iterations, and we utilize the adaptive weighting scheme in~\cite{rombach2022high}.
Specifically, $w_3$ is determined at each training iteration based on the ratio of $l_2$-norm of the gradient of other loss functions to the gradient of the vision-aided GAN loss at the last layer parameters of the GSDecoder. This ratio is then multiplied by 0.1 to set $w_3$.


\subsection{Multi-View Rectified Flow Model}
\label{app:sec:mvrf}

\paragraph{Architecture}

The architecture of our multi-view rectified flow (RF) model is primarily based on the Stable Diffusion 3 medium~\cite{esser2024scaling}\footnote{\href{https://huggingface.co/stabilityai/stable-diffusion-3-medium}{{https://huggingface.co/stabilityai/stable-diffusion-3-medium}}}, with modifications to fit our requirements. We expanded the input and output channels to accommodate concatenated latents, and updated the attention mechanism to incorporate cross-view attention. The model was initialized using pre-trained weights from Stable Diffusion 3. For the input and output layers, the extra channels were initialized by copying the pre-trained weights.

\vspace{\paramargin}
\paragraph{Training}
Following practices in Stable Diffusion 3~\cite{esser2024scaling}, we applied an SNR sampler and used three text encoders.

\vspace{\paramargin}
\paragraph{Sampling Process}
\begin{figure}[t]
\begin{algorithm}[H]
    \caption{Sampling Process of \textit{SplatFlow}} \label{alg:sampling}
    \small
    \textbf{Input:}
    \begin{algorithmic}
        \State $\boldsymbol{u}_\theta$ \hfill $\triangleright$ \textit{Velocity function of SplatFlow}
        \State $\boldsymbol{v}_\phi$ \hfill $\triangleright$ \textit{Velocity function of SD3}
        \State $t = [t_N, \ldots, t_0]$ \hfill $\triangleright$ \textit{Timesteps}
        \State $t_{stop}$ \hfill $\triangleright$ \textit{A time step to stop updating ray latents}
        \State $\boldsymbol{Y}_{t_N} = \bm{X}^{(1:K)}_{t_N} = (\bm{X}^1_{t_N} \dotsc \bm{X}^K_{t_N})\sim \mathcal{N}(0, I)$ \hfill $\triangleright$ \textit{Initial Noise}
    \end{algorithmic}
    \textbf{Sampling:}
    \begin{algorithmic}[1]
        \For{$i=N, \dotsc, 1$}
            \vspace{1mm}
            \State $\boldsymbol{\hat{v}}_{t_i} \gets \boldsymbol{u}_\theta(\boldsymbol{Y}_{t_i}, t_i)$ 
                        
            \vspace{1mm}
            
            \If{$i \geq t_{stop}$}
            \vspace{1mm}
                \If{$N - i \equiv 0 \mod 3 \And i \neq t_{stop} $} 
                    \vspace{1mm}
                    \State $\boldsymbol{\hat{v}}_{t_i}[:n] \gets \boldsymbol{v}_\phi(\boldsymbol{Y}_{t_i}[:n], t_i)$ \hfill $\triangleright$ \textit{Replace to SD3}
                \vspace{1mm}
                \EndIf
                \vspace{1mm}
                \State $\bm{\tilde{Y}}_{t_{0}} \gets \boldsymbol{Y}_{t_i}  - t_i \boldsymbol{u}_\theta(\boldsymbol{Y}_{t_i}, t_i)$ \hfill $\triangleright$ \textit{Predict Destination}
                \vspace{1mm}
                \State $\langle \bm{d}^{(1:K)}, \bm{m}^{(1:K)} \rangle \gets \bm{\tilde{Y}}_{t_{0}} [2n:]$ \hfill $\triangleright$ \textit{Extract Ray Latent}
                \vspace{1mm}
                \For{$j=1, \dotsc, K$}
                    \vspace{1mm}
                    \State $\langle \bm{K}^{j}, \bm{R}^{j}, \bm{T}^{j} \rangle \gets \text{ray\_optimize}(\langle \bm{d}^j, \bm{m}^j \rangle)$
                     \vspace{1mm}
                     \vspace{1mm}
                \EndFor
                \vspace{1mm}
                \State $\langle \bm{K}, \bm{R}^{(1:K)}, \bm{T}^{(1:K)} \rangle \gets \text{shared\_K}(\langle \bm{K}, \bm{R}, \bm{T} \rangle^{(1:K)})$
                \vspace{1mm}
                \State $\langle \bm{d}^{(1:K)}, \bm{m}^{(1:K)} \rangle \gets \text{plücker}(\langle \bm{K}, \bm{R}^{(1:K)}, \bm{T}^{(1:K)} \rangle)$
                \vspace{1mm}
                \State $\bm{r}_{t_0} \gets  \langle \bm{d}^{(1:K)}, \bm{m}^{(1:K)} \rangle$ \hfill $\triangleright$ \textit{Update Ray Destination}
                \vspace{1mm}
            \EndIf
            
            \vspace{1mm}
            
            \State $\boldsymbol{Y}_{t_{i-1}} \gets \boldsymbol{Y}_{t_i} + (t_{i-1} - t_i)\boldsymbol{\hat{v}}_{t_i}(\boldsymbol{Y}_{t_i}, t_i)$
            \vspace{1mm}
            \State $\bm{z} \sim \mathcal{N}(0, I)$
            \vspace{1mm}
            \State $\boldsymbol{Y}_{t_{i-1}}[2n:] \gets (1-t_{i-1})\bm{r}_{t_0} + t_{i-1} \bm{z}$
            \vspace{1mm}
        \EndFor
    \State \textbf{return} $\bm{Y}_{t_0}, \bm{K}, \bm{R}^{(1:K)}, \bm{T}^{(1:K)}$
    \end{algorithmic}
\end{algorithm}
\vspace{-8mm}
\end{figure}
Since our multi-view RF model generates images, depths, and camera poses simultaneously, we modified the sampling process to effectively handle these joint tasks. \Cref{alg:sampling} outlines our sampling procedure, emphasizing three key modifications:
\vspace{0.3cm}
\begin{itemize}
    \item  \textbf{Early stopping of camera pose updates}: We adopt the early stopping approach from RayDiffusion~\cite{zhang2024cameras}, where camera poses are determined early in the sampling process and remain fixed for subsequent steps. This prevents instability and helps maintain a consistent reference frame for the generated views.
    \item \textbf{Intermediate pose optimization with constrained manifold}: To improve camera pose estimation, we introduce a step to predict the sampling destination for ray latents, regress camera poses, and project the resulting Plücker ray representation onto a valid ray manifold at each sampling step. This helps avoid error accumulation and ensures that the poses remain accurate throughout the entire process.
    \item \textbf{Stable Diffusion 3 guidance for generalization}: We integrate vector fields from Stable Diffusion 3~\cite{esser2024scaling} into the sampling process before fixing the camera poses. This enhances the generalizability of our model, especially given the smaller in-the-wild 3D scene datasets we use~\cite{yu2023mvimgnet, ling2024dl3dv}. Applying this guidance early ensures consistency between multi-view images and depths while improving their quality. Also, we use the dual-mode toggling approach similar to Dual3D~\cite{li2024dual3d}, applying the guidance every three sampling steps to balance generalizability with 3D consistency.
\end{itemize}
\vspace{0.5cm}

To regress the camera poses for $j$-th view $\bm{K}^j, \bm{R}^j, \bm{T}^j$ from the Plücker ray representation with $h \times w$ rays, we use the same optimization process in RayDiffusion~\cite{zhang2024cameras} as:
\vspace{-2mm}
\begin{equation}
    \bm{c}^j = \argmin\limits_{\bm{p} \in \mathbb{R}^{3}}{\sum\limits_{\langle \bm{d}^j, \bm{m}^j \rangle \in \mathbb{R}}{\lVert \bm{p} \times \bm{d}^j - \bm{m}^j \rVert^2}},
\end{equation}
\vspace{-2mm}
\begin{equation}
    \bm{P}^j = \argmin\limits_{\lVert \bm{H} \rVert = 1}{\sum\limits_{i=1}^{h \times w}{\lVert \bm{H}\bm{d}^j_i \times \bm{u}_i \rVert}},
\end{equation}
where $\bm{u}$ is the per-pixel ray directions of an identity camera (\ie, $\bm{K}= \bm{I}$ and $\bm{R}= \bm{I}$). Then, the projection matrix $\bm{P}^j$ is decomposed into the intrinsic matrix $\bm{K}^j$ and the rotation matrix $\bm{R}^j$ via DLT~\cite{abdel1971direct}. We further optimize intrinsic and rotation matrices for $K$ views using an Adam~\cite{kingma2014adam} optimizer with 10 iterations (which adds negligible overhead), ensuring that all views share the same intrinsic matrix as:
\vspace{-2mm}
\begin{equation}
    \langle \bm{K}, \bm{R}^{(1:K)} \rangle = \argmin\limits_{\lVert \bm{R} \rVert = 1}{\sum\limits_{j=1}^{K}{\sum\limits_{i=1}^{h \times w}{\lVert \bm{R}^j\bm{d}^j_i \times \bm{u}_{\bm{K}, i} \rVert}}},
\end{equation}
where $\bm{u}_{\bm{K}}$ is the per-pixel ray directions of a camera with the intrinsic matrix $\bm{K}$ and the identity rotation matrix. 
Then, we calculate the translation vector of each view $\bm{T}^{j} = -{\bm{R}^{j}}^\top\bm{c}^j$.
We set the total sampling steps $N$ to 200 and $t_{stop}$ to 150.
We employ different classifier-free guidance~\cite{ho2022classifier} scales to solve ODE for each latent, where we use 7, 5, and 1 for image, depth, and ray latents, respectively. We set the classifier-free guidance scale to 3 for Stable Diffusion 3 guidance.

\begin{figure}[t]
\begin{algorithm}[H]
    \caption{Inpainting Process of \textit{SplatFlow}} \label{alg:inpainting}
    \small
    \textbf{Input:}
    \begin{algorithmic}
        \State $\boldsymbol{u}_\theta$ \hfill $\triangleright$ \textit{Velocity function of SplatFlow}
        \State $t = [t_N, \ldots, t_0]$ \hfill $\triangleright$ \textit{Timesteps}
        \State $t_{stop}$ \hfill $\triangleright$ \textit{A timestep to stop updating ray latents}
        \State $Y_{t_0}^{known}$ \hfill $\triangleright$ \textit{Known latents}
        \State $m$ \hfill $\triangleright$ \textit{Mask for known latents}
        \State $\boldsymbol{Y}_{t_N} = \bm{X}^{(1:K)}_{t_N} = (\bm{X}^1_{t_N} \dotsc \bm{X}^K_{t_N})\sim \mathcal{N}(0, I)$ \hfill $\triangleright$ \textit{Initial noise}
    \end{algorithmic}
    \textbf{Sampling:}
    \begin{algorithmic}[1]
        \For{$i=N, \dotsc, 1$}
            \vspace{1mm}
            \If{$i \geq t_{stop}$}
                \vspace{1mm}
                \State $\bm{\tilde{Y}}_{t_{0}} \gets \boldsymbol{Y}_{t_i}  - t_i \boldsymbol{u}_\theta(\boldsymbol{Y}_{t_i}, t_i)$ \hfill $\triangleright$ \textit{Predict Destination}
                \vspace{1mm}
                \State $\langle \bm{d}^{(1:K)}, \bm{m}^{(1:K)} \rangle \gets \bm{\tilde{Y}}_{t_{0}} [2n:]$ \hfill $\triangleright$ \textit{Extract Ray Latent}
                \vspace{1mm}
                               \For{$j=1, \dotsc, K$}
                    \vspace{1mm}
                    \State $\langle \bm{K}^{j}, \bm{R}^{j}, \bm{T}^{j} \rangle \gets \text{ray\_optimize}(\langle \bm{d}^j, \bm{m}^j \rangle)$
                     \vspace{1mm}
                     \vspace{1mm}
                \EndFor
                \vspace{1mm}
                \State $\langle \bm{K}, \bm{R}^{(1:K)}, \bm{T}^{(1:K)} \rangle \gets \text{shared\_K}(\langle \bm{K}, \bm{R}, \bm{T} \rangle^{(1:K)})$
                \vspace{1mm}
                \State $\langle \bm{d}^{(1:K)}, \bm{m}^{(1:K)} \rangle \gets \text{plücker}(\langle \bm{K}, \bm{R}^{(1:K)}, \bm{T}^{(1:K)} \rangle)$
                \vspace{1mm}
                \State $\bm{r}_{t_0} \gets  \langle \bm{d}^{(1:K)}, \bm{m}^{(1:K)} \rangle$ \hfill $\triangleright$ \textit{Update Ray Destination}
                \vspace{1mm}
            \EndIf
            
            \vspace{1mm}
            
            \State $\boldsymbol{Y}_{t_{i-1}}^\text{unknown} \gets \boldsymbol{Y}_{t_i} + (t_{i-1} - t_i)\boldsymbol{u}_\theta(\boldsymbol{Y}_{t_i}, t_i)(\boldsymbol{Y}_{t_i}, t_i)$
            \vspace{1mm}
            \State $\bm{z} \sim \mathcal{N}(0, I)$            
            \vspace{1mm}
            \State $\boldsymbol{Y}_{t_{i-1}}^\text{unknown}[2n:] \gets (1-t_{i-1})\bm{r}_{t_0} + t_{i-1} \bm{z}$
            \vspace{1mm}
            \State $\bm{\epsilon} \sim \mathcal{N}(0, I)$
            \vspace{1mm}
            \State $Y_{t_{i-1}}^\text{known} = (1-t_{i-1}) Y_{t_0}^\text{known} + t_{i-1} \bm{\epsilon}$
            \vspace{1mm}
            \State $Y_{t_{i-1}} = m \odot Y_{t_{i-1}}^\text{known} + (1-m) \odot Y_{t_{i-1}}^\text{unknown}$
            \vspace{1mm}
        \EndFor
    \State \textbf{return} $\bm{Y}_{t_0}, \bm{K}, \bm{R}^{(1:K)}, \bm{T}^{(1:K)}$
    \end{algorithmic}
\end{algorithm}
\vspace{-8mm}
\end{figure}

\subsection{Inpainting Process}
\label{app:sec:editing}

By integrating the RePaint~\cite{lugmayr2022repaint} into the rectified flow model, our multi-view RF model becomes adaptable to 3DGS editing and training-free downstream tasks such as 3D object replacement, novel view synthesis, and camera pose estimation. \Cref{alg:inpainting} provides an overview of our inpainting process, which incorporates an early stopping strategy and intermediate ray inversion during the sampling process. Furthermore, we utilize RePaint~\cite{lugmayr2022repaint} by leveraging the inversion of known latents to refine the denoised unknown latents at the final stage of each sampling step. As the inpainting process depends on conditions derived from known latents, we exclude the use of Stable Diffusion 3 guidance during the inpainting process.

%

\section{Experimental Setup Details}
\label{app:sec:exp_details}
In this section, we provide comprehensive details regarding the experimental setups used in our paper, extending beyond the brief description given in Section~\textcolor{cvprblue}{5}.

\subsection{Implementation Details}

\paragraph{Dataset}
As described in~\cite{yu2023mvimgnet}, the MVImgNet dataset consists of 219,188 scenes annotated with camera parameters. 
After removing erroneous scenes, we retained approximately 210K scenes. 
From these, 10K scenes were allocated as the validation set, with 1.25K scenes designated for the validation of each specific task.
The rationale for sampling is that fully evaluating all 10K scenes is computationally intensive.
The DL3DV~\cite{ling2024dl3dv} dataset originally contained 10K scenes, but during our experimental period, only 7K scenes were available. Therefore, we utilized the 7K available sequences and allocated 300 sequences as the validation set.
Since neither dataset includes text annotations for each scene, we extracted text descriptions by utilizing the Llava-One Vision Qwen7B model.
A random image was selected to generate the corresponding text descriptions.

\vspace{\paramargin}
\paragraph{Training configuration}

Excluding the validation set described above, we used the remaining dataset to train SplatFlow. Both the GSDecoder and the multi-view RF model were trained with an 8-view setup, sampling 8 viewpoints per scene. Specifically:
\vspace{0.3cm}
\begin{itemize} \item \textbf{GSDecoder:} Trained for 400K iterations with a batch size of 8, using the AdamW optimizer~\cite{loshchilov2017decoupled} and a learning rate of $5 \times 10^{-5}$. The vision-aided GAN loss was activated at 200K iterations, during which the discriminator learning rate was doubled to $1 \times 10^{-4}$. For depth estimation, we used the DepthAnythingV2 Small model~\cite{yang2024depth}\footnote{\href{https://huggingface.co/depth-anything/Depth-Anything-V2-Small}{https://huggingface.co/depth-anything/Depth-Anything-V2-Small}}.
\item \textbf{Multi-view RF model:} Trained for 100K iterations with a batch size of 256, using the AdamW optimizer with a learning rate of $1 \times 10^{-4}$. The learning rate was linearly warmed up for the initial 1K steps. For extracting the depth map, we used the DepthAnythingV2 Small model as in the GSDecoder.
\end{itemize}
\vspace{0.3cm}

\subsection{Detailed Setups in Text-to-3DGS Generation}

\paragraph{Evaluation protocol} We evaluated the performance of our 3DGS generation model using text annotations from the validation sets of the MVImgNet and DL3DV datasets. The corresponding ground-truth images were used as reference images for calculating Fréchet Inception Distance (FID) and CLIP scores. Specifically, we used 10K reference images from MVImgNet and 2.4K reference images from DL3DV.

FID score was calculated using CleanFID~\cite{parmar2022aliased}\footnote{Implementation available at \href{https://github.com/GaParmar/clean-fid}{https://github.com/GaParmar/clean-fid}} to assess the distance between the generated and reference images. CLIP scores were computed using the "openai/clip-vit-base-patch16" model to measure the alignment between the generated images and text descriptions.
These quantitative measurements are conducted for the rendered 8-views.


\subsection{Detailed Setups in 3DGS Editing}

\lstset{
    basicstyle=\ttfamily\footnotesize,
    breaklines=true,
}

\paragraph{Evaluation protocol}
We conducted a benchmark on 100 scenes from the MVImgNet dataset to evaluate object replacement capabilities. For this, we used GPT-4o with the following prompt:
\begin{lstlisting}
You are a vision-language model designed to create captions that describe object replacements in images. Given an input image and its caption, your task is to produce a new caption where the primary object is replaced with a different one, maintaining the overall context and scene.

Guidelines:
1. Object Replacement Focus: Change the main object in the caption to a different but plausible one for the scene. Keep other details consistent with the original setting (e.g., background, lighting).

2. Natural Integration: Ensure the new object fits logically within the environment described. Avoid improbable replacements that clash with the scene's context or elements.

3. Clarity and Directness: Use clear and straightforward language to describe the new object in place of the original, reflecting the same style as the given caption.

4. Single Object Focus: Most images will contain a single object, so focus solely on replacing that object without altering other aspects of the scene unless explicitly instructed.

5. If the given caption describe empty scene like empty hallway, add a 'new object' to the scene.

6. Just return the text.

Example:

- Input Caption: "A white tag with a green leaf design and the text \"HEY!\" on it.|Leaf-shaped tag on hanger, black and white checkered background."
- Target Caption: "A sleek metallic spoon with a reflective surface on a plaid fabric."
\end{lstlisting}
Out of 100 scenes, GPT-4o successfully generated captions for 98 scenes. Two scenes failed due to response refusal from GPT-4o.
Evaluation metrics were then calculated on 8 rendered views per method, using the newly generated captions to guide editing.

\vspace{\paramargin}
\paragraph{Comparison Methods.}
We used DGE~\cite{chen2024dge} and MVInpainter~\cite{cao2024mvinpainter} as baselines by using their official implementations with the following configurations:

\begin{itemize}
    \item \textbf{DGE}~\cite{chen2024dge}: To create 3D Gaussian Splatting (3DGS) representations as the initial point for DGE, all viewpoint images were utilized. Subsequently, 3DGS editing was performed using the provided captions.
    
    \item \textbf{MVInpainter}~\cite{cao2024mvinpainter}: Similar to our method, MVInpainter extracts $8$ views and generates masks for these views. These masks are then used with Stable Diffusion 2, a text-to-image inpainting model, to edit the first view. The remaining views are inpainted based on the edited primary view.
\end{itemize}

\section{Additional Experimental Results}
\label{app:sec:additional_results}

To comprehensively validate the effectiveness of SplatFlow, we provide additional experimental results in this section.

\subsection{Ablation on GSDecoder Design Choice}
\begin{table}[t]
    \centering
    \setlength\tabcolsep{4pt}
    \resizebox{\linewidth}{!}{
    \begin{tabular}{lcccc}
       \toprule
        Method & PSNR$\uparrow$ & LPIPS$\downarrow$ & SSIM$\uparrow$ & FID-50K$\downarrow$ \\
       \midrule
       w/o Depth Latent (200K Iterations) & 25.64 & 0.2507 & 0.7993 & 16.29 \\
       w/ Depth Latent (200K Iterations) & 26.19 & 0.2260 & 0.8169 & 11.92 \\
      \midrule
       w/ Depth Latent (400K Iterations) & 26.68 & 0.2129 & 0.8251 & 8.80 \\
       \rowcolor{gray!25} + Vision-Aided GAN Loss & \textbf{26.84} & \textbf{0.2048} & \textbf{0.8256} & \textbf{5.81} \\
       \arrayrulecolor{black}\bottomrule
    \end{tabular}
    }
    \vspace{-1mm}
    \vspace{\abovetabcapmargin}
    \caption{\textbf{Ablation study on GSDecoder design choices.} Evaluations are performed using PSNR, LPIPS, SSIM, and FID, highlighting the impact of incorporating depth latents and vision-aided GAN loss in improving 3DGS quality.}
    \vspace{-3mm}
    \label{tab:gs_decoder_abl}
\end{table}

To validate the effectiveness of our design choices in the GSDecoder, we conducted ablation studies focusing on two key aspects: (1) the incorporation of depth latents, and (2) the impact of the vision-aided GAN loss. We analyzed these effects by comparing four variants of our GSDecoder:
\begin{itemize}
    \item \textbf{Without Depth Latents (200K iterations):} A baseline variant that excludes depth latents during training to evaluate the effect of incorporating depth information.
    \item \textbf{With Depth Latents (200K iterations):} This version includes depth latents to assess their contribution to improving the quality of the generated 3D Gaussian Splatting.
    \item \textbf{With Depth Latents (400K iterations):} We extended the training by 200K iterations to examine the impact of prolonged training without the vision-aided GAN loss.
    \item \textbf{With Depth Latents + Vision-Aided GAN Loss (400K iterations):} This variant applies the vision-aided GAN loss starting after 200K iterations to evaluate the impact of adversarial training on enhancing 3DGS quality.
\end{itemize}
\vspace{1.2mm}
For training, we utilized the MVImgNet dataset excluding the 10K validation split. We evaluated the generated outputs using 5 rendered images per scene, measured by PSNR, LPIPS~\cite{zhang2018unreasonable}, SSIM~\cite{wang2004image}, and FID~\cite{heusel2017gans}. For FID calculations, we sampled 50K reference images.

The results of this ablation study are illustrated in Table~\ref{tab:gs_decoder_abl}. As shown, incorporating depth latents and vision-aided GAN loss both contributed significantly to improving the quality of 3DGS. 1) Depth latents: Incorporating depth latents led to substantial improvements in PSNR, LPIPS, SSIM, and FID metrics compared to the baseline without depth information. This demonstrates that including depth information enhances the quality and consistency of the decoded scenes. 2) Vision-aided GAN loss: Adding vision-aided GAN loss after 200K iterations yielded the best performance across all metrics. Compared to training without vision-aided GAN loss, it significantly improved perceptual quality, as evidenced by better LPIPS and FID scores.

\begin{figure}[t]
    \centering

    \begin{subfigure}[b]{0.19\linewidth}
        \centering
        \includegraphics[width=\linewidth]{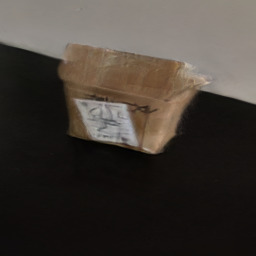}
        \vspace{-5mm}
        \subcaption*{\centering \scriptsize w/o Depth \newline (200K iter)}
    \end{subfigure}
    \begin{subfigure}[b]{0.19\linewidth}
        \centering
        \includegraphics[width=\linewidth]{figure/asset/gsdecoder_qual/wdlatent/novel_view_1.png}
        \vspace{-5mm}
        \subcaption*{\centering \scriptsize w/ Depth \newline (200K iter)}
    \end{subfigure}
    \begin{subfigure}[b]{0.19\linewidth}
        \centering
        \includegraphics[width=\linewidth]{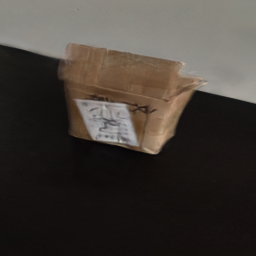}
        \vspace{-5mm}
        \subcaption*{\centering \scriptsize w/o GAN Loss \newline (400K iter)}
    \end{subfigure}
    \begin{subfigure}[b]{0.19\linewidth}
        \centering
        \includegraphics[width=\linewidth]{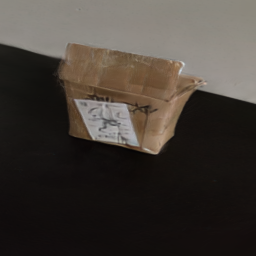}
        \vspace{-5mm}
        \subcaption*{\centering \scriptsize w/ GAN Loss\newline (400K iter)}
    \end{subfigure}
    \begin{subfigure}[b]{0.19\linewidth}
        \centering
        \includegraphics[width=\linewidth]{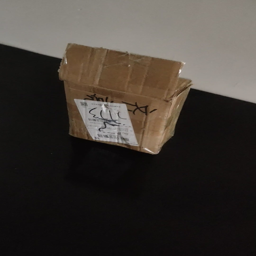}
        \vspace{-5mm}
        \subcaption*{\centering \scriptsize Target\newline View}
    \end{subfigure}

        \begin{subfigure}[b]{0.19\linewidth}
        \centering
        \includegraphics[width=\linewidth]{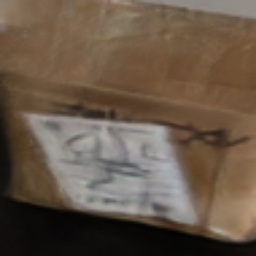}
        \vspace{-5mm}
        \subcaption*{\centering \scriptsize Zoom:\newline w/o Depth\newline (200K iter)}
    \end{subfigure}
    \begin{subfigure}[b]{0.19\linewidth}
        \centering
        \includegraphics[width=\linewidth]{figure/asset/gsdecoder_qual/wdlatent/output_image.jpg}
        \vspace{-5mm}
        \subcaption*{\centering \scriptsize Zoom:\newline w/ Depth\newline (200K iter)}
    \end{subfigure}
    \begin{subfigure}[b]{0.19\linewidth}
        \centering
        \includegraphics[width=\linewidth]{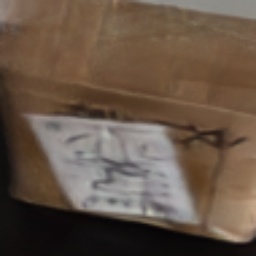}
        \vspace{-5mm}
        \subcaption*{\centering \scriptsize Zoom:\newline w/o GAN Loss\newline (400K iter)}
    \end{subfigure}
    \begin{subfigure}[b]{0.19\linewidth}
        \centering
        \includegraphics[width=\linewidth]{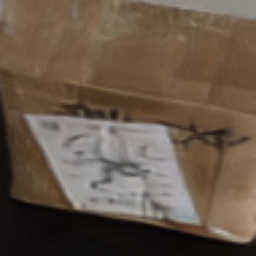}
        \vspace{-5mm}
        \subcaption*{\centering \scriptsize Zoom:\newline w/ GAN Loss\newline (400K iter)}
    \end{subfigure}
    \begin{subfigure}[b]{0.19\linewidth}
        \centering
        \includegraphics[width=\linewidth]{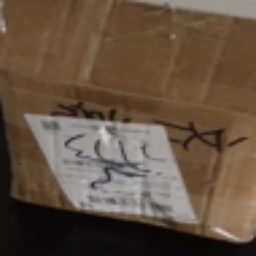}
        \vspace{-5mm}
        \subcaption*{\centering \scriptsize Zoom:\newline Target\newline View}
    \end{subfigure}
    \vspace{\abovefigcapmargin}
    \caption{\textbf{Ablation study on GSDecoder design choices.} The first row shows the original views, while the second row provides zoomed-in details for better visualization. Incorporating depth latents and vision-aided GAN loss enhances the realism and quality of generated 3D Gaussian Splatting (3DGS) scenes.}
    \vspace{-3mm}
    \label{fig:gsdecoder_qualitative}
\end{figure}

Additionally, the qualitative comparison of the four GSDecoder variants is presented in Fig.~\ref{fig:gsdecoder_qualitative}. The images demonstrate that incorporating depth latents significantly enhances the sharpness and detail of the generated scenes compared to the baseline without depth information, leading to more accurate reconstructions of the target view. Furthermore, adding the vision-aided GAN loss at 400K iterations results in the most visually compelling outputs, with enhanced texture details and consistency across views. This progression from the baseline to the final variant clearly highlights the positive impact of both depth information and adversarial training on the quality of novel view synthesis. Specifically, the final configuration (w/ GAN Loss, 400K iter) shows improvements in fine-grained textures and overall coherence, making it visually closer to the target view.

\subsection{Ablation on Sampling Process}

We modified the sampling process to enhance the quality of joint image, depth, and camera pose generation. Here, we present the ablation study for evaluating the impact of the main modifications: 1) early stopping of camera pose updates and 2) Stable Diffusion 3 guidance.

\begin{table}[t]
    \centering
    \setlength\tabcolsep{6pt}
    \resizebox{\linewidth}{!}{
    \begin{tabular}{lcc}
       \toprule
        Method & FID-10K$\downarrow$ & CLIPScore$\uparrow$ \\
       \midrule
       Stop-Ray ($t_{\text{stop}} = 100$) & 35.55 & 31.37 \\
       Stop-Ray ($t_{\text{stop}} = 50$) & 37.89 & 31.41 \\
       w/o Stop-Ray ($t_{\text{stop}} = 0$) & 47.32 & 30.12 \\
       \midrule
       \rowcolor{gray!25} SplatFlow - Default ($t_{\text{stop}} = 150$) & \textbf{34.85} & \textbf{31.43} \\
       \arrayrulecolor{black}\bottomrule
    \end{tabular}
    }
    \vspace{-1mm}
    \vspace{\abovetabcapmargin}
    \caption{\textbf{Impact of the Stop-Ray modification.} Evaluations are conducted using FID-10K and CLIPScore metrics to assess the effectiveness of stopping camera ray updates at different timesteps in the sampling process. }
    \vspace{-1mm}
    \label{tab:stop_ray_ablation}
\end{table}

\vspace{\paramargin}
\paragraph{Effect of Early Stopping}

To assess the impact of early stopping for camera pose updates, we varied the stopping step ($t_{\text{stop}}$) at different values: 150 (the original, base setup), 100, 50, and 0 (no early stopping) during 200 total sampling steps. As shown in Table~\ref{tab:stop_ray_ablation}, stopping early at $t_{\text{stop}} = 150$ results in the best FID-10K and CLIPScore, indicating that fixing the camera poses early stabilizes the generated views. At $t_{\text{stop}} = 100$, there is a slight degradation in FID-10K and CLIPScore compared to $t_{\text{stop}} = 150$, and the performance further drops at $t_{\text{stop}} = 50$, which suggests that extending the camera pose updates introduces more degradation in the generated views. When no early stopping is applied ($t_{\text{stop}} = 0$), both metrics degrade significantly, highlighting the increased instability in camera pose updates over the entire sampling process. These results underline that stopping the ray updates early, preferably at $t_{\text{stop}} = 150$, is crucial for maintaining high-quality generation.

\vspace{\paramargin}
\paragraph{Effect of Stable Diffusion 3 guidance}
\begin{table}[t]
    \centering
    \setlength\tabcolsep{8pt}
    \resizebox{\linewidth}{!}{
    \begin{tabular}{lcc}
       \toprule
        Method & FID-10K$\downarrow$ & CLIPScore$\uparrow$ \\
       \midrule
       SplatFlow (w/o SD3 Guidance) & 34.88 & 30.67 \\
       \rowcolor{gray!25} SplatFlow (full model) & \textbf{34.85} & \textbf{31.43} \\
       \arrayrulecolor{black}\bottomrule
    \end{tabular}
    }
    \vspace{-1mm}
    \vspace{\abovetabcapmargin}
    \caption{\textbf{Impact of Stable Diffusion 3 Guidance.} The table compares the FID-10K and CLIPScore metrics for SplatFlow with and without SD3 guidance.}
    \vspace{-3mm}
    \label{tab:sd3_guidance_ablation}
\end{table}
To measure the effects of Stable Diffusion 3 (SD3) guidance, we compared the original SplatFlow model, which includes SD3 guidance, against a variant without it. As shown in Table~\ref{tab:sd3_guidance_ablation}, removing SD3 guidance leads to a slight increase in FID-10K (34.88 compared to 34.85) and a notable drop in CLIPScore (from 31.43 to 30.67). These results indicate that SD3 guidance contributes positively to the consistency between generated images and the provided text prompts, thereby improving the alignment and quality of the generated outputs. Including SD3 guidance ensures better generalizability and alignment, particularly for smaller in-the-wild datasets.

\subsection{More Results on Text-to-3DGS Generation}

\paragraph{Generalizability evaluation}
\begin{table}[t]
    \centering
    \setlength\tabcolsep{3pt}
    \resizebox{\linewidth}{!}{
    \begin{tabular}{lccc}
       \toprule
        Method   & BRISQUE$\downarrow$ & NIQE$\downarrow$ & CLIPScore$\uparrow$ \\
       \midrule
       DreamFusion~\cite{poole2022dreamfusion} & 90.2 & 10.48 & -   \\
       Magic3D~\cite{lin2023magic3d} & 92.8 & 11.20 & -   \\
       LatentNeRF~\cite{metzer2023latent} & 88.6 & 9.19 & - \\
       SJC~\cite{wang2023score} & 82.0 & 10.15 & - \\
       Fantasia3D~\cite{chen2023fantasia3d} & 69.6 & 7.65 & - \\
       ProlificDreamer~\cite{wang2024prolificdreamer} & 61.5 & 7.07 & - \\
       \arrayrulecolor{gray}\midrule
       Director3D~\cite{li2024director3d} & 37.1 & 6.41 & 32.0 \\
       Director3D (w/ SDS++)~\cite{li2024director3d} & 32.3 & 4.35 & 32.9 \\
       \midrule
       \rowcolor{gray!25} \textbf{SplatFlow} & \textbf{16.8} & 5.88 & 28.9 \\
      \rowcolor{gray!25} \textbf{SplatFlow (w/ SDS++)} & 19.6 & \textbf{4.24} & \textbf{33.2}  \\
       \arrayrulecolor{black}\bottomrule
    \end{tabular}
    }
    \vspace{-1mm}
    \vspace{\abovetabcapmargin}
    \caption{\textbf{Quantitative results in Single-Object-with-Surrounding set of T3Bench~\cite{he2023t3bench}.} For the CLIPScore, we report our reproduced score due to an error in the measurement of Director3D~\cite{li2024director3d}.}
    \vspace{\belowtabcapmargin}
    \label{tab:t3bench}
\end{table}

Although our SplatFlow is trained on the MVImgNet~\cite{yu2023mvimgnet} and the DL3DV~\cite{ling2024dl3dv} datasets, we conducted an experiment on Single-Ojbect-with-Surrounding sets of T3Bench~\cite{he2023t3bench} to validate the generalizability of our SplatFlow for unseen domain texts.  
Following evaluation protocols in Director3D~\cite{li2024director3d}, we utilized the BRISQUE~\cite{mittal2012no} and the NIQE~\cite{mittal2012making} to evaluate the image quality and the CLIPScore~\cite{hessel2021clipscore} to measure alignment with text prompts.

\Cref{tab:t3bench} demonstrates that our SplatFlow outperforms the previous text-to-3D generation methods across all metrics. Notably, our SplatFlow achieves a significantly lower score than other methods in the BRISQUE metric even without the refining process.
Our SplatFlow performs worse than Director3D~\cite{li2024director3d} in CLIPScore when both models are evaluated without a refining process, suggesting that SplatFlow has lower generalizability to text prompts due to its smaller training dataset. However, SplatFlow generates significantly higher-quality images compared to Director3D. This allows the refining process to focus primarily on aligning with the text prompts rather than enhancing image quality, leading to a substantial improvement in CLIPScore and ultimately surpassing Director3D~\cite{li2024director3d}.

\vspace{\paramargin}
\paragraph{Qualitative results}

\begin{figure*}[ht]
    \centering
    \setlength\tabcolsep{5pt}
    \begin{tabular}{cccccc}
        \toprule
        \multicolumn{6}{l}{\textbf{MVImgNet~\cite{yu2023mvimgnet} validation set}} \\
        \toprule
        \adjincludegraphics[clip,width=0.19\linewidth,trim={0 0 0 0}]{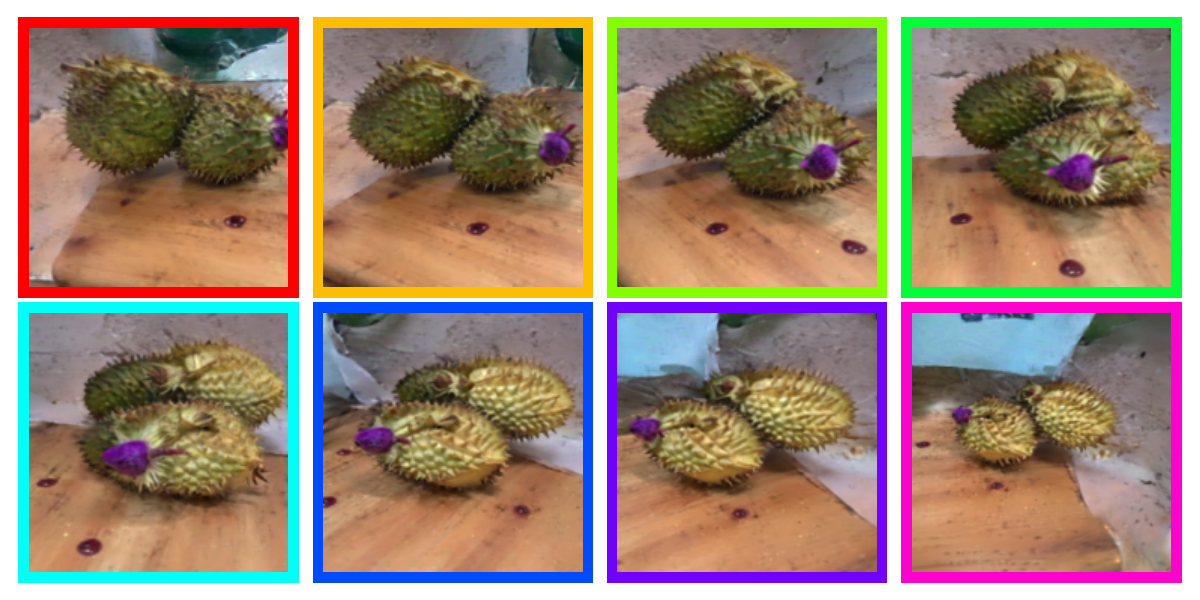} &
        \adjincludegraphics[clip,width=0.095\linewidth,trim={0 0 0 0}]{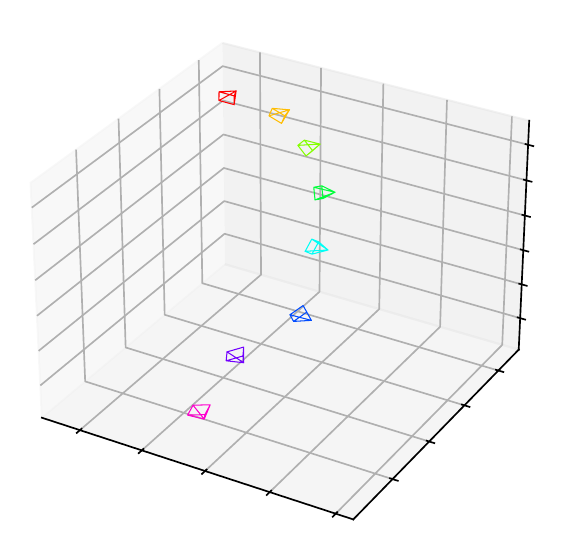} &
        \adjincludegraphics[clip,width=0.19\linewidth,trim={0 0 0 0}]{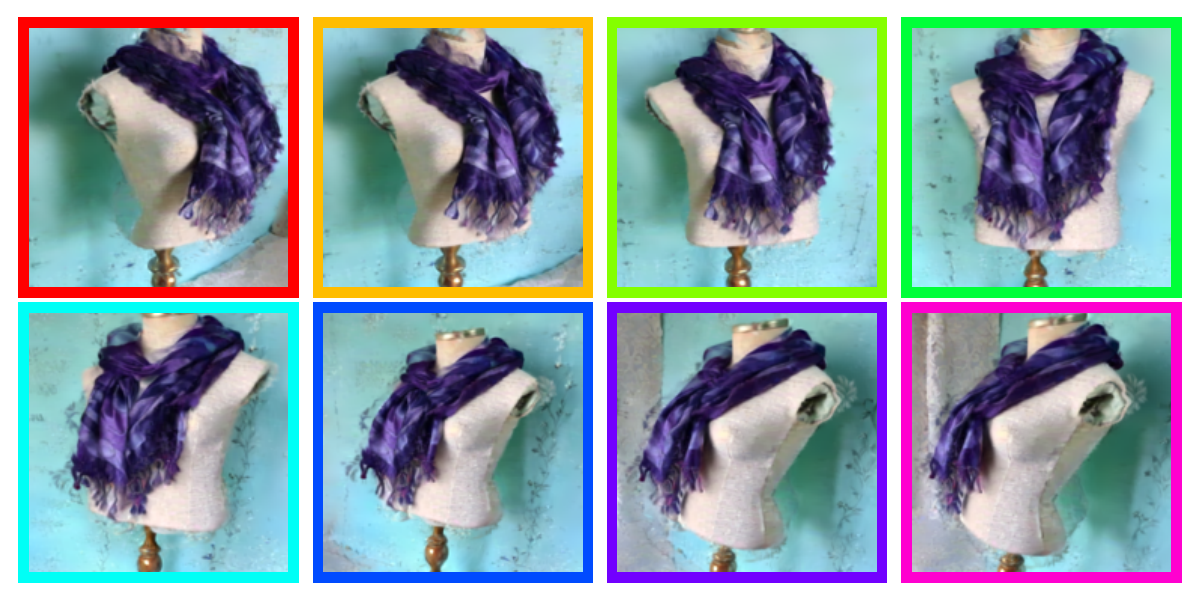} &
        \adjincludegraphics[clip,width=0.095\linewidth,trim={0 0 0 0}]{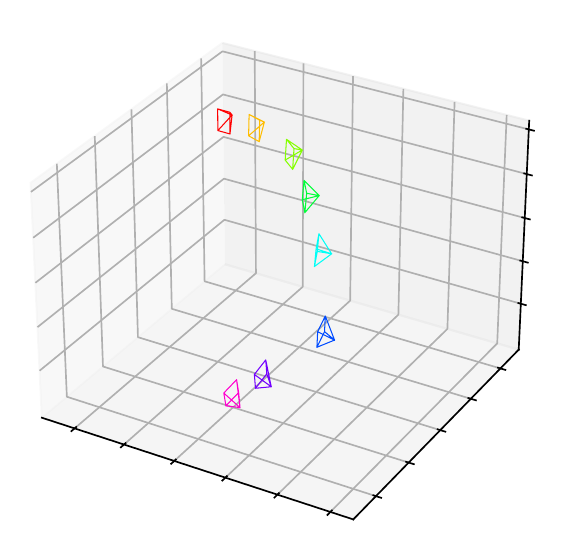} &
        \adjincludegraphics[clip,width=0.19\linewidth,trim={0 0 0 0}]{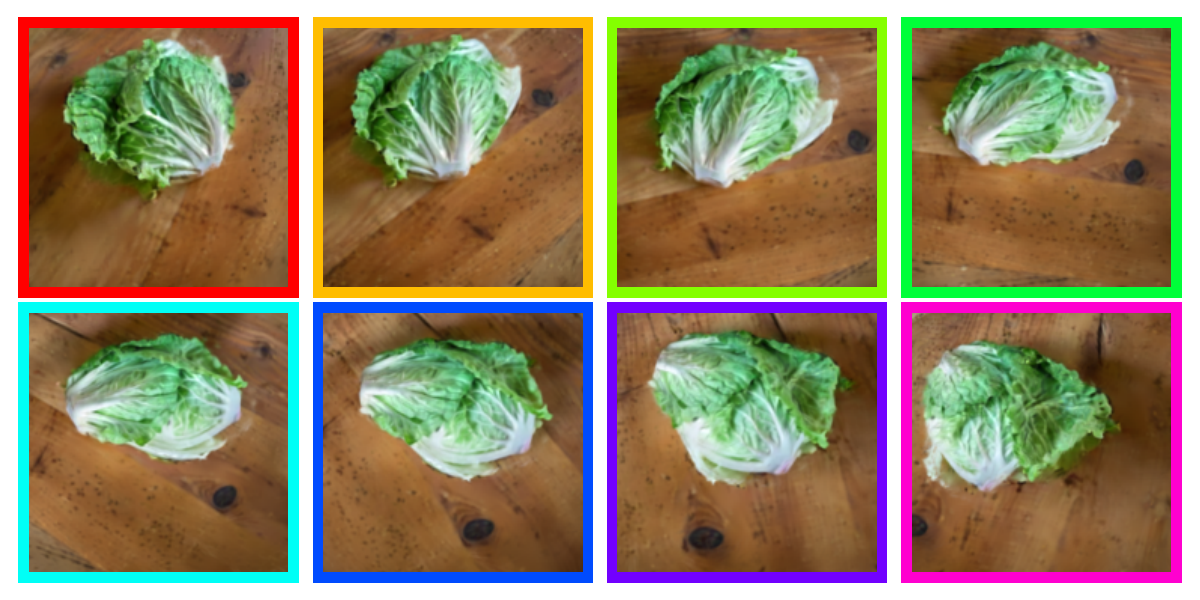} &
        \adjincludegraphics[clip,width=0.095\linewidth,trim={0 0 0 0}]{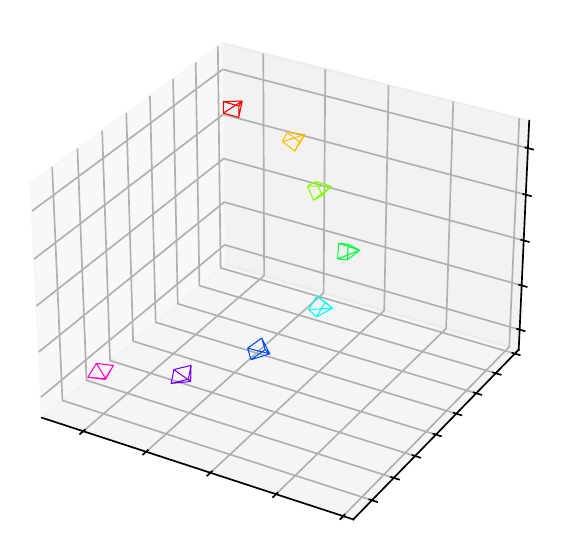} \\
        \multicolumn{2}{p{0.3\linewidth}}{\centering\scriptsize \textit{``Two durians with spiky skins and a purple tag"}} & \multicolumn{2}{p{0.3\linewidth}}{\centering\scriptsize \textit{``Mannequin torso with scarf, turquoise background"}} & \multicolumn{2}{p{0.3\linewidth}}{\centering\scriptsize \textit{``Top view of a cabbage on a wooden surface"}} \\

        \arrayrulecolor{white}\midrule
        
        \adjincludegraphics[clip,width=0.19\linewidth,trim={0 0 0 0}]{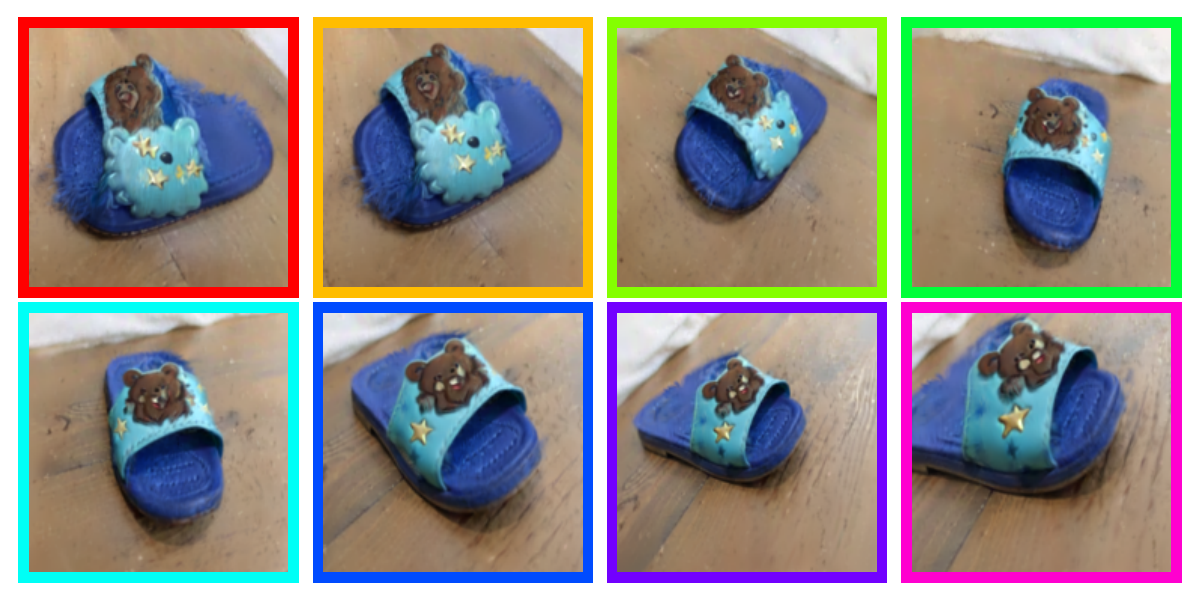} &
        \adjincludegraphics[clip,width=0.095\linewidth,trim={0 0 0 0}]{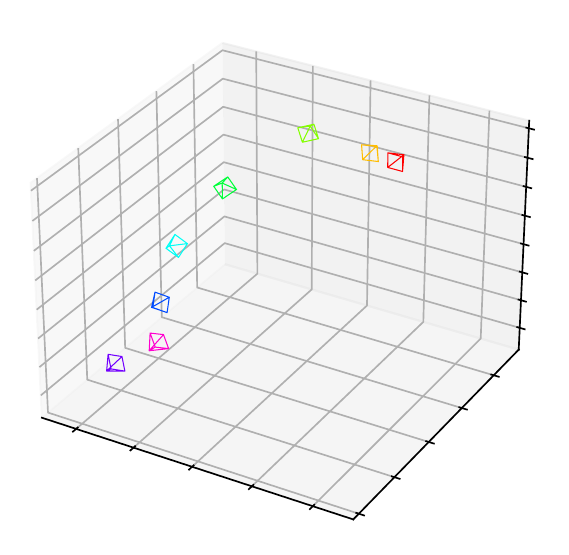} &
        \adjincludegraphics[clip,width=0.19\linewidth,trim={0 0 0 0}]{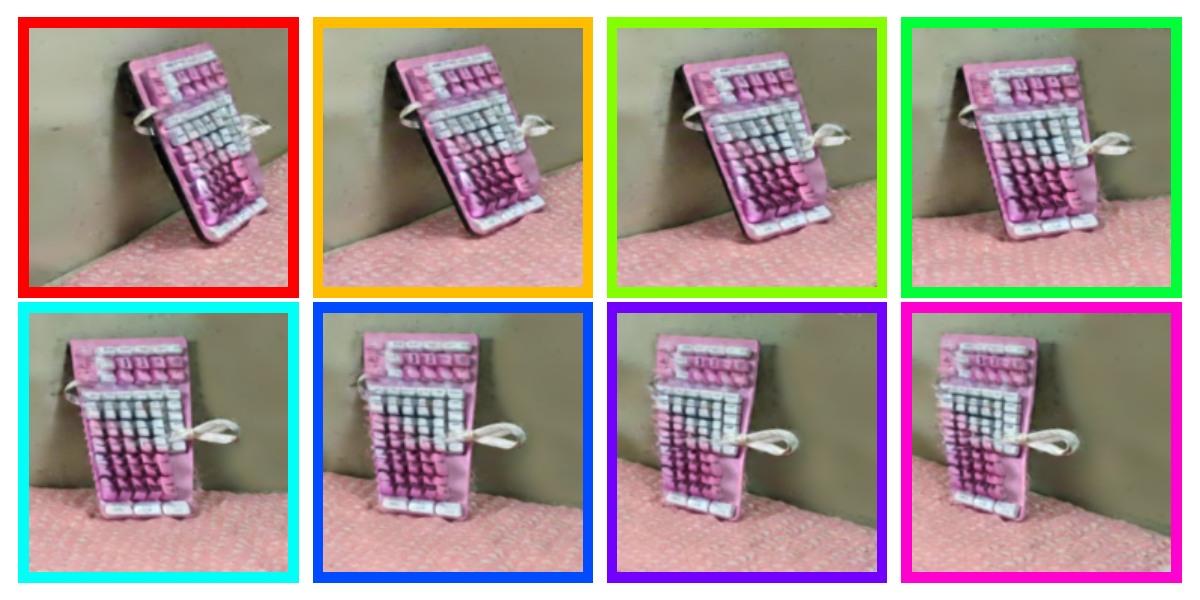} &
        \adjincludegraphics[clip,width=0.095\linewidth,trim={0 0 0 0}]{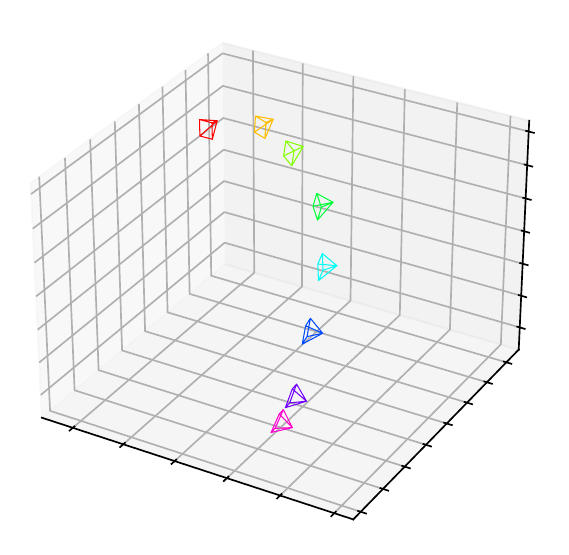} &
        \adjincludegraphics[clip,width=0.19\linewidth,trim={0 0 0 0}]{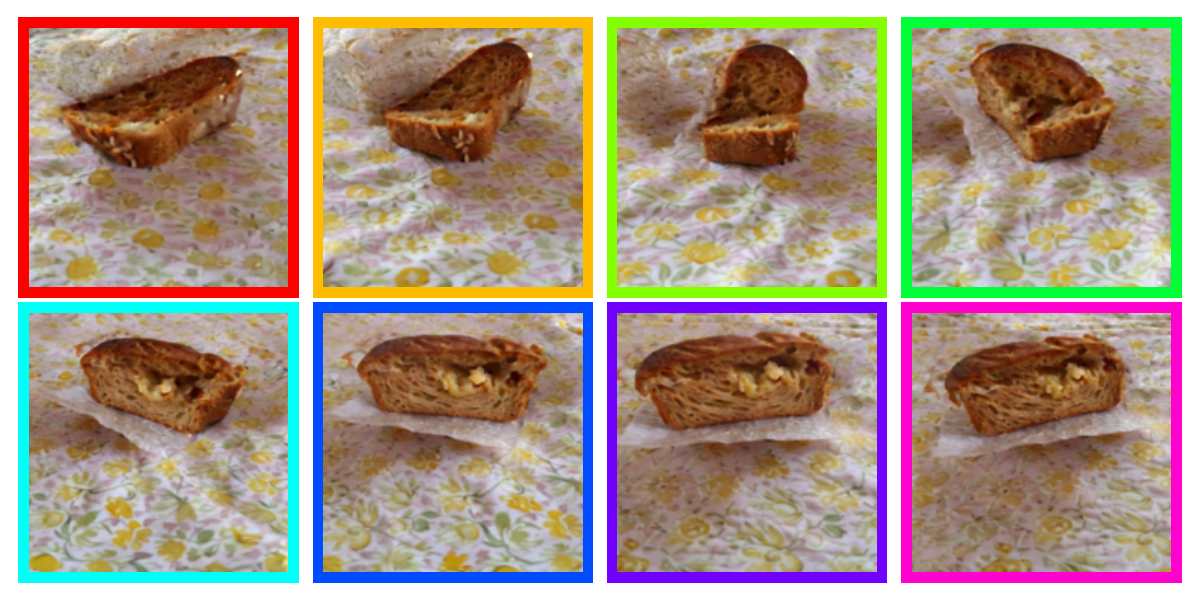} &
        \adjincludegraphics[clip,width=0.095\linewidth,trim={0 0 0 0}]{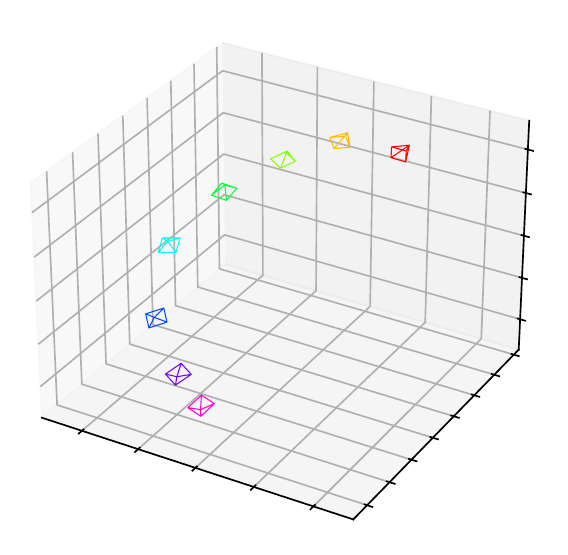} \\
        \multicolumn{2}{p{0.3\linewidth}}{\centering\scriptsize \textit{``Blue slide sandal with bear face and stars"}} & \multicolumn{2}{p{0.3\linewidth}}{\centering\scriptsize \textit{``Pink and white keyboard with pink keys"}} & \multicolumn{2}{p{0.3\linewidth}}{\centering\scriptsize \textit{``A piece of bread, resting on a patterned surface"}} \\

        \arrayrulecolor{white}\midrule

        \arrayrulecolor{black}\toprule
        \multicolumn{6}{l}{\textbf{DL3DV~\cite{ling2024dl3dv} validation set}} \\
        \toprule
        \adjincludegraphics[clip,width=0.19\linewidth,trim={0 0 0 0}]{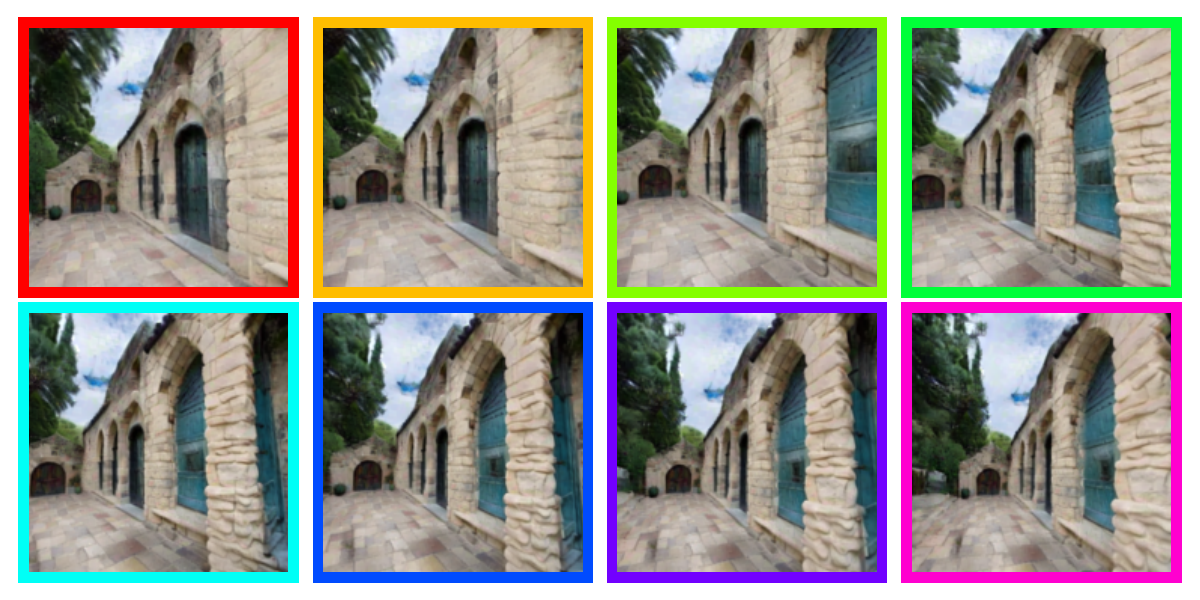} &
        \adjincludegraphics[clip,width=0.095\linewidth,trim={0 0 0 0}]{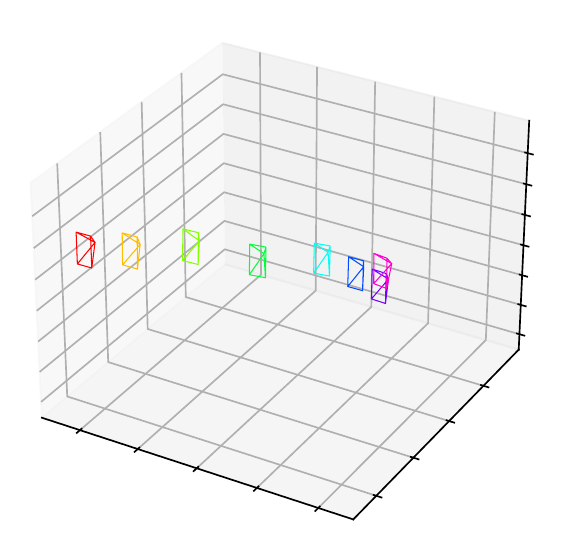} &
        \adjincludegraphics[clip,width=0.19\linewidth,trim={0 0 0 0}]{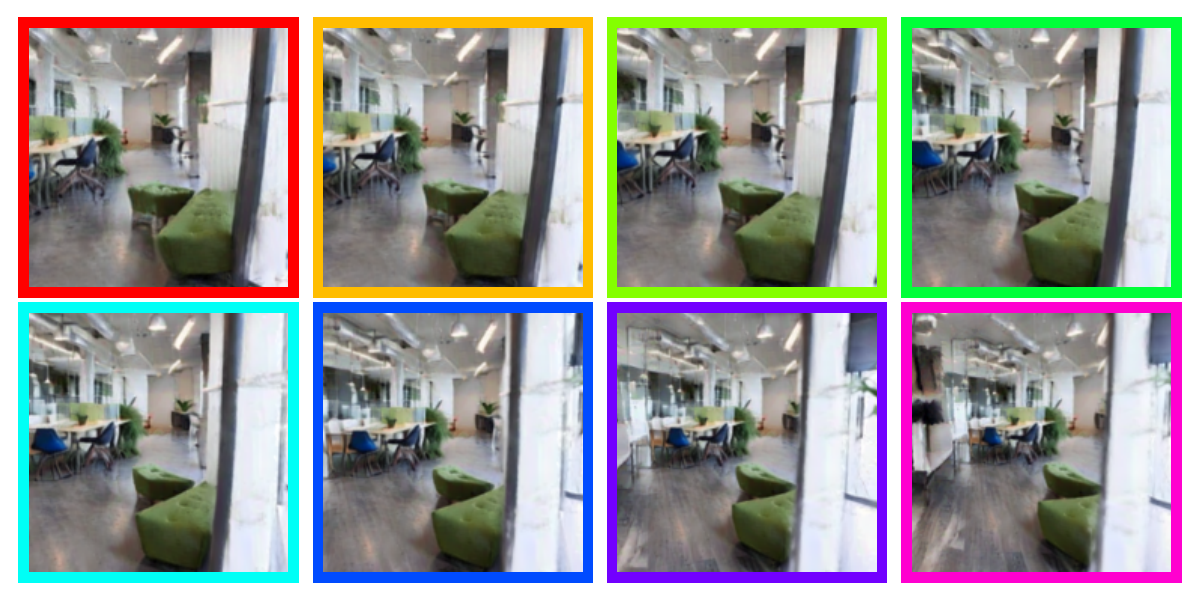} &
        \adjincludegraphics[clip,width=0.095\linewidth,trim={0 0 0 0}]{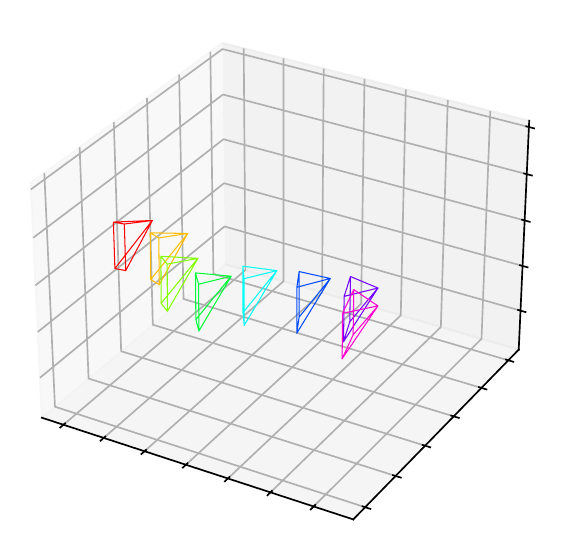} &
        \adjincludegraphics[clip,width=0.19\linewidth,trim={0 0 0 0}]{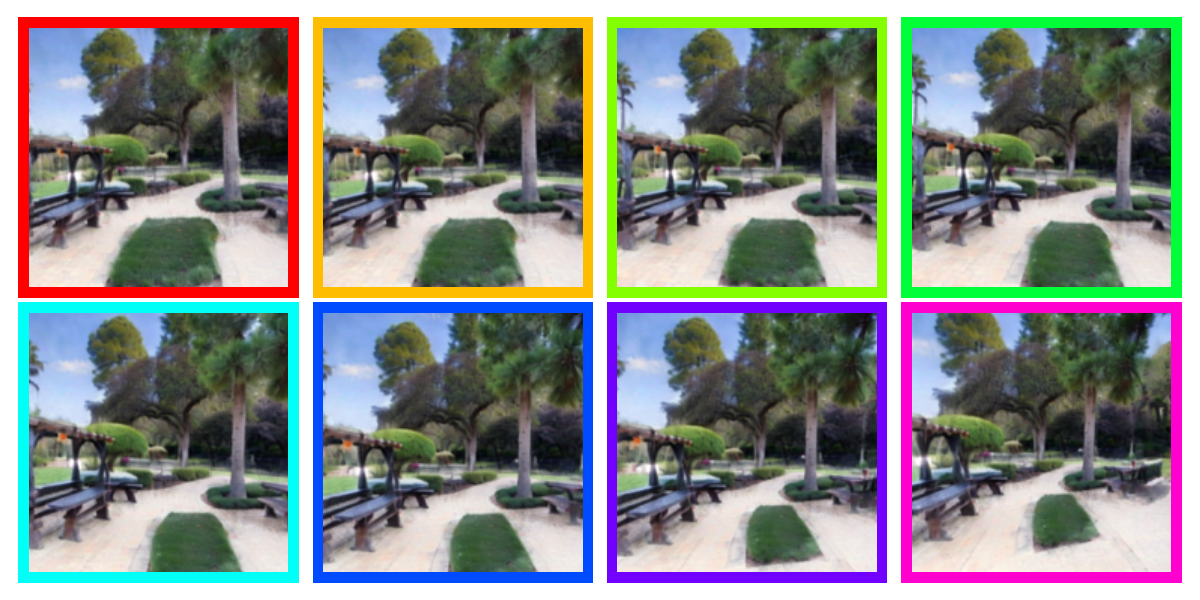} &
        \adjincludegraphics[clip,width=0.095\linewidth,trim={0 0 0 0}]{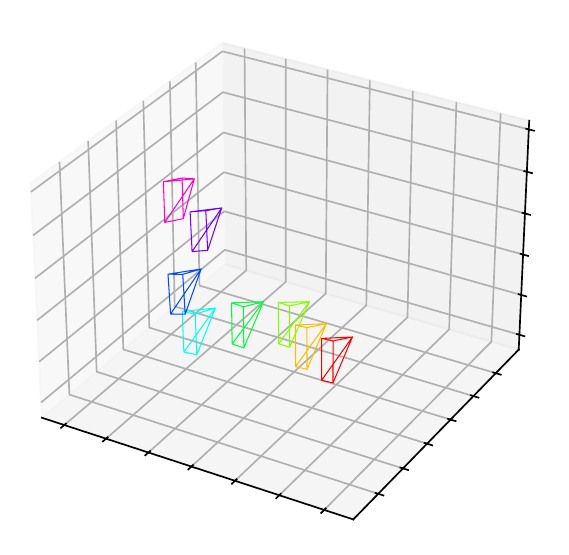} \\
        \multicolumn{2}{p{0.3\linewidth}}{\centering\scriptsize \textit{``A stone building with a tiled floor and an arched window"}} & \multicolumn{2}{p{0.3\linewidth}}{\centering\scriptsize \textit{``A modern office space with a bench, chairs, and tables"}} & \multicolumn{2}{p{0.3\linewidth}}{\centering\scriptsize \textit{``A serene park with picnic tables, benches, and a gazebo"}} \\
\arrayrulecolor{white}\midrule
        
        \adjincludegraphics[clip,width=0.19\linewidth,trim={0 0 0 0}]{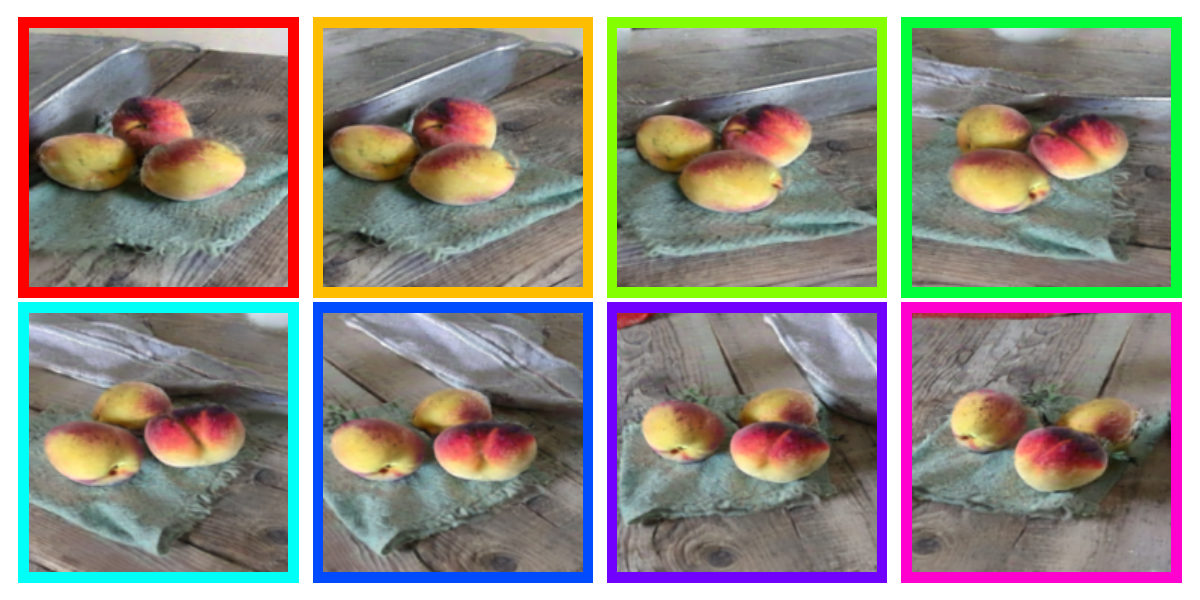} &
        \adjincludegraphics[clip,width=0.095\linewidth,trim={0 0 0 0}]{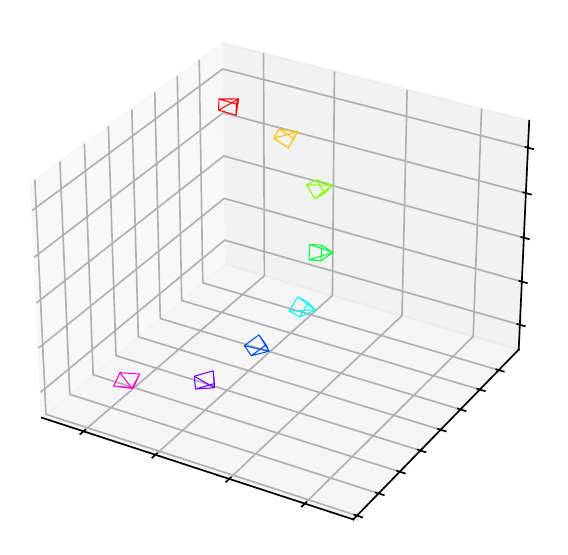} &
        \adjincludegraphics[clip,width=0.19\linewidth,trim={0 0 0 0}]{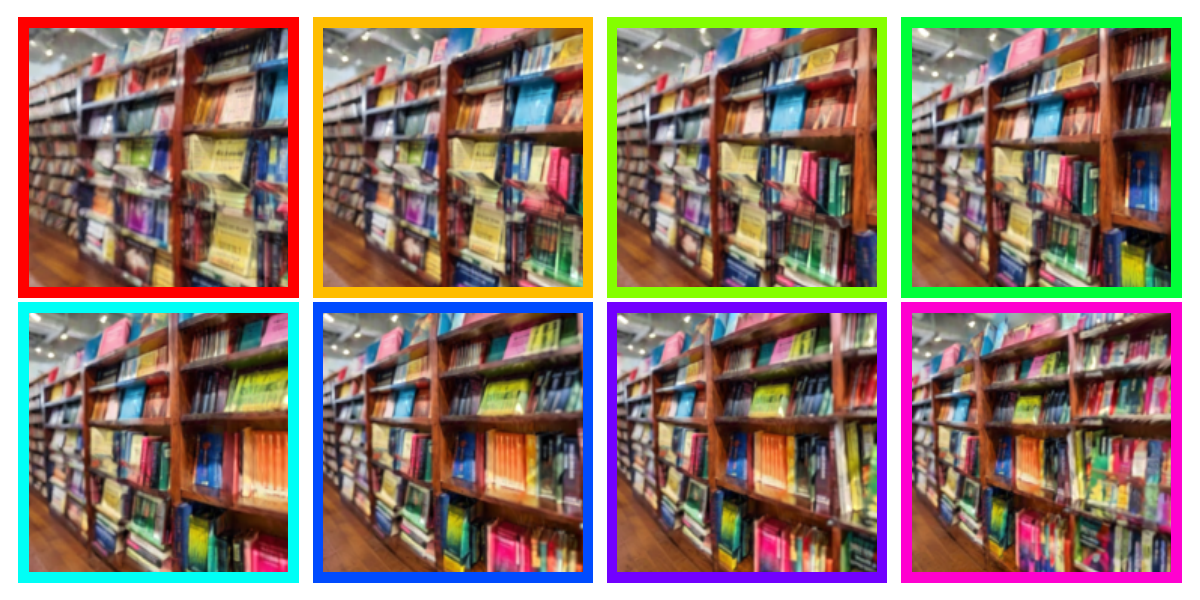} &
        \adjincludegraphics[clip,width=0.095\linewidth,trim={0 0 0 0}]{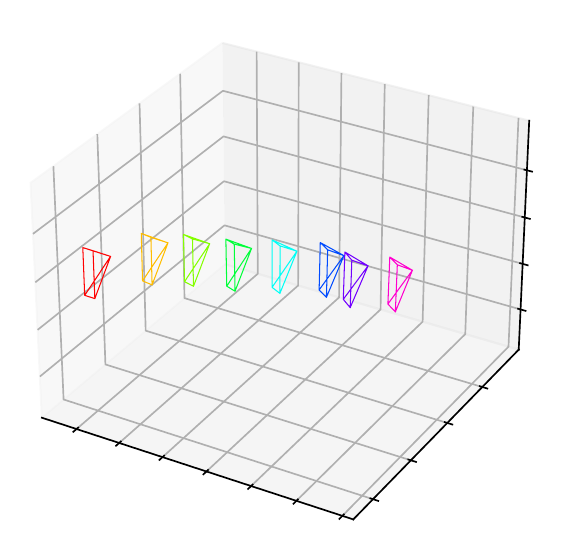} &
        \adjincludegraphics[clip,width=0.19\linewidth,trim={0 0 0 0}]{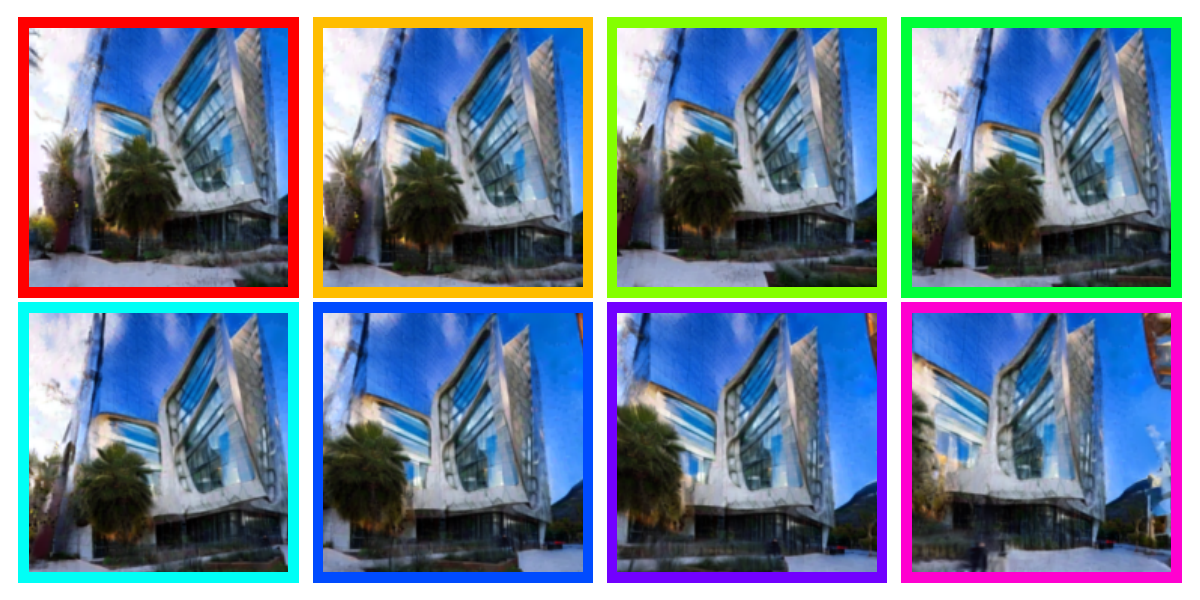} &
        \adjincludegraphics[clip,width=0.095\linewidth,trim={0 0 0 0}]{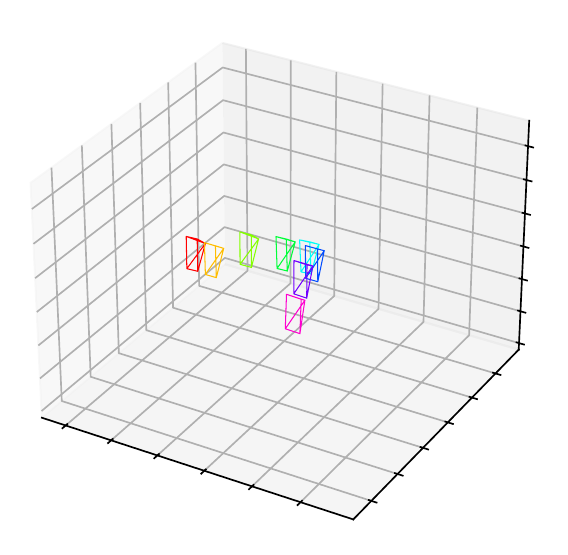} \\
        \multicolumn{2}{p{0.3\linewidth}}{\centering\scriptsize \textit{``A blue cloth on a wooden surface with three peaches"}} & \multicolumn{2}{p{0.3\linewidth}}{\centering\scriptsize \textit{``Rows of colorful books on shelves in a bookstore"}} & \multicolumn{2}{p{0.3\linewidth}}{\centering\scriptsize \textit{``A modern building with a unique facade and large windows"}} \\

        \arrayrulecolor{white}\midrule
        
        \arrayrulecolor{black}\toprule
        \multicolumn{6}{l}{\textbf{T3Bench~\cite{he2023t3bench} - Single-Object-with-Surrounding set}} \\
        \toprule
        \adjincludegraphics[clip,width=0.19\linewidth,trim={0 0 0 0}]{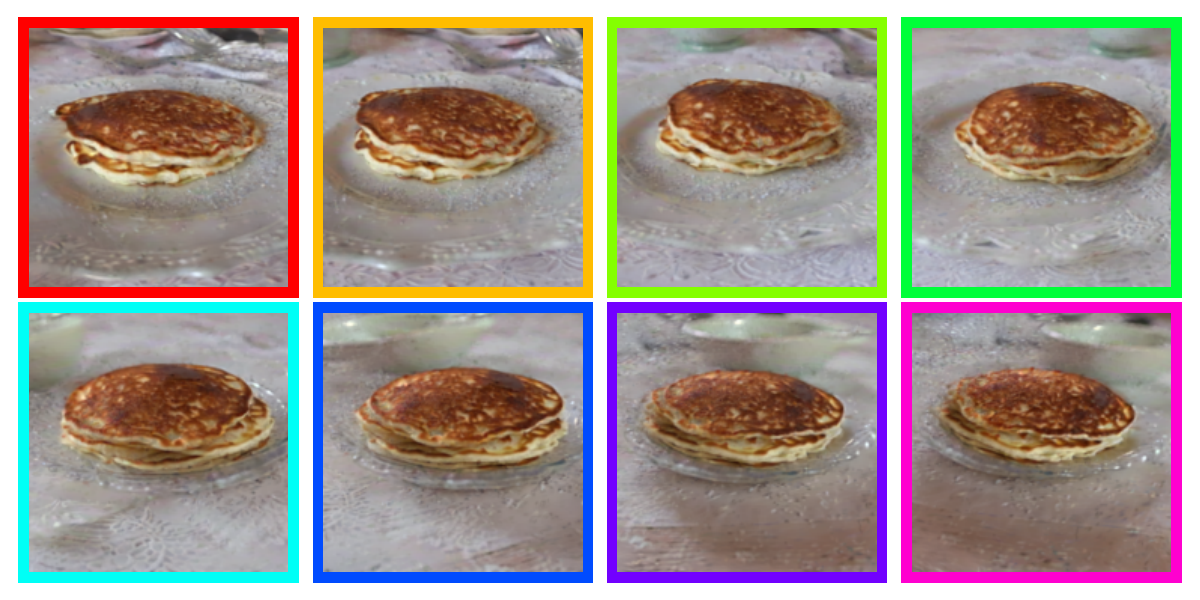} &
        \adjincludegraphics[clip,width=0.095\linewidth,trim={0 0 0 0}]{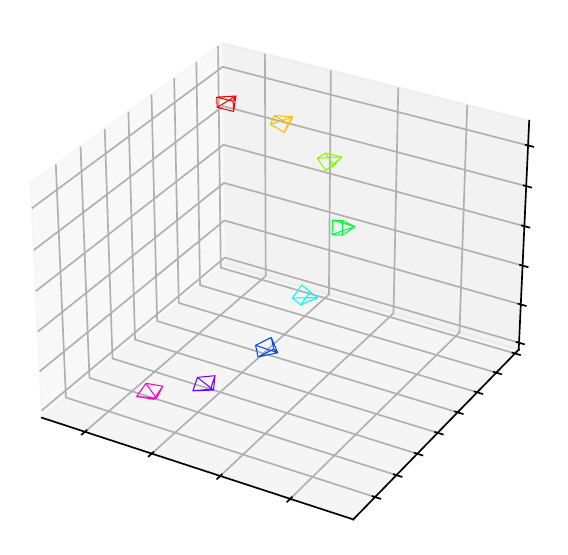} &
        \adjincludegraphics[clip,width=0.19\linewidth,trim={0 0 0 0}]{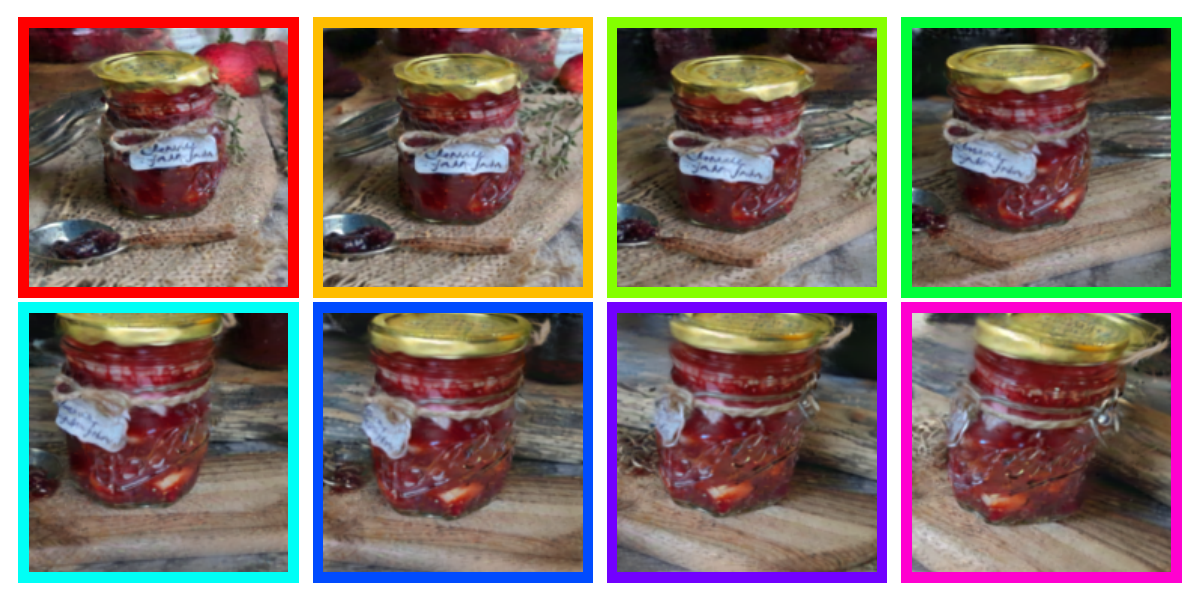} &
        \adjincludegraphics[clip,width=0.095\linewidth,trim={0 0 0 0}]{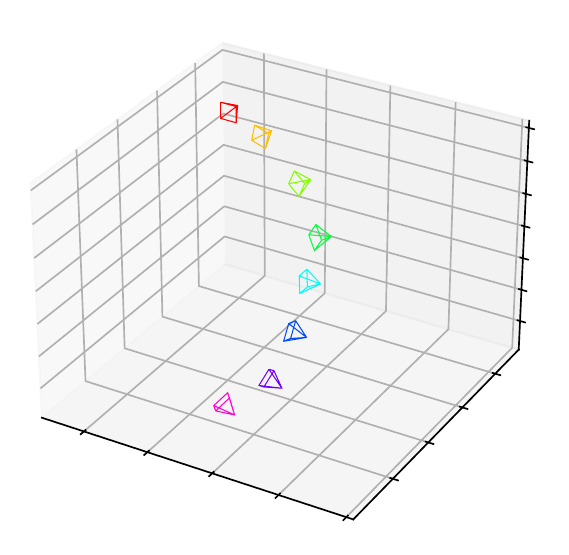} &
        \adjincludegraphics[clip,width=0.19\linewidth,trim={0 0 0 0}]{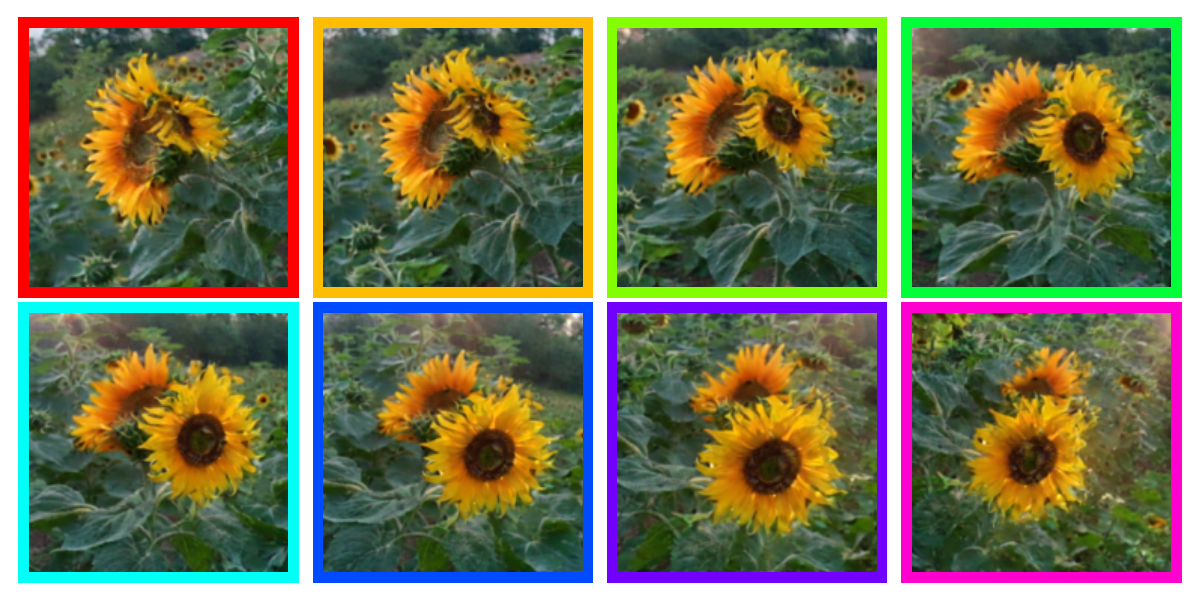} &
        \adjincludegraphics[clip,width=0.095\linewidth,trim={0 0 0 0}]{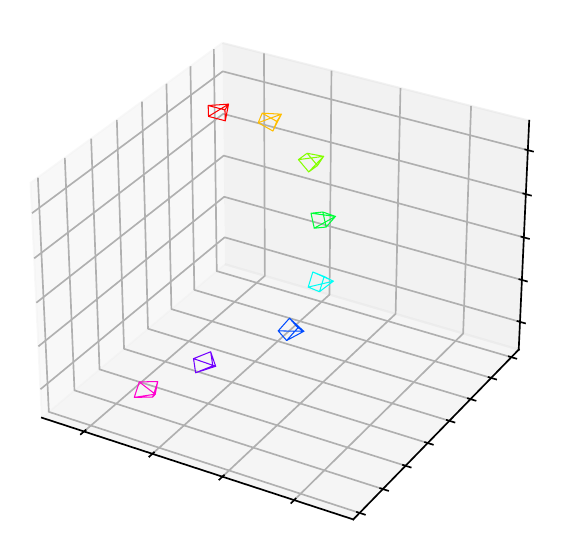} \\
        \multicolumn{2}{p{0.3\linewidth}}{\centering\scriptsize \textit{``A stack of pancakes on a breakfast table"}} & \multicolumn{2}{p{0.3\linewidth}}{\centering\scriptsize \textit{``A jar of homemade jam on a kitchen counter"}} & \multicolumn{2}{p{0.3\linewidth}}{\centering\scriptsize \textit{``A bright sunflower in a field"}} \\
\arrayrulecolor{white}\midrule

        \adjincludegraphics[clip,width=0.19\linewidth,trim={0 0 0 0}]{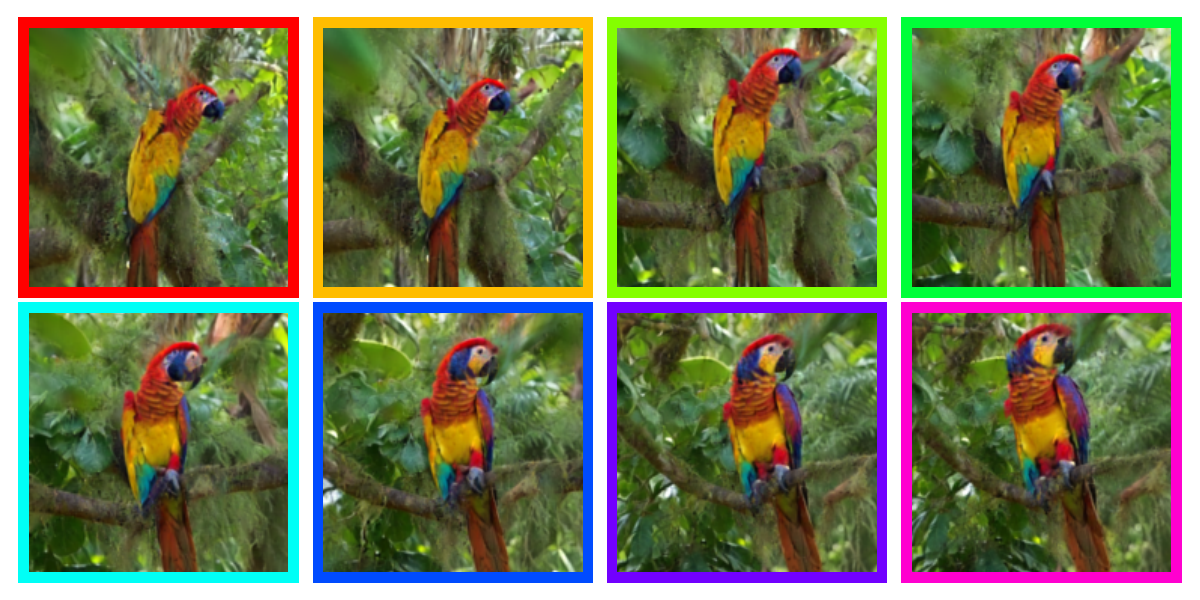} &
        \adjincludegraphics[clip,width=0.095\linewidth,trim={0 0 0 0}]{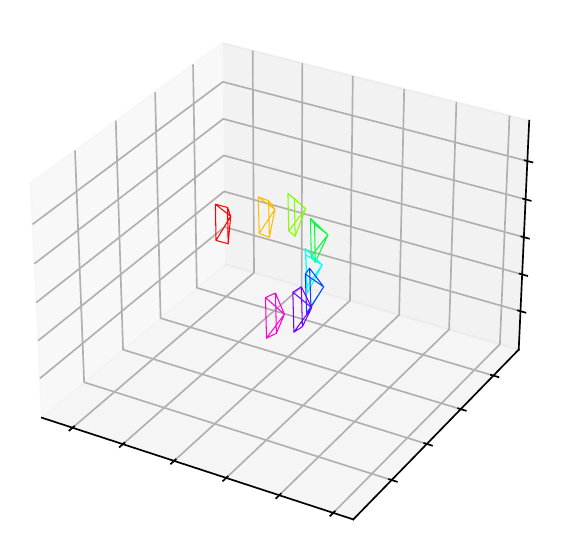} &
        \adjincludegraphics[clip,width=0.19\linewidth,trim={0 0 0 0}]{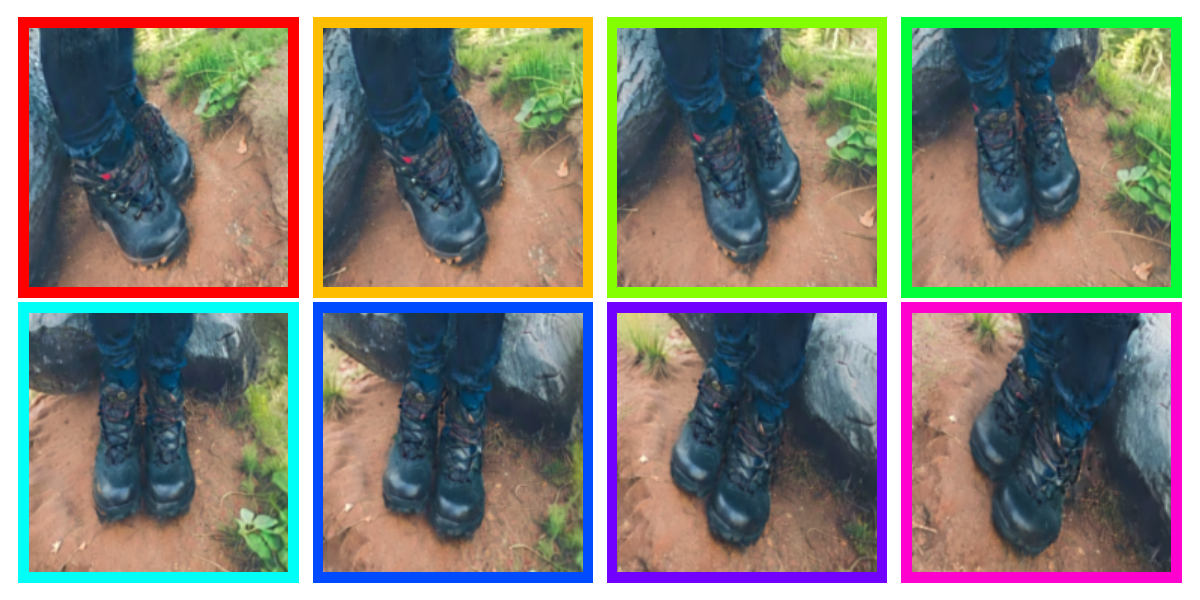} &
        \adjincludegraphics[clip,width=0.095\linewidth,trim={0 0 0 0}]{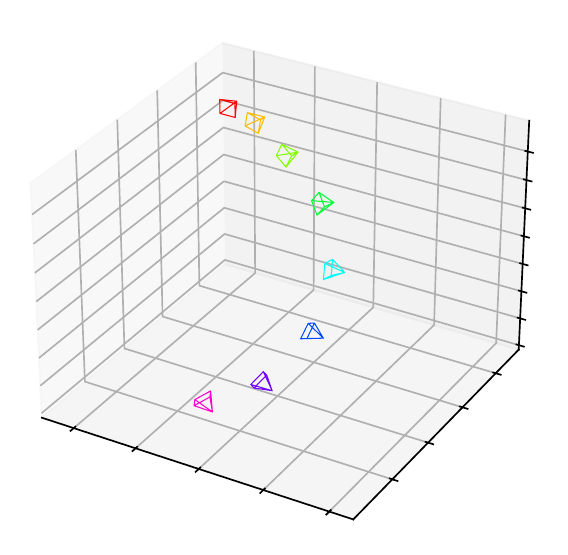} &
        \adjincludegraphics[clip,width=0.19\linewidth,trim={0 0 0 0}]{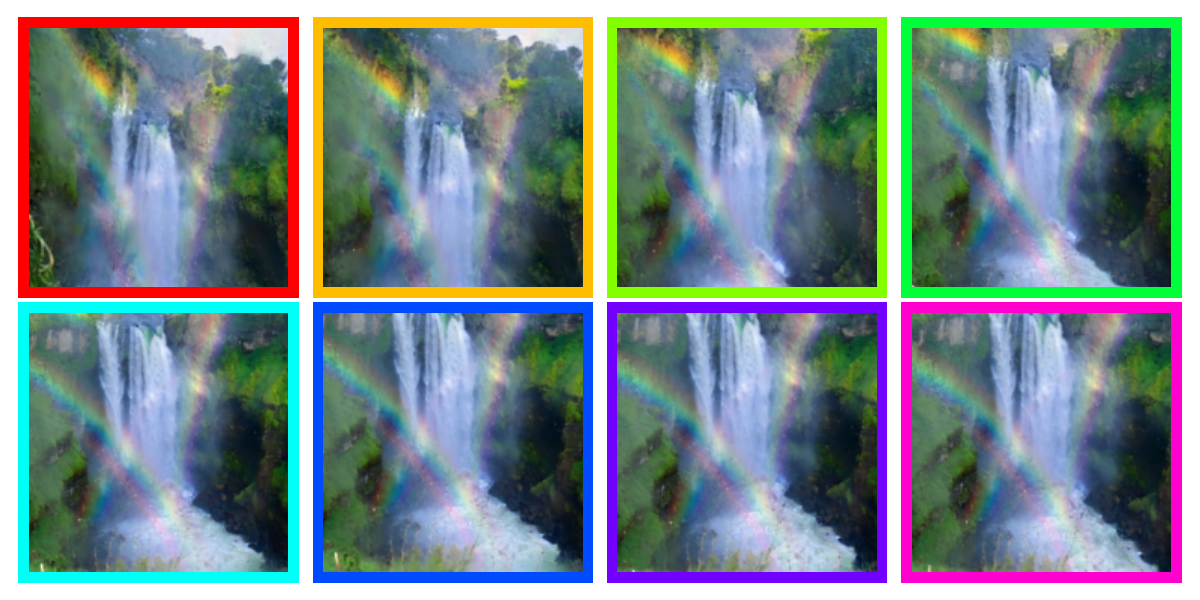} &
        \adjincludegraphics[clip,width=0.095\linewidth,trim={0 0 0 0}]{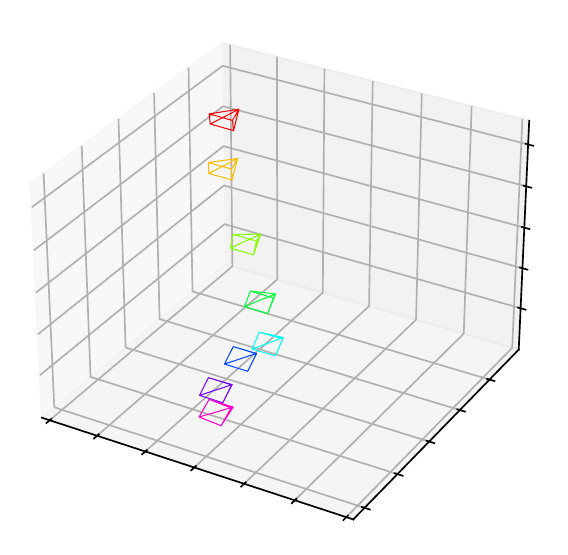} \\
        \multicolumn{2}{p{0.3\linewidth}}{\centering\scriptsize \textit{``A colorful parrot on a jungle tree"}} & \multicolumn{2}{p{0.3\linewidth}}{\centering\scriptsize \textit{``A pair of hiking boots on a trail"}} & \multicolumn{2}{p{0.3\linewidth}}{\centering\scriptsize \textit{``A rainbow over a waterfall"}} \\
        \arrayrulecolor{white}\midrule
        
        \adjincludegraphics[clip,width=0.19\linewidth,trim={0 0 0 0}]{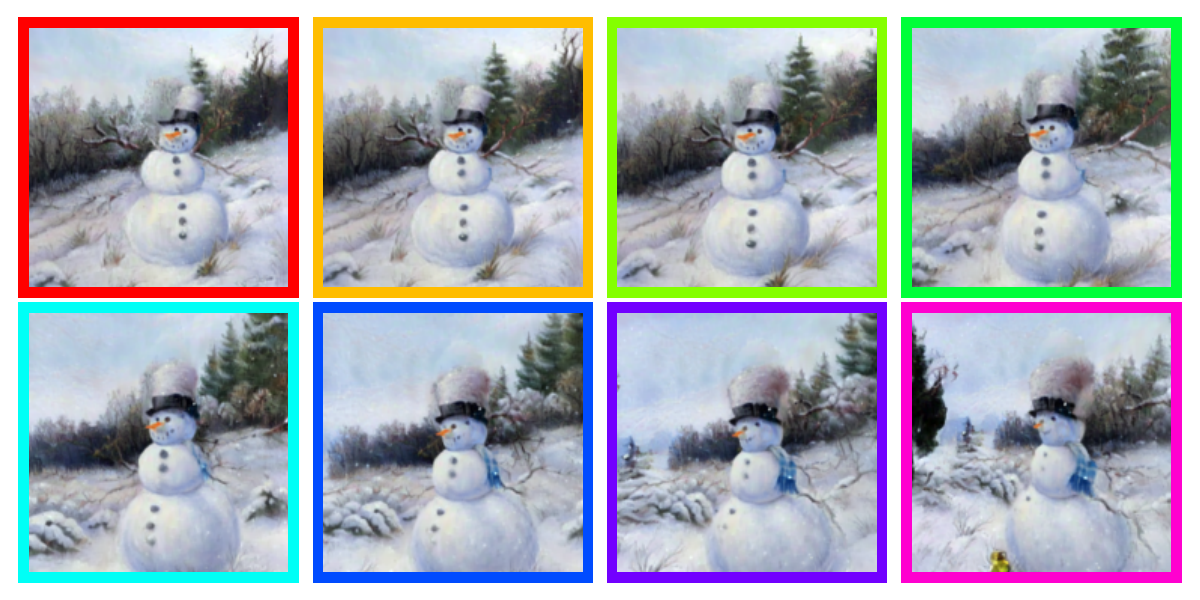} &
        \adjincludegraphics[clip,width=0.095\linewidth,trim={0 0 0 0}]{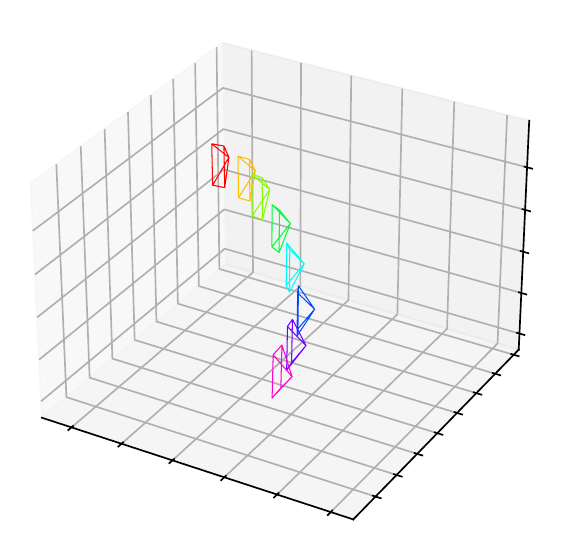} &
        \adjincludegraphics[clip,width=0.19\linewidth,trim={0 0 0 0}]{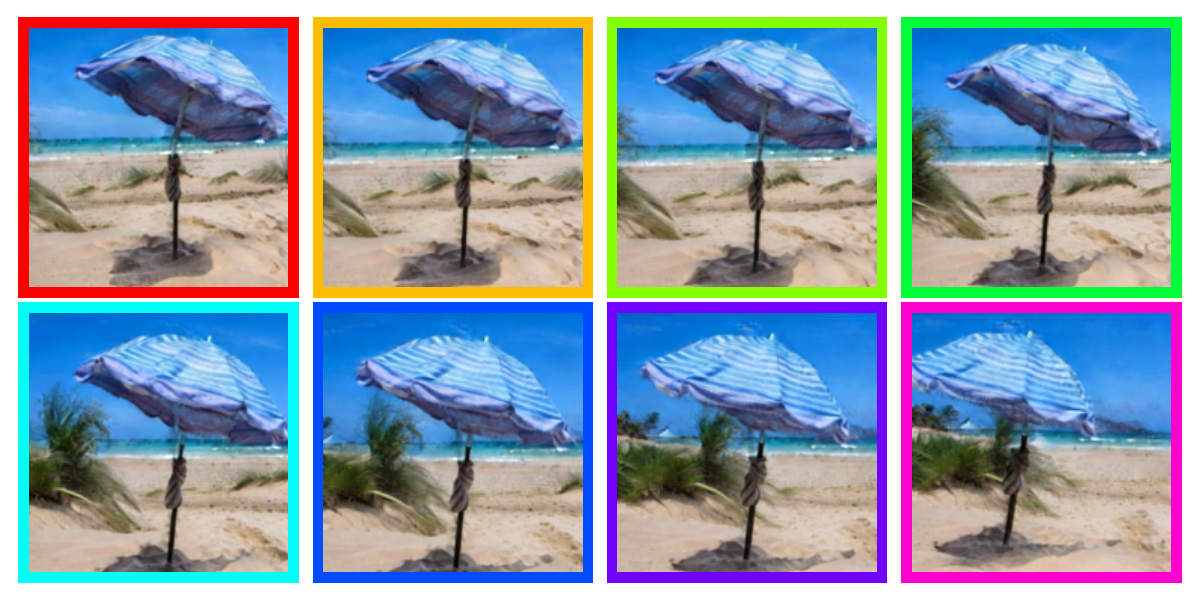} &
        \adjincludegraphics[clip,width=0.095\linewidth,trim={0 0 0 0}]{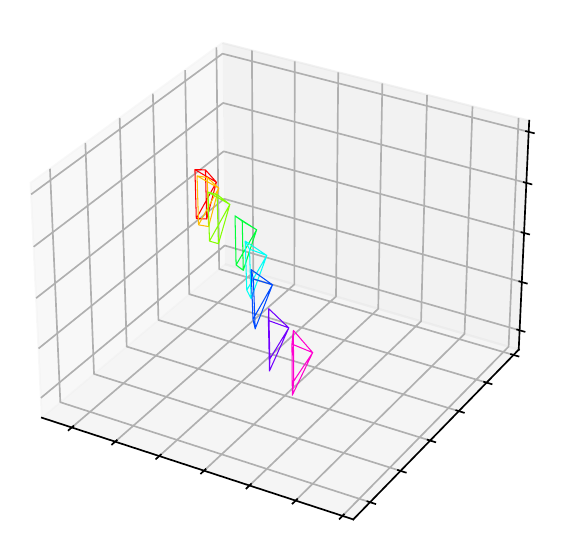} &
        \adjincludegraphics[clip,width=0.19\linewidth,trim={0 0 0 0}]{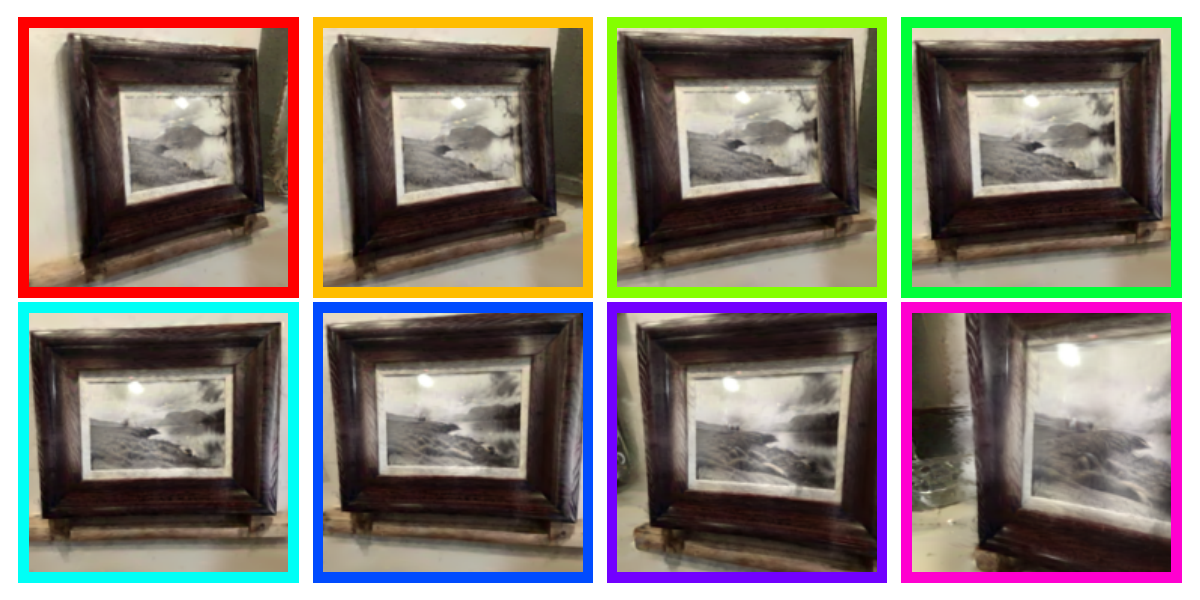} &
        \adjincludegraphics[clip,width=0.095\linewidth,trim={0 0 0 0}]{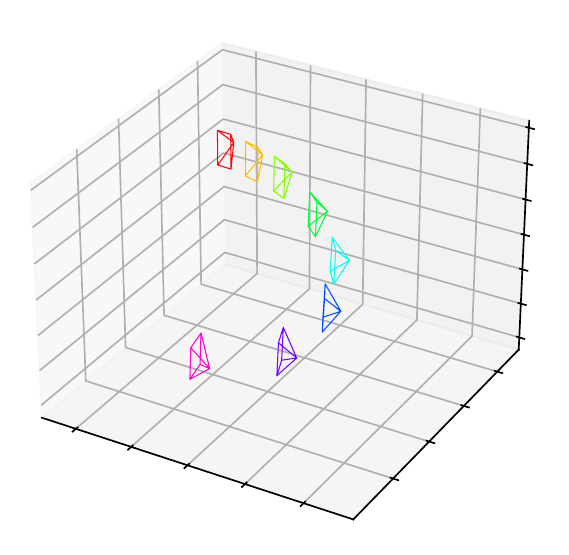} \\
        \multicolumn{2}{p{0.3\linewidth}}{\centering\scriptsize \textit{``A snowman wearing a scarf in a winter landscape"}} & \multicolumn{2}{p{0.3\linewidth}}{\centering\scriptsize \textit{``A striped beach umbrella standing tall on a sandy beach"}} & \multicolumn{2}{p{0.3\linewidth}}{\centering\scriptsize \textit{``A black and white photograph framed in dark mahogany"}} \\
        \arrayrulecolor{black}\bottomrule
    \end{tabular}
    
    \vspace{\abovefigcapmargin}
    \caption{\textbf{Additional qualitative results in 3DGS generation on MVImgNet~\cite{yu2023mvimgnet}, DL3DV~\cite{ling2024dl3dv}, and T3Bench~\cite{he2023t3bench}.} We show eight rendered scenes and camera poses from given text prompts, where image border colors match each camera.}
    \vspace{\belowfigcapmargin}
    \label{fig:additional_qual}
\end{figure*}

As shown in \cref{fig:additional_qual}, our SplatFlow generates realistic 3DGS and camera poses from various text prompts across MVImgNet~\cite{yu2023mvimgnet}, DL3DV~\cite{ling2024dl3dv}, and T3Bench~\cite{he2023t3bench}. 
Interestingly, our SplatFlow primarily produces a straight-line camera trajectory for scenery-based descriptions, while creating a circular camera trajectory for object-centric descriptions.

\begin{figure}[t!]
    \centering
    \setlength\tabcolsep{1pt}
    \begin{tabular}{cccccccc}
        \toprule
        \raisebox{0.01\linewidth}{\rotatebox{90}{\small Orig.}} 
        \adjincludegraphics[clip,width=0.11\linewidth,trim={0 0 0 0}]{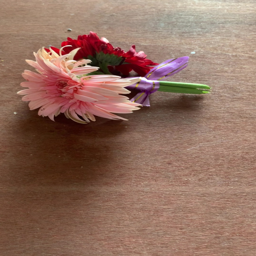} &
        \adjincludegraphics[clip,width=0.11\linewidth,trim={0 0 0 0}]{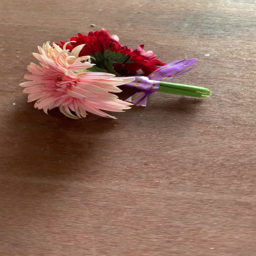} &
        \adjincludegraphics[clip,width=0.11\linewidth,trim={0 0 0 0}]{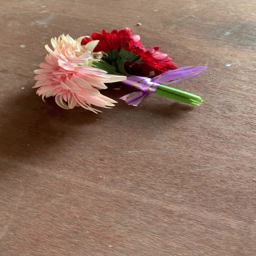} &
        \adjincludegraphics[clip,width=0.11\linewidth,trim={0 0 0 0}]{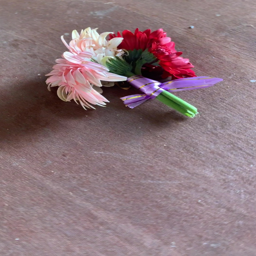} &
        \adjincludegraphics[clip,width=0.11\linewidth,trim={0 0 0 0}]{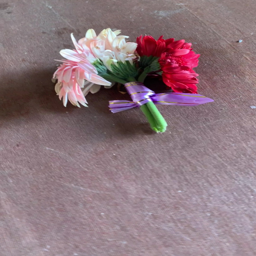} &
        \adjincludegraphics[clip,width=0.11\linewidth,trim={0 0 0 0}]{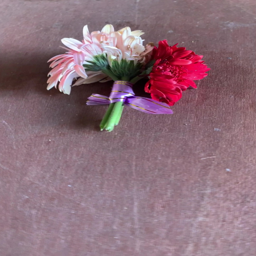} &
        \adjincludegraphics[clip,width=0.11\linewidth,trim={0 0 0 0}]{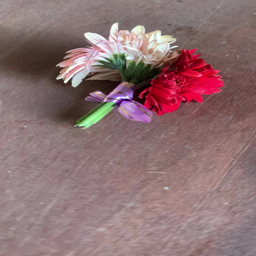} &
        \adjincludegraphics[clip,width=0.11\linewidth,trim={0 0 0 0}]{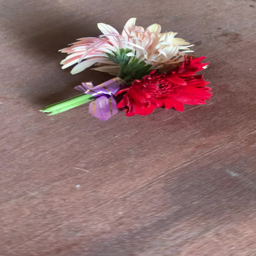}\\
        
        \raisebox{0.01\linewidth}{\rotatebox{90}{\small Edit}} 
        \adjincludegraphics[clip,width=0.11\linewidth,trim={0 0 0 0}]{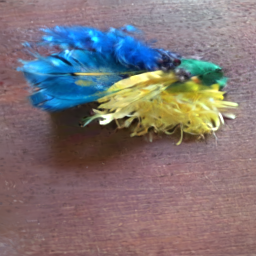} &
        \adjincludegraphics[clip,width=0.11\linewidth,trim={0 0 0 0}]{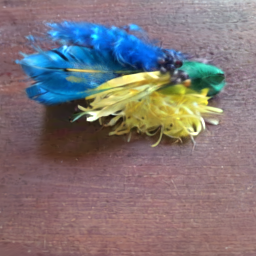} &
        \adjincludegraphics[clip,width=0.11\linewidth,trim={0 0 0 0}]{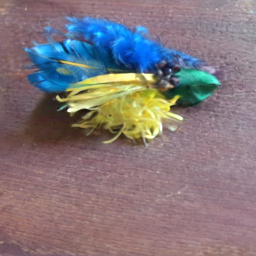} &
        \adjincludegraphics[clip,width=0.11\linewidth,trim={0 0 0 0}]{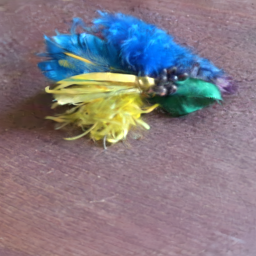} &
        \adjincludegraphics[clip,width=0.11\linewidth,trim={0 0 0 0}]{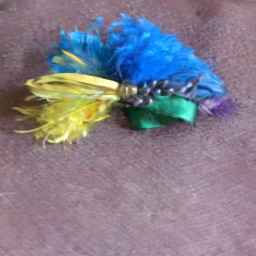} &
        \adjincludegraphics[clip,width=0.11\linewidth,trim={0 0 0 0}]{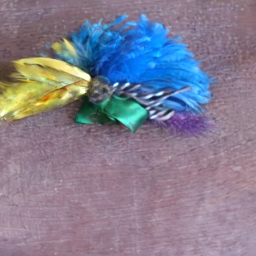} &
        \adjincludegraphics[clip,width=0.11\linewidth,trim={0 0 0 0}]{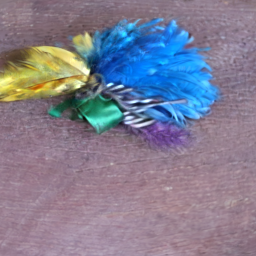} &
        \adjincludegraphics[clip,width=0.11\linewidth,trim={0 0 0 0}]{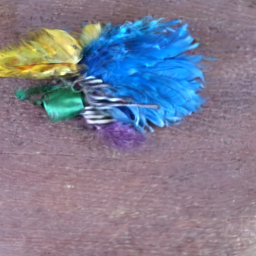}\\
        
        \multicolumn{8}{p{0.98\linewidth}}{\centering\footnotesize \quad\quad \textit{"pink and red flower"} $\to$ \textit{"blue and yellow feather"}} \\
        
        \arrayrulecolor{gray}\midrule

        \raisebox{0.01\linewidth}{\rotatebox{90}{\small Orig.}} 
        \adjincludegraphics[clip,width=0.11\linewidth,trim={0 0 0 0}]{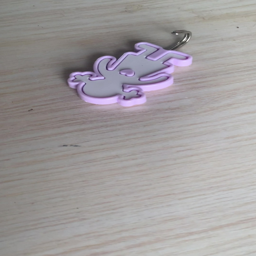} &
        \adjincludegraphics[clip,width=0.11\linewidth,trim={0 0 0 0}]{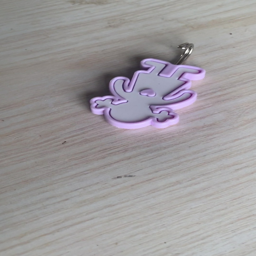} &
        \adjincludegraphics[clip,width=0.11\linewidth,trim={0 0 0 0}]{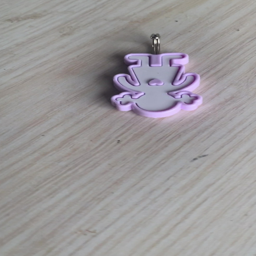} &
        \adjincludegraphics[clip,width=0.11\linewidth,trim={0 0 0 0}]{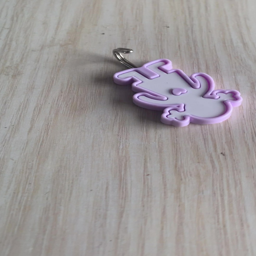} &
        \adjincludegraphics[clip,width=0.11\linewidth,trim={0 0 0 0}]{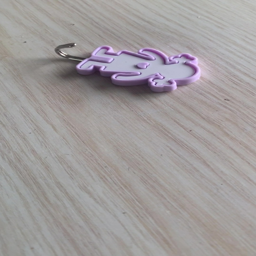} &
        \adjincludegraphics[clip,width=0.11\linewidth,trim={0 0 0 0}]{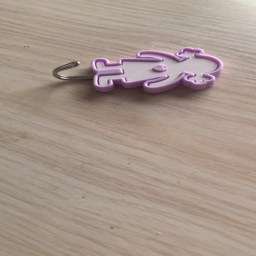} &
        \adjincludegraphics[clip,width=0.11\linewidth,trim={0 0 0 0}]{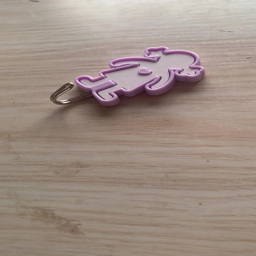} &
        \adjincludegraphics[clip,width=0.11\linewidth,trim={0 0 0 0}]{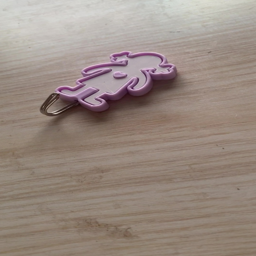} \\

        \raisebox{0.01\linewidth}{\rotatebox{90}{\small Edit}} 
        \adjincludegraphics[clip,width=0.11\linewidth,trim={0 0 0 0}]{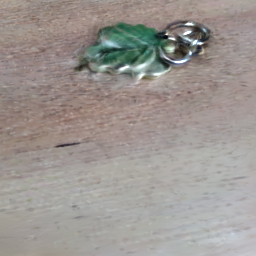} &
        \adjincludegraphics[clip,width=0.11\linewidth,trim={0 0 0 0}]{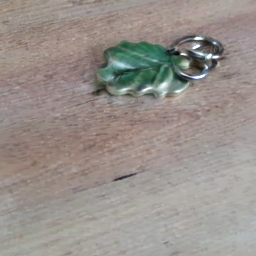} &
        \adjincludegraphics[clip,width=0.11\linewidth,trim={0 0 0 0}]{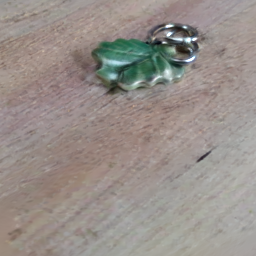} &
        \adjincludegraphics[clip,width=0.11\linewidth,trim={0 0 0 0}]{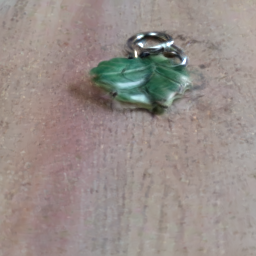} &
        \adjincludegraphics[clip,width=0.11\linewidth,trim={0 0 0 0}]{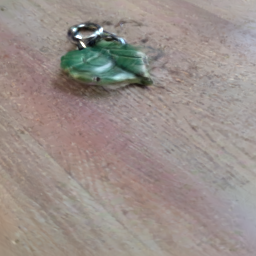} &
        \adjincludegraphics[clip,width=0.11\linewidth,trim={0 0 0 0}]{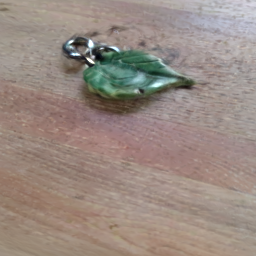} &
        \adjincludegraphics[clip,width=0.11\linewidth,trim={0 0 0 0}]{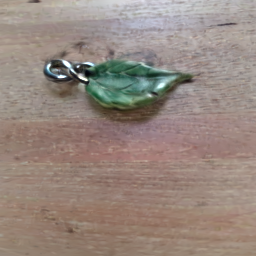} &
        \adjincludegraphics[clip,width=0.11\linewidth,trim={0 0 0 0}]{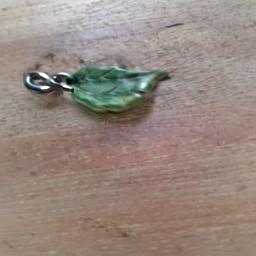} \\

        \multicolumn{8}{p{0.98\linewidth}}{\centering\footnotesize \quad\quad \textit{"purple cartoon-shaped keychain"} $\to$ \textit{"green leaf-shaped keychain"}} \\

        \midrule

        \raisebox{0.01\linewidth}{\rotatebox{90}{\small Orig.}} 
        \adjincludegraphics[clip,width=0.11\linewidth,trim={0 0 0 0}]{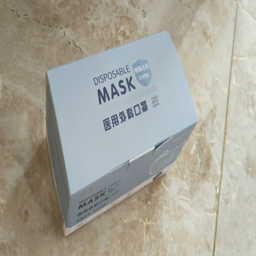} &
        \adjincludegraphics[clip,width=0.11\linewidth,trim={0 0 0 0}]{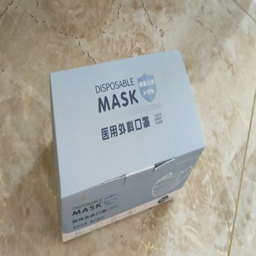} &
        \adjincludegraphics[clip,width=0.11\linewidth,trim={0 0 0 0}]{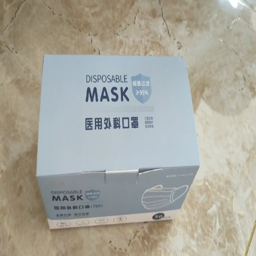} &
        \adjincludegraphics[clip,width=0.11\linewidth,trim={0 0 0 0}]{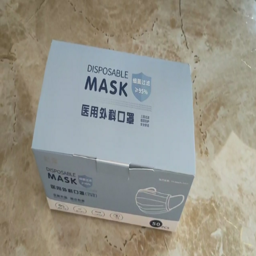} &
        \adjincludegraphics[clip,width=0.11\linewidth,trim={0 0 0 0}]{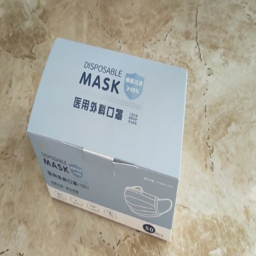} &
        \adjincludegraphics[clip,width=0.11\linewidth,trim={0 0 0 0}]{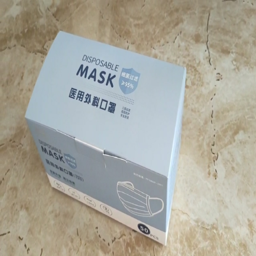} &
        \adjincludegraphics[clip,width=0.11\linewidth,trim={0 0 0 0}]{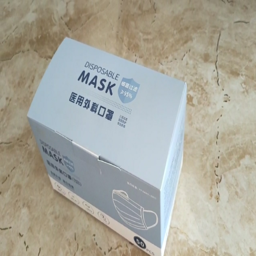} &
        \adjincludegraphics[clip,width=0.11\linewidth,trim={0 0 0 0}]{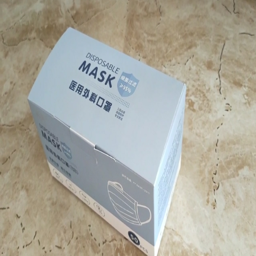} \\

        \raisebox{0.01\linewidth}{\rotatebox{90}{\small Edit}} 
        \adjincludegraphics[clip,width=0.11\linewidth,trim={0 0 0 0}]{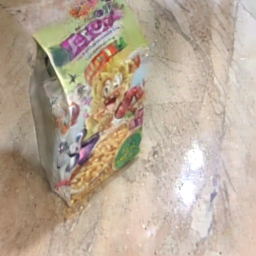} &
        \adjincludegraphics[clip,width=0.11\linewidth,trim={0 0 0 0}]{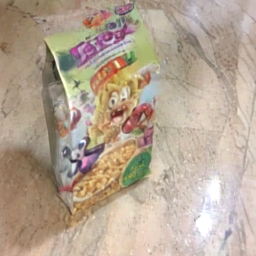} &
        \adjincludegraphics[clip,width=0.11\linewidth,trim={0 0 0 0}]{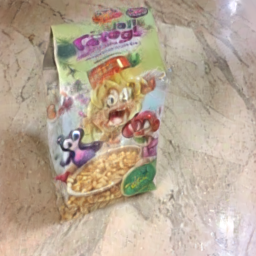} &
        \adjincludegraphics[clip,width=0.11\linewidth,trim={0 0 0 0}]{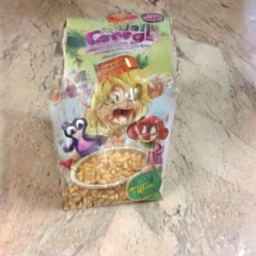} &
        \adjincludegraphics[clip,width=0.11\linewidth,trim={0 0 0 0}]{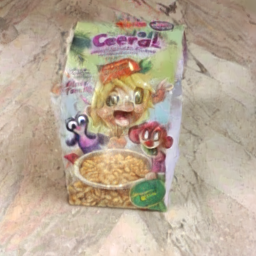} &
        \adjincludegraphics[clip,width=0.11\linewidth,trim={0 0 0 0}]{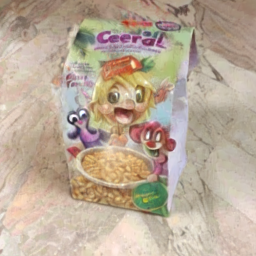} &
        \adjincludegraphics[clip,width=0.11\linewidth,trim={0 0 0 0}]{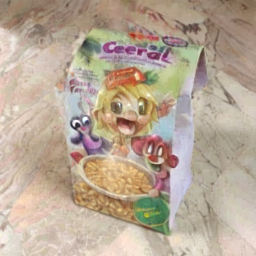} &
        \adjincludegraphics[clip,width=0.11\linewidth,trim={0 0 0 0}]{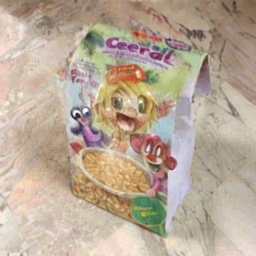} \\

        \multicolumn{8}{p{0.98\linewidth}}{\centering\footnotesize \quad\quad \textit{"mask box"} $\to$ \textit{"cereal box"}} \\
        \arrayrulecolor{black}\bottomrule
    \end{tabular}
    
    \vspace{\abovefigcapmargin}
    \caption{\textbf{Additional qualitative results in 3DGS editing.}}
    \vspace{\belowfigcapmargin}
    \label{fig:edit_supp}
\end{figure}

\subsection{More Results on 3DGS Editing}

\paragraph{3D object replacement}
Our SplatFlow enables 3DGS editing through a modified SDEdit~\cite{meng2021sdedit} combined with the inpainting process outlined in~\Cref{alg:inpainting}. Specifically, for 3D object replacement, we perform an inversion on the masked region at $t=190$ out of 200 total sampling steps, treating the remaining area as known regions by utilizing the foreground mask of the main object. Additional qualitative results on the 3D object replacement are presented in \cref{fig:edit_supp}, confirming its effectiveness in 3DGS editing on various objects.

\begin{figure}[t!]
    \centering
    \setlength\tabcolsep{1pt}
    \begin{tabular}{cccccc}
        \toprule
        \raisebox{0.03\linewidth}{\rotatebox{90}{\small Input}} 
        \adjincludegraphics[clip,width=0.15\linewidth,trim={0 0 0 0}]{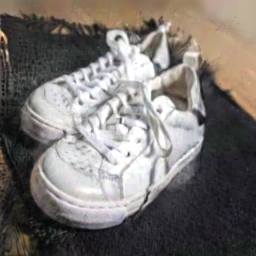} &
        \adjincludegraphics[clip,width=0.15\linewidth,trim={0 0 0 0}]{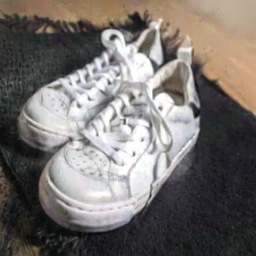} &
        \adjincludegraphics[clip,width=0.15\linewidth,trim={0 0 0 0}]{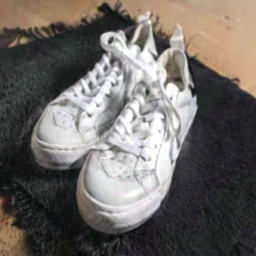} &
        \adjincludegraphics[clip,width=0.15\linewidth,trim={0 0 0 0}]{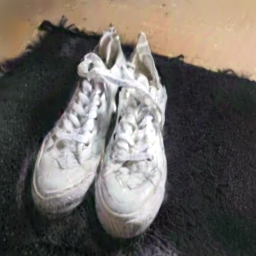} &
        \adjincludegraphics[clip,width=0.15\linewidth,trim={0 0 0 0}]{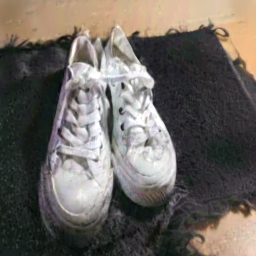} &
        \adjincludegraphics[clip,width=0.15\linewidth,trim={0 0 0 0}]{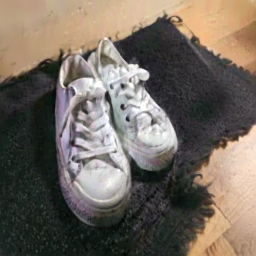} \\
        
        \raisebox{0.02\linewidth}{\rotatebox{90}{\small Guide}} 
        \adjincludegraphics[clip,width=0.15\linewidth,trim={0 0 0 0}]{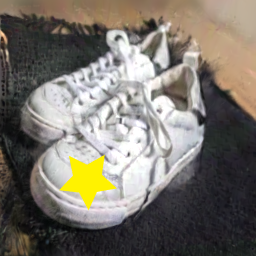} &
        \adjincludegraphics[clip,width=0.15\linewidth,trim={0 0 0 0}]{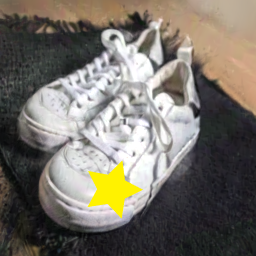} &
        \adjincludegraphics[clip,width=0.15\linewidth,trim={0 0 0 0}]{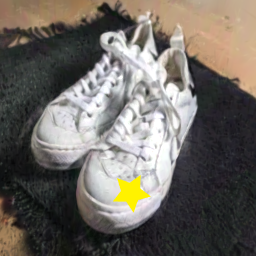} &
        \adjincludegraphics[clip,width=0.15\linewidth,trim={0 0 0 0}]{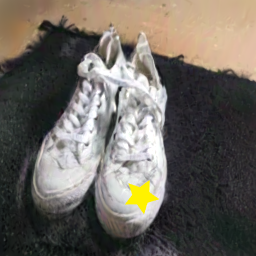} &
        \adjincludegraphics[clip,width=0.15\linewidth,trim={0 0 0 0}]{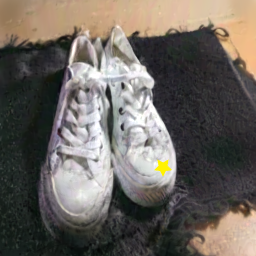} &
        \adjincludegraphics[clip,width=0.15\linewidth,trim={0 0 0 0}]{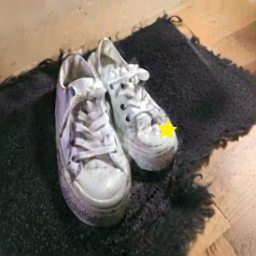} \\

        \raisebox{0.04\linewidth}{\rotatebox{90}{\small Edit}} 
        \adjincludegraphics[clip,width=0.15\linewidth,trim={0 0 0 0}]{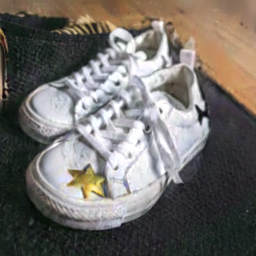} &
        \adjincludegraphics[clip,width=0.15\linewidth,trim={0 0 0 0}]{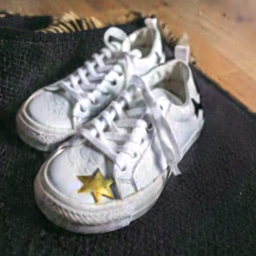} &
        \adjincludegraphics[clip,width=0.15\linewidth,trim={0 0 0 0}]{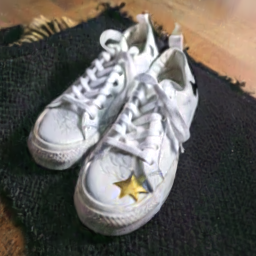} &
        \adjincludegraphics[clip,width=0.15\linewidth,trim={0 0 0 0}]{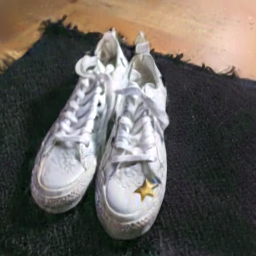} &
        \adjincludegraphics[clip,width=0.15\linewidth,trim={0 0 0 0}]{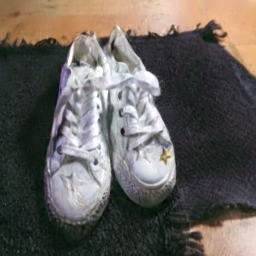} &
        \adjincludegraphics[clip,width=0.15\linewidth,trim={0 0 0 0}]{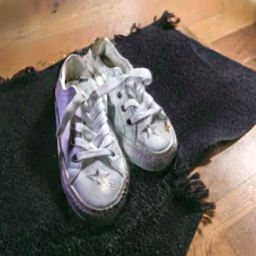} \\
        
        \multicolumn{6}{p{0.98\linewidth}}{\centering\footnotesize \quad\quad \textit{"$\dots$ sneakers $\dots$"} $\to$ \textit{"$\dots$ sneakers \textbf{\textcolor{Dandelion}{with a star mark}} $\dots$"}} \\
        
        \arrayrulecolor{gray}\midrule

        \raisebox{0.03\linewidth}{\rotatebox{90}{\small Input}} 
        \adjincludegraphics[clip,width=0.15\linewidth,trim={0 0 0 0}]{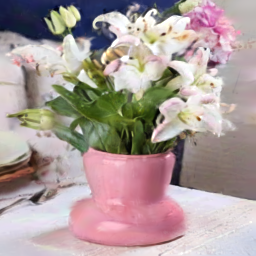} &
        \adjincludegraphics[clip,width=0.15\linewidth,trim={0 0 0 0}]{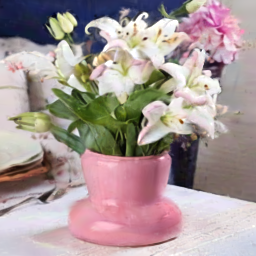} &
        \adjincludegraphics[clip,width=0.15\linewidth,trim={0 0 0 0}]{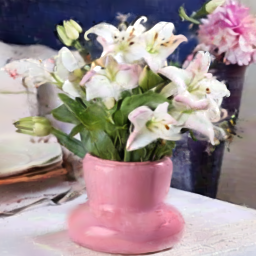} &
        \adjincludegraphics[clip,width=0.15\linewidth,trim={0 0 0 0}]{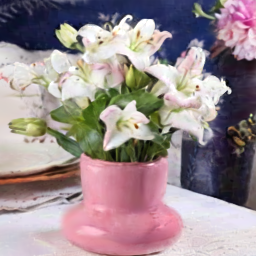} &
        \adjincludegraphics[clip,width=0.15\linewidth,trim={0 0 0 0}]{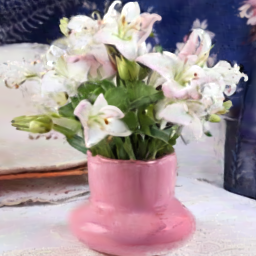} &
        \adjincludegraphics[clip,width=0.15\linewidth,trim={0 0 0 0}]{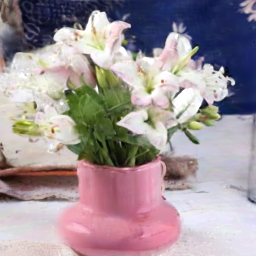} \\
        
        \raisebox{0.02\linewidth}{\rotatebox{90}{\small Guide}} 
        \adjincludegraphics[clip,width=0.15\linewidth,trim={0 0 0 0}]{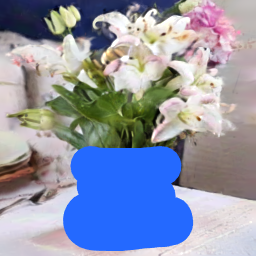} &
        \adjincludegraphics[clip,width=0.15\linewidth,trim={0 0 0 0}]{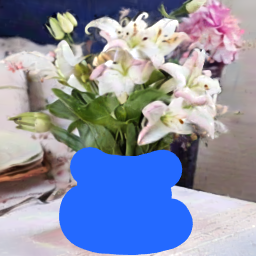} &
        \adjincludegraphics[clip,width=0.15\linewidth,trim={0 0 0 0}]{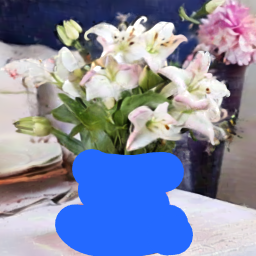} &
        \adjincludegraphics[clip,width=0.15\linewidth,trim={0 0 0 0}]{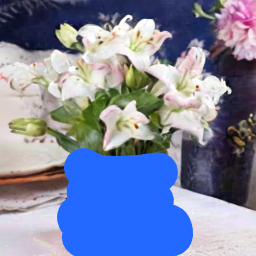} &
        \adjincludegraphics[clip,width=0.15\linewidth,trim={0 0 0 0}]{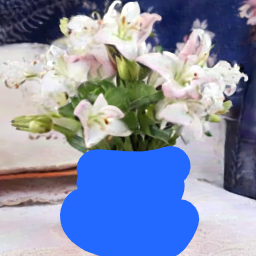} &
        \adjincludegraphics[clip,width=0.15\linewidth,trim={0 0 0 0}]{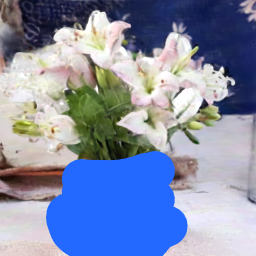} \\
        
        \raisebox{0.04\linewidth}{\rotatebox{90}{\small Edit}} 
        \adjincludegraphics[clip,width=0.15\linewidth,trim={0 0 0 0}]{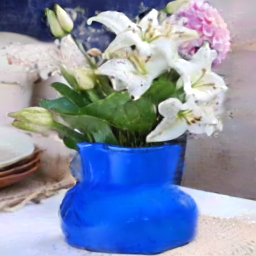} &
        \adjincludegraphics[clip,width=0.15\linewidth,trim={0 0 0 0}]{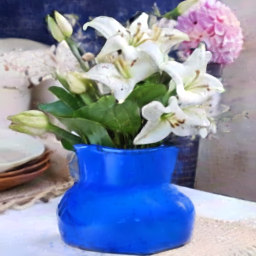} &
        \adjincludegraphics[clip,width=0.15\linewidth,trim={0 0 0 0}]{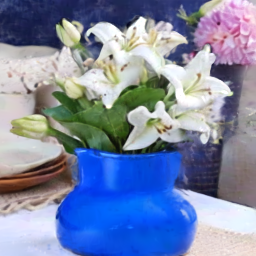} &
        \adjincludegraphics[clip,width=0.15\linewidth,trim={0 0 0 0}]{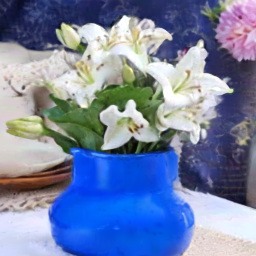} &
        \adjincludegraphics[clip,width=0.15\linewidth,trim={0 0 0 0}]{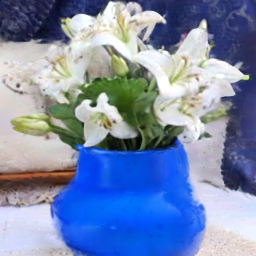} &
        \adjincludegraphics[clip,width=0.15\linewidth,trim={0 0 0 0}]{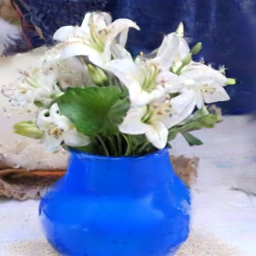} \\
        
        \multicolumn{6}{p{0.98\linewidth}}{\centering\footnotesize \quad\quad \textit{"$\dots$ a \textbf{\textcolor{Rhodamine}{pink}} ceramic vase $\dots$"} $\to$ \textit{"$\dots$ a \textbf{\textcolor{blue}{blue}} ceramic vase $\dots$"}} \\

        \arrayrulecolor{black}\bottomrule
    \end{tabular}
    
    \vspace{\abovefigcapmargin}
    \caption{\textbf{Qualitative results on 3DGS editing with user-provided strokes.} Despite the rough strokes applied to the rendered scene, our SplatFlow enables seamless and natural 3DGS editing.}
    \vspace{-1mm}
    \label{fig:sdedit_appendix}
\end{figure}

\vspace{\paramargin}
\paragraph{3DGS editing with strokes}
Interestingly, our SplatFlow can selectively edit specific portions of the generated 3DGS based on user-provided input strokes.  Specifically, we perform an inversion on all rendered multi-view images with strokes at $t=100$ out of 200 total sampling steps, followed by denoising the latents using edited captions. As shown in~\cref{fig:sdedit_appendix}, even with rough stroke inputs that are 3D inconsistent, the edited 3DGS maintains a highly natural appearance.

\section{Anaylsis and Discussion}

\subsection{Depth Map Visualization}

\begin{figure}[t!]
    \centering
    \setlength\tabcolsep{2pt}
    \begin{tabular}{cccccc}
        \toprule
        \raisebox{0.03\linewidth}{\rotatebox{90}{\small RGB}} 
        \adjincludegraphics[clip,width=0.14\linewidth,trim={0 0 0 0}]{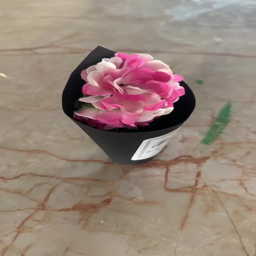} &
        \adjincludegraphics[clip,width=0.14\linewidth,trim={0 0 0 0}]{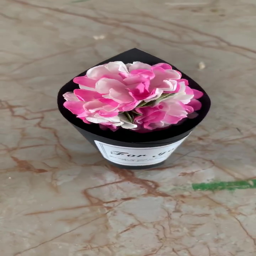} &
        \adjincludegraphics[clip,width=0.14\linewidth,trim={0 0 0 0}]{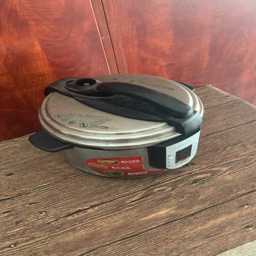} &
        \adjincludegraphics[clip,width=0.14\linewidth,trim={0 0 0 0}]{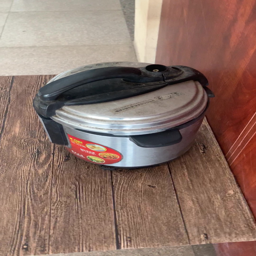} &
        \adjincludegraphics[clip,width=0.14\linewidth,trim={0 0 0 0}]{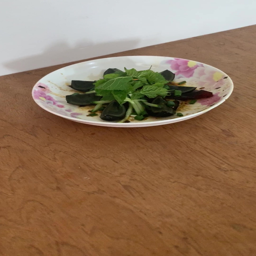} &
        \adjincludegraphics[clip,width=0.14\linewidth,trim={0 0 0 0}]{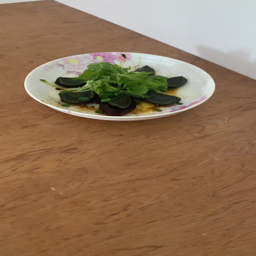}\\
        
        \raisebox{0.01\linewidth}{\rotatebox{90}{\small GT~\cite{yang2024depth}}} 
        \adjincludegraphics[clip,width=0.14\linewidth,trim={0 0 0 0}]{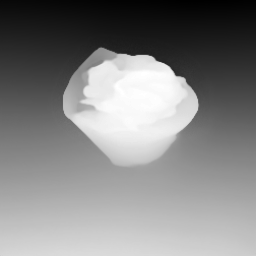} &
        \adjincludegraphics[clip,width=0.14\linewidth,trim={0 0 0 0}]{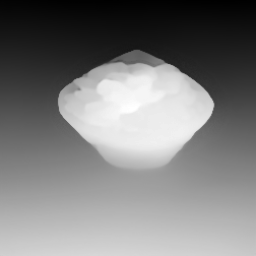} &
        \adjincludegraphics[clip,width=0.14\linewidth,trim={0 0 0 0}]{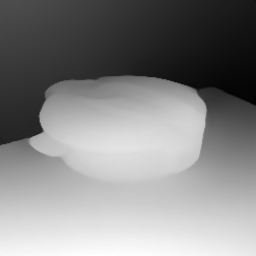} &
        \adjincludegraphics[clip,width=0.14\linewidth,trim={0 0 0 0}]{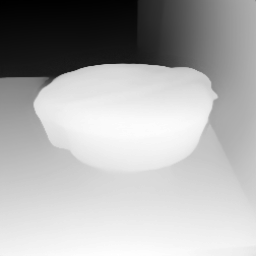} &
        \adjincludegraphics[clip,width=0.14\linewidth,trim={0 0 0 0}]{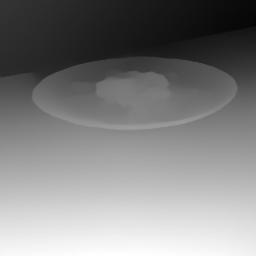} &
        \adjincludegraphics[clip,width=0.14\linewidth,trim={0 0 0 0}]{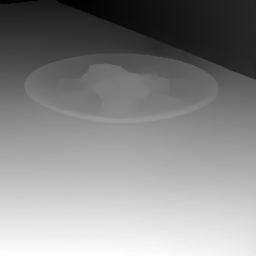}\\
        
        \raisebox{0.04\linewidth}{\rotatebox{90}{\small Ours}} 
        \adjincludegraphics[clip,width=0.14\linewidth,trim={0 0 0 0}]{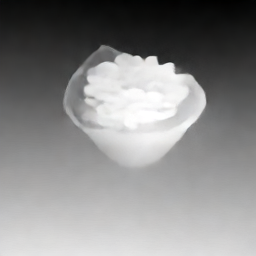} &
        \adjincludegraphics[clip,width=0.14\linewidth,trim={0 0 0 0}]{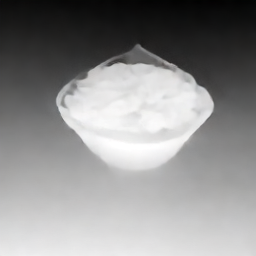} &
        \adjincludegraphics[clip,width=0.14\linewidth,trim={0 0 0 0}]{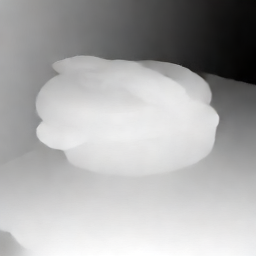} &
        \adjincludegraphics[clip,width=0.14\linewidth,trim={0 0 0 0}]{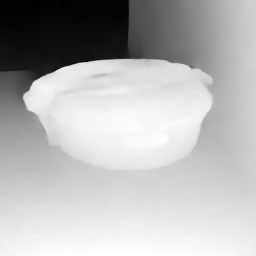} &
        \adjincludegraphics[clip,width=0.14\linewidth,trim={0 0 0 0}]{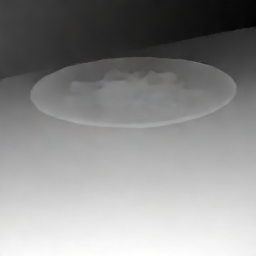} &
        \adjincludegraphics[clip,width=0.14\linewidth,trim={0 0 0 0}]{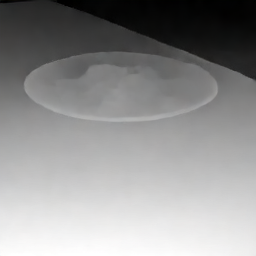}\\
        \bottomrule
    \end{tabular}

    \vspace{\abovefigcapmargin}
    \caption{\textbf{Depth map visualization.} Given multi-view images, we show generated depths that are jointly inpainted with camera poses.}
    \vspace{\belowfigcapmargin}
    \label{fig:repaint_depth}
\end{figure}

Note that better performance is achieved when the depth map from DepthAnythingV2~\cite{yang2024depth} is not used for camera pose estimation. \Cref{fig:repaint_depth} shows the generated depth maps obtained by jointly generating depth and ray latents from multi-view images. Notably, the generated depth maps capture the details of the given images more effectively than the ground truth provided by DepthAnythingV2~\cite{yang2024depth}. Therefore, more detailed depth maps allow our multi-view RF to achieve more accurate camera pose estimation.

\subsection{3D Consistency Analysis for the GSDecoder}
\begin{figure*}[t!]
    \centering
    \setlength\tabcolsep{0.5pt}
    \begin{tabular}{cccccccc}
        \toprule

        \raisebox{0.0\linewidth}{\rotatebox{90}{\small Multi-View Gen}} 
        \adjincludegraphics[clip,width=0.12\linewidth,trim={0 0 0 0}]{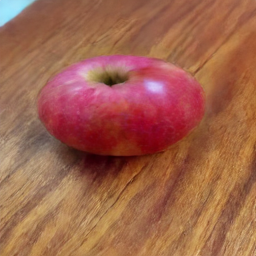} &
        \adjincludegraphics[clip,width=0.12\linewidth,trim={0 0 0 0}]{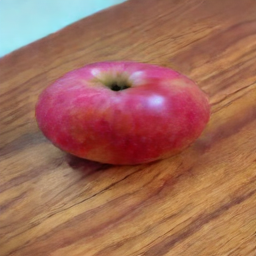} &
        \adjincludegraphics[clip,width=0.12\linewidth,trim={0 0 0 0}]{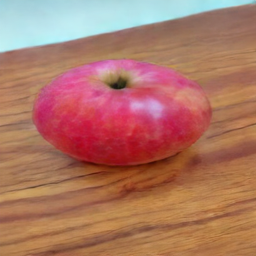} &
        \adjincludegraphics[clip,width=0.12\linewidth,trim={0 0 0 0}]{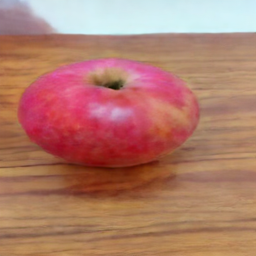} &
        \adjincludegraphics[clip,width=0.12\linewidth,trim={0 0 0 0}]{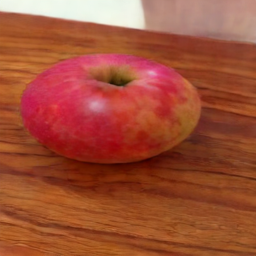} &
        \adjincludegraphics[clip,width=0.12\linewidth,trim={0 0 0 0}]{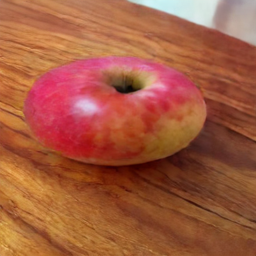} &
        \adjincludegraphics[clip,width=0.12\linewidth,trim={0 0 0 0}]{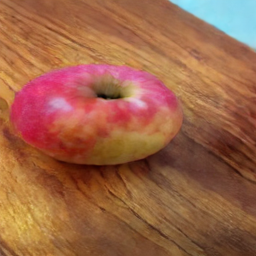} &
        \adjincludegraphics[clip,width=0.12\linewidth,trim={0 0 0 0}]{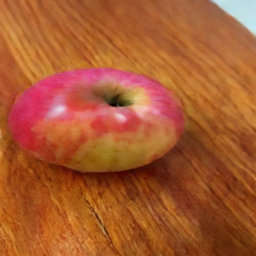} \\
        
        \raisebox{0.0\linewidth}{\rotatebox{90}{\small Rendered Img}} 
        \adjincludegraphics[clip,width=0.12\linewidth,trim={0 0 0 0}]{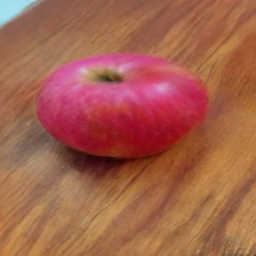} &
        \adjincludegraphics[clip,width=0.12\linewidth,trim={0 0 0 0}]{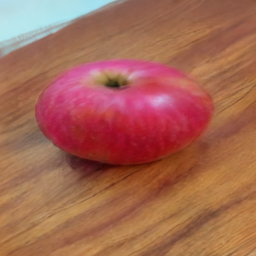} &
        \adjincludegraphics[clip,width=0.12\linewidth,trim={0 0 0 0}]{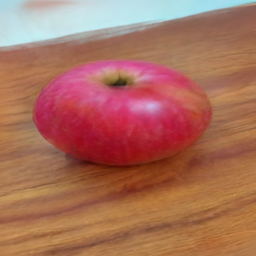} &
        \adjincludegraphics[clip,width=0.12\linewidth,trim={0 0 0 0}]{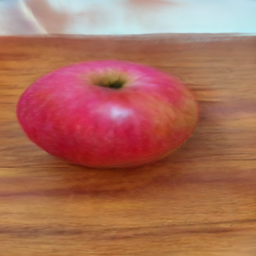} &
        \adjincludegraphics[clip,width=0.12\linewidth,trim={0 0 0 0}]{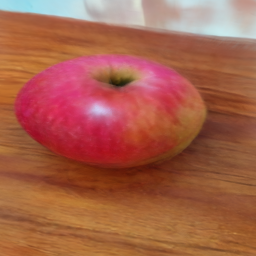} &
        \adjincludegraphics[clip,width=0.12\linewidth,trim={0 0 0 0}]{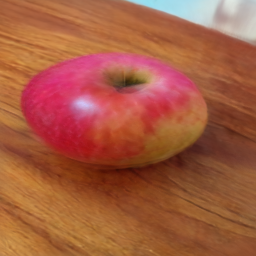} &
        \adjincludegraphics[clip,width=0.12\linewidth,trim={0 0 0 0}]{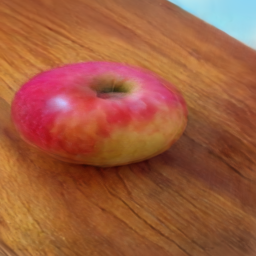} &
        \adjincludegraphics[clip,width=0.12\linewidth,trim={0 0 0 0}]{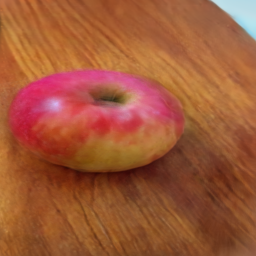} \\
        
        \multicolumn{8}{p{0.98\linewidth}}{\centering \textit{Red apple on wooden surface}} \\
        
        \arrayrulecolor{gray}\midrule
        
        \raisebox{0.0\linewidth}{\rotatebox{90}{\small Multi-View Gen}} 
        \adjincludegraphics[clip,width=0.12\linewidth,trim={0 0 0 0}]{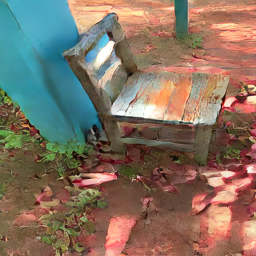} &
        \adjincludegraphics[clip,width=0.12\linewidth,trim={0 0 0 0}]{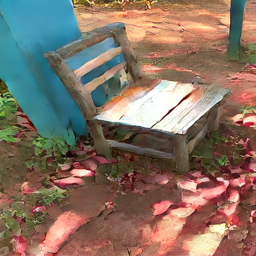} &
        \adjincludegraphics[clip,width=0.12\linewidth,trim={0 0 0 0}]{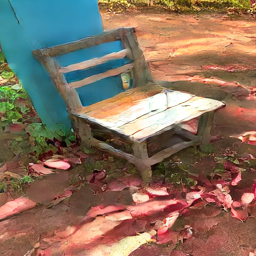} &
        \adjincludegraphics[clip,width=0.12\linewidth,trim={0 0 0 0}]{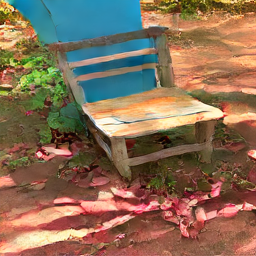} &
        \adjincludegraphics[clip,width=0.12\linewidth,trim={0 0 0 0}]{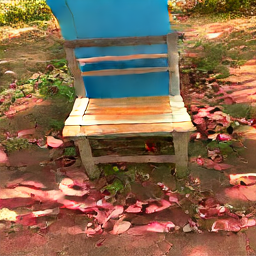} &
        \adjincludegraphics[clip,width=0.12\linewidth,trim={0 0 0 0}]{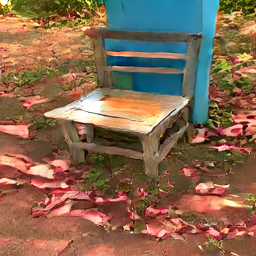} &
        \adjincludegraphics[clip,width=0.12\linewidth,trim={0 0 0 0}]{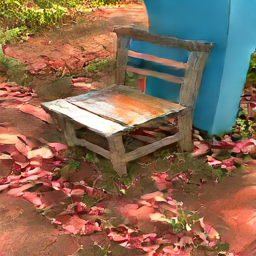} &
        \adjincludegraphics[clip,width=0.12\linewidth,trim={0 0 0 0}]{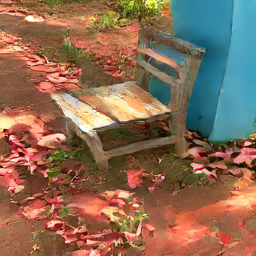} \\
        
        \raisebox{0.0\linewidth}{\rotatebox{90}{\small Rendered Img}} 
        \adjincludegraphics[clip,width=0.12\linewidth,trim={0 0 0 0}]{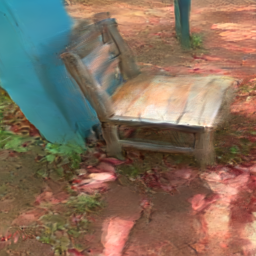} &
        \adjincludegraphics[clip,width=0.12\linewidth,trim={0 0 0 0}]{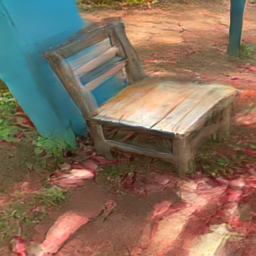} &
        \adjincludegraphics[clip,width=0.12\linewidth,trim={0 0 0 0}]{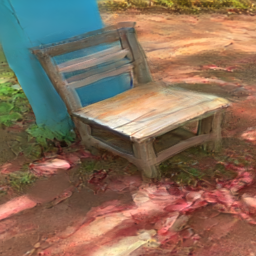} &
        \adjincludegraphics[clip,width=0.12\linewidth,trim={0 0 0 0}]{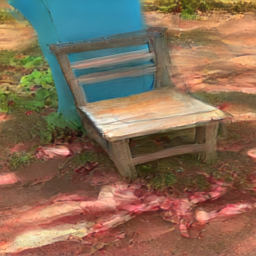} &
        \adjincludegraphics[clip,width=0.12\linewidth,trim={0 0 0 0}]{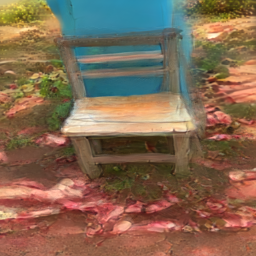} &
        \adjincludegraphics[clip,width=0.12\linewidth,trim={0 0 0 0}]{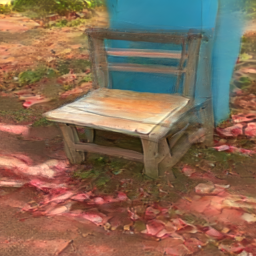} &
        \adjincludegraphics[clip,width=0.12\linewidth,trim={0 0 0 0}]{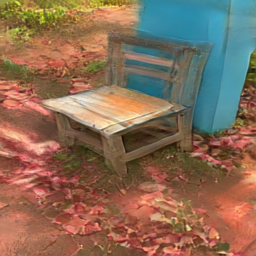} &
        \adjincludegraphics[clip,width=0.12\linewidth,trim={0 0 0 0}]{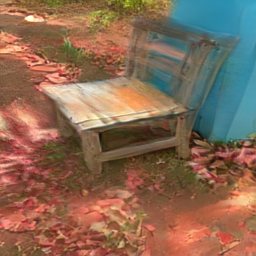} \\

        \multicolumn{8}{p{0.98\linewidth}}{\centering \quad\quad \textit{Wooden chair with blue poles, leaves on ground}} \\
        
        \arrayrulecolor{black}\bottomrule
    \end{tabular}

    \caption{\textbf{GSDecoder analysis.} Comparison between multi-view generated images (top row of each pair) and 3D Gaussian Splatting (3DGS) rendered outputs (bottom row of each pair) using the GSDecoder. The GSDecoder helps smooth out inconsistencies present in the multi-view generated images, such as variations in color and shape, resulting in more coherent 3D representations. For example, inconsistencies in the apple's color and shape, and in the seat cushion's color of the wooden chair, are corrected.}
    \vspace{\belowfigcapmargin}
    \label{fig:gsdecoder_analysis}
\end{figure*}

In this section, we explore the role of the GSDecoder by comparing the multi-view generated images with the 3D Gaussian Splatting (3DGS) rendered outputs, as shown in Fig.~\ref{fig:gsdecoder_analysis}. The multi-view images are decoded using the Stable Diffusion 3 decoder, while the 3DGS is rendered using our GSDecoder in a feed-forward manner. Our objective is to analyze how the GSDecoder contributes to achieving consistency across the 3D space by transforming the multi-view latent representations into a coherent 3D representation.

From our analysis, we observe that the GSDecoder can help mitigate inconsistencies present in the multi-view generated images. Specifically, there are instances where slight variations in the appearance of objects, such as colors or shapes, are noticeable in the multi-view generated images, which can lead to a lack of 3D coherence in the generated scene. The GSDecoder processes these inconsistencies by smoothing them, resulting in more consistent 3DGS parameters across multiple views.
In the case of the red apple on a wooden surface, the GSDecoder addresses inconsistencies in the apple's color and shape that are evident in the multi-view images. For the wooden chair with blue poles and leaves on the ground, the GSDecoder helps align the color of the seat cushion, making it more consistent across different views. Overall, the GSDecoder prioritizes 3D consistency at the cost of slightly blurring some details, ultimately enhancing the coherence of the generated scene.

\subsection{Discussions}

\paragraph{Inference time}
It takes about 20 seconds to generate one 3D scene from text prompts and 5 minutes for the refining process (\ie, SDS++~\cite{li2024director3d}). This is similar to Director3D~\cite{li2024director3d} which also utilizes the same refining process.

\paragraph{Future works and limitations} As our SplatFlow is trained only on MVImgNet~\cite{yu2023mvimgnet} and DL3DV~\cite{ling2024dl3dv}, its generalizability to unseen text prompts has room for improvement. Specifically, training with a large text-to-image dataset~\cite{schuhmann2022laion} and other multi-view image datasets~\cite{reizenstein2021common} can improve the generalizability. Additionally, we believe that the inversion technique or inpainting method suitable for the rectified flow model can be combined with our multi-view RF model to achieve better quality 3DGS editing.

\paragraph{Ethical Considerations}
Advancements in 3D generation technology raise several ethical considerations that require careful attention. A significant concern is the potential misuse of generated 3D content, as it could be exploited to create deceptive or misleading visuals. Fabricated content could be presented as authentic, potentially leading to harm or misinformation. To prevent the misuse of this technology, it is crucial to establish clear guidelines for responsible use and enforce ethical standards. Implementing robust safeguards and obtaining informed consent, especially when processing images that contain personal information, are crucial steps to prevent the misuse of this technology.

\subsection{Extended Related Works}
We sincerely appreciate the reviewers’ valuable feedback and have decided to add relevant discussions in this section.
Previous works have attempted to ensure 3D consistency by leveraging diffusion models~\cite{anciukevivcius2024denoising, szymanowicz2023viewset}. 
Concurrent works have further taken a step toward direct 3D Gaussian Splatting (3DGS) by replacing conventional diffusion decoders with Gaussian Splatting-based decoders~\cite{lin2025diffsplat, liang2024wonderland, szymanowicz2025bolt3d, go2025videorfsplatdirectsceneleveltextto3d, yang2024prometheus,schwarz2025generative, park2025steerx}.
We believe this direction is not only more versatile but also paves the way for more straightforward and effective editing through generative modeling.

\end{document}